\documentclass[acmlarge,natbib,screen]{acmart}

\AtBeginDocument{%
  \providecommand\BibTeX{{%
    \normalfont B\kern-0.5em{\scshape i\kern-0.25em b}\kern-0.8em\TeX}}}

\setcopyright{acmcopyright}
\copyrightyear{2023}
\acmYear{2023}
\acmDOI{10.1145/3635718}

\acmJournal{TKDD}
\usepackage{amsmath}
\usepackage{flushend}
\usepackage{caption}
\usepackage{tabularx}
\usepackage{graphicx}
\usepackage{subfigure}
\usepackage{amsthm}
\usepackage{algorithm}
\usepackage{algorithmic}
\usepackage{multirow}
\usepackage{xcolor}
\usepackage{booktabs}
\usepackage{float}
\usepackage{bm}
\usepackage{xspace}
\usepackage{threeparttable}

\usepackage{amssymb}
\usepackage{url}
\usepackage[misc]{ifsym}
\usepackage{enumerate}
\usepackage{balance}
\usepackage{ulem}
\usepackage{makecell}
\usepackage{colortbl}

\newcommand{\ie}{\textit{i}.\textit{e}.}
\newcommand{\eg}{\textit{e}.\textit{g}.} 
\newcommand{\wrt}{\textit{w}.\textit{r}.\textit{t}} 

\newtheorem{Def}{Definition}

\newtheorem*{Pro*}{Problem}

\newcommand{\model}{AttnTUL\xspace}
\newcommand{\modell}{TUL-L\xspace}
\newcommand{\modelg}{TUL-G\xspace}
\newcommand{\modelsa}{TUL-SA\xspace}
\newcommand{\modelea}{TUL-EA\xspace}
\newcommand{\modelts}{TUL-TS\xspace}

\begin{document}

% \balance
% \input{coverletter}

\title{Trajectory-User Linking via Hierarchical Spatio-Temporal Attention Networks}

% \author{
% \IEEEauthorblockN{
% Wei Chen{\textsuperscript{1}}, Chao Huang{\textsuperscript{2}},  Yanwei Yu{\textsuperscript{1, \Letter}}, Yongguo Jiang{\textsuperscript{1}},  Junyu Dong{\textsuperscript{1}}}\\
% \IEEEauthorblockA{
% \textsuperscript{1}College of Computer Science and Technology, Ocean University of China, Qingdao, China\\
% \textsuperscript{2}Department of Computer Science, The University of Hong Kong, Hong Kong, China\\
% weichen@stu.ouc.edu.cn, chuang@cs.hku.hk,\\ \{yuyanwei, jiangyg,  dongjunyu\}@ouc.edu.cn}\\
% \thanks{Corresponding author: Yanwei Yu.}
% }

\author{Wei~Chen}
\email{onedeanxxx@gmail.com}
\affiliation{%
  \institution{Ocean University of China}
  \streetaddress{Songling RD 238}
  \city{Qingdao}
  \state{Shandong}
  \country{China}
  \postcode{266100}
}

\author{Chao~Huang}
\email{chaohuang75@gmail.com}
\affiliation{%
  \institution{The University of Hong Kong}
  \streetaddress{Pokfulam}
  \city{Hong Kong}
  \country{China}
}

\author{Yanwei~Yu}
\authornote{Corresponding author.}
\email{yuyanwei@ouc.edu.cn}
\affiliation{%
  \institution{Ocean University of China}
  \streetaddress{Songling RD 238}
  \city{Qingdao}
  \state{Shandong}
  \country{China}
  \postcode{266100}
}

\author{Yongguo~Jiang}
\email{jiangyg@ouc.edu.cn}
\affiliation{%
  \institution{Ocean University of China}
  \streetaddress{Songling RD 238}
  \city{Qingdao}
  \state{Shandong}
  \country{China}
  \postcode{266100}
}

\author{Junyu~Dong}
\email{dongjunyu@ouc.edu.cn}
\affiliation{
  \institution{Ocean University of China}
  \streetaddress{Songling RD 238}
  \city{Qingdao}
  \state{Shandong}
  \country{China}
  \postcode{266100}
}

\renewcommand{\shortauthors}{Wei Chen et al.}

\begin{abstract}

Trajectory-User Linking (TUL) is crucial for human mobility modeling by linking different trajectories to users with the exploration of complex mobility patterns. Existing works mainly rely on the recurrent neural framework to encode the temporal dependencies in trajectories, have fall short in capturing spatial-temporal global context for TUL prediction. To fill this gap, this work presents a new hierarchical spatio-temporal attention neural network, called \textit{\model}, to jointly encode the local trajectory transitional patterns and global spatial dependencies for TUL. Specifically, our first model component is built over the graph neural architecture to preserve the local and global context and enhance the representation paradigm of geographical regions and user trajectories. Additionally, a hierarchically structured attention network is designed to simultaneously encode the intra-trajectory and inter-trajectory dependencies, with the integration of the temporal attention mechanism and global elastic attentional encoder. Extensive experiments demonstrate the superiority of our \model method as compared to state-of-the-art baselines on various trajectory datasets.
The source code of our model is available at \url{https://github.com/Onedean/AttnTUL}. 

% Trajectory-User Linking (TUL), which links trajectories to users, has been a challenging problem due to the high dimensionality, sparsity, and hierarchical structures in mobility data. 
% Existing works mainly rely on Recurrent Neural Network (RNN), which fails to model long-term dependencies in trajectories, and ignores the contribution of global spatial modeling to TUL prediction. In this paper, we present a novel hierarchical spatio-temporal attention neural network, called \textit{\model}, which simultaneously models the local sequential transition patterns and global spatial correlations for solving the TUL task. 
% Specifically, we first model both micro and macro characteristics of users' trajectories by constructing local and global spatial graphs, and use Graph Convolutional Network (GCN) to learn embeddings for the divided grids and trajectories respectively. 
% Second, we design a hierarchical attention network, which is composed of a temporal self-attention encoder and a global elastic attention encoder to obtain local representation that incorporates sequential dependencies within the trajectory and global representation that captures the complex correlations among trajectories for each trajectory, respectively. 
% Finally, we use a linking layer to fuse the two representations to classify trajectories by users.
% Extensive experiments show that \model yields significant improvements over state-of-the-art baselines on three types of real-life mobility datasets (\ie, check-in dataset, human GPS trajectory dataset, and vehicle GPS trajectory dataset). 

\end{abstract}




\maketitle

% \vspace{-2mm}
\section{Introduction}

The development of mobile computing techniques enables the collection of massive mobility data from various sources, including location-based social networks, geo-tagged social media and GPS enabled mobile applications~\cite{zheng2015trajectory,wang2020deep}. Among various mobility modeling applications, a recently introduced \textit{Trajectory-User Linking} (TUL) task, which aims to link anonymous trajectories to users who generate them, is crucial and beneficial for a broad range of spatial-temporal mining applications, ranging from location-based recommendations~\cite{liu2019geo} to urban anomaly detection~\cite{zhang2019decomposition,2018deepcrime}. For example, vehicle-mounted terminals and ride-sharing services usually anonymize user identities in collecting mobility data due to the privacy issue. Effective linking trajectories with their corresponding users can not only enable accurate location-based recommendations, but also identify abnormal events from users' GPS traces. Therefore, the linking results between users and their anonymous trajectories are important for advancing the business intelligence and smart city applications~\cite{chen2021curriculum,feng2019dplink}. 

% Nowadays, massive mobility data can be collected from a variety of sources including location-sharing social networks, geo-tagged social media, location-based online services, and other GPS enabled smartphone applications. 
% These mobility data provides us with unprecedented information to understand human mobility patterns and stimulates a number of trajectory mining tasks for various applications~\citep{zheng2015trajectory,wang2020deep}. 
% \textit{Trajectory-User Linking} (TUL), which links trajectories to users who generate them, is a recently introduced trajectory mining task with a broad range of applications, ranging from personalized recommendation systems, location-based services, to potential criminals identification~\citep{gao2017identifying,feng2018deepmove}. 
% %~\cite{gao2017identifying,feng2020predicting}
% For example, vehicle-mounted terminals and location-sharing service Apps (\eg, DiDi and Uber) usually anonymize user identities for the sake of privacy when collecting users' trajectories. 
% However, linking trajectories with their corresponding users plays a crucial role in various trajectory mining applications, \eg, it helps to make more accurate and personalized location recommendations from check-in data~\citep{chen2021curriculum}, and it may help in identifying the potential criminals from users' GPS traces. 
% In addition, it is in a great need of linking user identities across multiple online services to fuse the separated data to provide better business intelligence~\citep{feng2019dplink}. 

\begin{figure}[h]
    \begin{center}
    \includegraphics[width=.9\columnwidth]{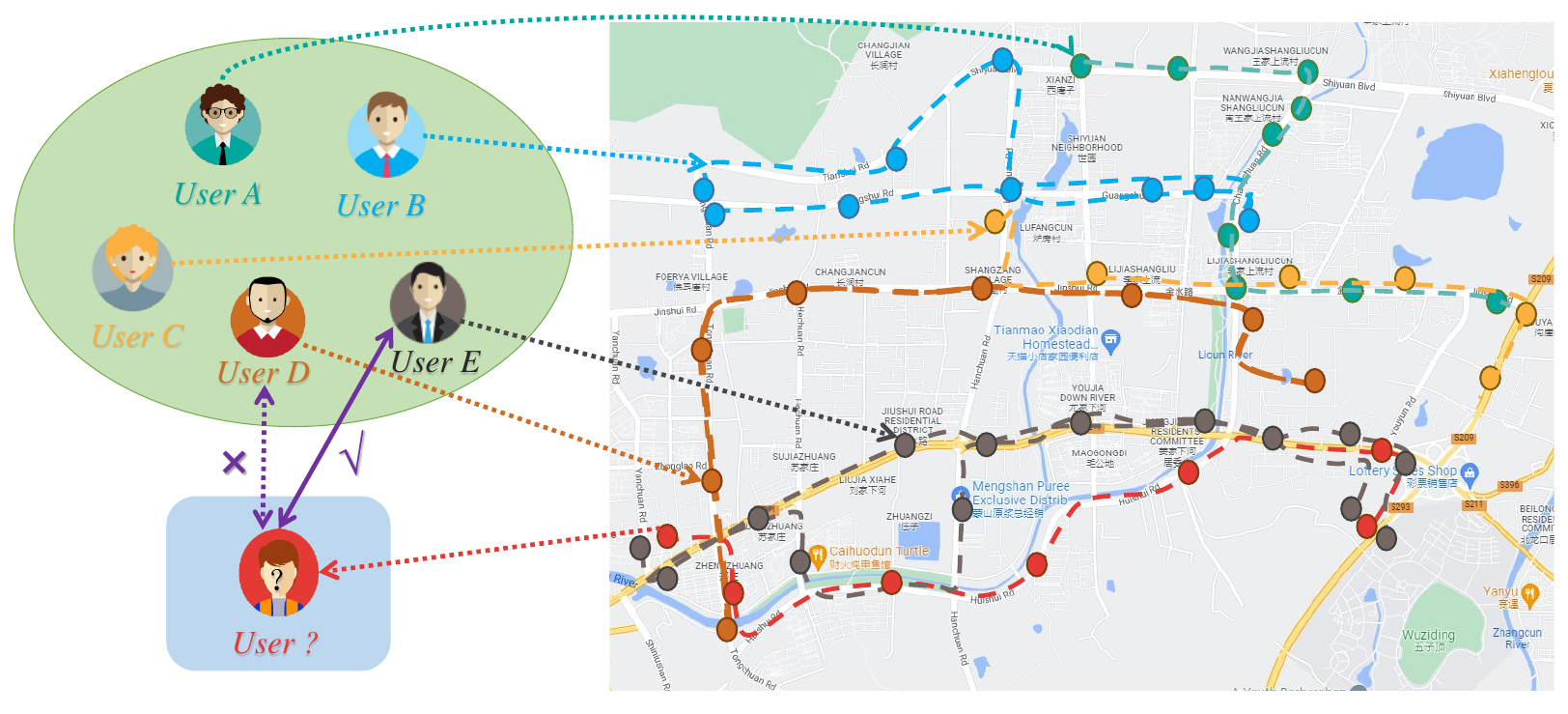}
    \caption{An illustrated example of linking anonymous mobility trajectories with corresponding users. Based on the mobility trajectories of five users, we find that the anonymous trajectory (red) is much closer to the trajectory of user $E$ (dark gray). Thus, we can link the anonymous trajectory with user $E$.}
    \label{fig:example}
    \end{center}
\end{figure}

In this paper, we tackle the trajectory-user linking challenge via exploring the spatio-temporal movement patterns from the mobility data. 
Figure~\ref{fig:example} presents an intuitive example of linking anonymous trajectories with the corresponding users. Our goal in this work is to link anonymous mobility trajectories to the users who are most likely to produce them based on the historical trajectories. 
Due to the strong representation ability of deep learning techniques, existing studies attempt to solve the TUL problem based on various trajectory representation learning models~\cite{miao2020trajectory,zhou2018trajectory,gao2017identifying}. For example, 
TULVAE~\cite{zhou2018trajectory} incorporates VAE model into the TUL problem to learn hierarchical semantics of check-in trajectories with RNN to improve the prediction accuracy. 
DeepTUL~\cite{miao2020trajectory} proposes to integrate recurrent networks with attention mechanism to model higher-order and multi-periodic mobility patterns by learning from historical traces. %to alleviate the data sparsity problem. 
AdattTUL~\cite{gao2020adversarial} and TGAN~\cite{zhou2021improving} utilize Generation Adversarial Network (GAN) to study the TUL problem with adversarial mobility modeling. 
Besides, SML-TUL~\cite{zhou2021self} uses contrastive learning to encode the predictive representations from the user mobility itself constrained by spatio-temporal factors. 
Recently, there has been a surge in the development of graph-based models~\cite{zhou2021trajectory,byun2023aspect,deng2023s2tul} that aim to enhance the acquisition of trajectory representations for TUL problem by adeptly capturing intricate user visitation preferences, discerning patterns of transition between check-in points, and delving into other pertinent contextual factors.

Despite their effectiveness, four key limitations exist in these methods. \textit{First}, all existing methods still suffer from \textit{data sparsity}, and cannot learn quality representations over low-sampling trajectories (\eg, check-ins of inactive users). \textit{Second}, existing methods generally rely on the recurrent neural frameworks to model trajectory sequence, and thus can hardly capture the \textit{long-term dependencies} among the long sequence mobility traces~\cite{khandelwal2018sharp}. Previous approaches show sub-optimal performance when dealing with long sequence data in our evaluation. 
\textit{Third}, most of existing works are limited to inject the global spatial and temporal context into the intra- and inter-trajectory pattern modeling. In practical scenarios, human trajectory are often exhibited with hierarchical mobility patterns across different time granularities~\cite{zhou2018trajectory}. 
\textit{Lastly}, few of previous approaches incorporate the \textit{rich external contextual features} into the representation of human mobility data.

% In this paper, we are interested in linking trajectories to their potential users for spatio-temporal mobility data. 
% Recently, several studies~\citep{miao2020trajectory,zhou2018trajectory,gao2017identifying,gao2020adversarial} have been done to solve the TUL problem via trajectory representation learning.  
% Although representation learning methods successfully solve the high dimensionality in TUL task, existing methods still have four key limitations. 
% \textit{First}, all existing methods still suffer from \textit{data sparsity}, and do not work well for low-sampling trajectories (\eg, check-ins generated by inactive users).   
% \textit{Second}, existing methods generally rely on RNN models, which fail to capture \textit{long-term dependencies} for long sequence trajectory data. 
% In~\citep{khandelwal2018sharp}, it is shown that RNNs have a certain degree of forgetfulness in processing long-sequence data. Previous approaches also show this issue when dealing with long sequence data in the subsequent experiments.  
% \textit{Third}, all existing works focus on the local spatial sequence modeling of trajectory data, while ignoring the \textit{global modeling}. 
% In fact, human trajectories often contain hierarchical structures at different time granularity, implying hierarchical mobility patterns~\citep{zhou2018trajectory}. 
% \textit{Lastly}, previous approaches only utilize spatial feature and/or temporal feature, and fail to utilize \textit{rich contextual features} in mobility data such as motion state.
% Lastly, integrating more diverse contextual features.  

To address the aforementioned challenges, we propose \model, a hierarchical spatio-temporal attention network, to realize high predictability for TUL problem. Particularly, \model\ is built on a graph neural architecture to encode both local and global trajectory transitional patterns. Our graph-based message passing paradigm is able to alleviate the data sparsity issue by effectively performing knowledge transfer among geographical regions and trajectories based on our generated hierarchical spatial graphs. To capture the intra-trajectory transitional regularities, we develop a temporal self-attention mechanism to encode the long-term mobility dependencies. To model the inter-trajectory relationships, \model\ enhances the trajectory representation paradigm by preserving the spatial context across the entire urban space, with the introduced global elastic attentive encoder. The integrated local and global embeddings will be fed into the prediction layer for linking trajectories to their corresponding users. With the design of local and global hierarchical modeling, our proposed \model\ method is general and robust for both dense GPS trajectory data and sparse check-in trajectory data

% It is worth noting that the previous studies basically regards the sparse check-in sequence as a trajectory for TUL task. We are the first to try to design a more general TUL model for both dense GPS trajectory data and sparse check-in trajectory data. Due to the design of local and global hierarchical modeling, our model can effectively handle these two types of trajectories. 

In our evaluation, experimental results on three types of real-life mobility datasets show that our model significantly outperforms several strong baselines (\textbf{9.20}\% ACC@1 gain and \textbf{12.73}\% Macro-F1 gain on average) in TUL task on both sparse and dense trajectory data.  Additionally, our perform ablation study to justify the model design rationale with component-wise effect investigation.

% In \model, we first construct a local spatial graph and a global spatial graph to model micro and macro characteristics of users' trajectories, and use GCN on the two graphs to obtain the initial embeddings for the divided grids and trajectories. 
% Local and global graph modelling also effectively alleviates data sparsity. 
% Then, \model learns the local representation that incorporates sequential dependencies within trajectory and the global representation that captures complex correlations among trajectories respectively via the designed hierarchical spatio-temporal attention network. 
% Finally, a linking layer is designed to fuse the two representations to link trajectories to their corresponding users. 
% Experimental results on three types of real-life mobility datasets show that our model significantly outperforms several strong baselines (\textbf{22.49}\% ACC@1 gain and \textbf{27.43}\% Macro-F1 gain on average) in TUL task on both sparse and dense trajectory data. 

Our contributions can be summarized as follows: 

\begin{itemize}
    \item We propose \model, a hierarchical spatio-temporal attention network model to solve the TUL problem. Our \model simultaneously models local and global spatial and temporal characteristics of users' mobility trajectories over a graph neural architecture.
    
    % \item We design a hierarchical attention network, which is composed of a temporal self-attention encoder to learn the local sequential dependencies within trajectory, and a global elastic attention encoder to capture the complex correlations among trajectories.  
    \item We design a hierarchical spatio-temporal attention network which contains a temporal self-attention encoder to learn the local sequential dependencies within trajectory, and a global elastic attention encoder to capture the complex inter-trajectory dependencies.  
    
    \item We conduct extensive experiments on three types of real-life mobility datasets. Results show that our model significantly outperforms state-of-the-art baselines by 6.04\%$\sim$14.07\% and 6.56\%$\sim$29.54\% improvements in terms of ACC@1 and Macro-F1. 
    
    % To support the reproductivity of our experimental results, we release our source code with anonymous format at: \url{https://anonymous.4open.science/r/AttnTUL}.
\end{itemize}

\section{Related Work}

In literature, measuring the similarity or distance between trajectories which is essential to trajectory pattern mining. 
Similarity measures, \eg, Dynamic Time Warping (DTW) \cite{keogh2000scaling}, Longest Common Sub-Sequence (LCSS) \cite{ying2010mining}, Trajectory-Hausdorff Distance~\cite{Atev2010Hausdorff}, Spatio-Temporal Linear Combine distance~\cite{shang2017trajectory}, and Spatiotemporal Signature~\cite{jin2019moving}, are often used to discover the user similarity from their trajectories. \textit{However, such approaches are artificially designed, and thus only suitable for specific scenario.}  
Recently, deep representation learning has been used for trajectory similarity computation~\cite{li2018deep,yao2019computing,yao2020linear,zhang2020trajectory,yang2021t3s}. \textit{However, these methods focus more on improving the efficiency of trajectory similarity computation.}  
Trajectory classification is another way to understand mobility patterns.  
Existing trajectory classification works focus on labeling trajectories as different motion patterns, such as Driving, Biking and Walking in transportation classification~\cite{zheng2008understanding} and Occupied, Non-occupied and Parked in taxi status inference~\cite{zhu2012inferring}. 
These approaches mainly rely on extraction of spatio-temporal characteristics of trajectories.

TUL problem was recently introduced in~\cite{gao2017identifying}, which links trajectories to their generating-users, and gradually becomes a hot topic in spatio-temporal data mining by classifying trajectories by users. 
Due to its broad range of applications in personalized recommendation systems, location-based services and urban planning, it gradually becomes a hot topic in spatio-temporal data mining. 
Several methods have been proposed to solve the TUL problem~\cite{gao2017identifying,zhou2018trajectory,miao2020trajectory,yu2020tulsn,sun2021trajectory}.  
TULER~\cite{gao2017identifying} utilizes RNN based models to learn sequential transition patterns from trajectories, and links them to users. 
It first uses word embedding to learn the representations for locations in the trajectories and feed them into RNN model to capture mobility patterns for TUL. 
However, the standard RNN based models suffer from data sparsity problem and lacking of understanding hierarchical semantics of human mobility. 
In their follow-up work~\cite{zhou2018trajectory}, TULVAE is proposed to improve the prediction accuracy by incorporating VAE into TUL task to learn hierarchical semantics of check-in sequences. 
However, it fails to utilize existing abundant features and also do not consider multi-periodic mobility regularities.
DeepTUL~\cite{miao2020trajectory} proposes using the attentive recurrent network in TUL to alleviate the data sparsity problem by leveraging historical data, and capture multi-periodic regularities of human mobility to improve prediction accuracy. 
AdattTUL~\cite{gao2020adversarial} and TGAN~\cite{zhou2021improving} introduce Generation Adversarial Network (GAN) to deal with the TUL problem. 
Recently, SML-TUL~\cite{zhou2021self} uses contrastive learning to learn the predictive representations from the user mobility itself constrained by the spatio-temporal factors. After that, MainTUL~\cite{chen2022MainTUL} attempts to alleviate the trajectory user linking problem on sparse check-in data sets by introducing distillation technology and using historical trajectory information.
\textit{Nevertheless, these methods use RNNs for modeling or prediction, which cannot effectively model long-term dependencies of trajectories, and all above methods ignore the contribution of global spatial modeling to TUL prediction.} 
% \textit{However, recurrent network used in DeepTUL cannot effectively model long-term dependencies of trajectories, and all above methods ignore the contribution of global spatial modeling to TUL prediction.} 

Given the multifaceted nature of the data within Location-Based Social Networks (LBSN), an array of scholarly endeavors~\cite{zhou2021trajectory,byun2023aspect,deng2023s2tul} pertaining to TUL problem have undertaken the exploration of user preferences through the discernment of influences stemming from diverse sources, including user relationships, temporal factors, geographical factors, and so on. Such endeavors are geared towards mitigating challenges such as data sparsity problem. Notably, the incorporation of graph-based models [10, 15], which inherently excel in capturing intricate and diverse relationships, has recently been introduced into the realm of TUL problem, further enriching the methodological landscape. GNNTUL~\cite{zhou2021trajectory} first propose a novel end-to-end model, composed of a graph neural network (GNN) module and a classifier, to learn human mobility and associate the traces to the users effectively and efficiently. After that, S2TUL~\cite{deng2023s2tul} captures more complex movement relationships by constructing different homogeneous and heterogeneous graphs, and integrates them into a unified semi-supervised framework for the TUL problem. Contemporaneously, ANES~\cite{byun2023aspect} introduces a aspect-oriented network embedding for social link inference, a novel approach that leverages user trajectory data and bipartite graphs to learn aspect-oriented relations between users and Points-of-Interest(POI). \textit{Still and all, the above methods only emphasize partial relational information, neglecting the integration and synergy of local and global information. In this study, we initiated a comprehensive effort to carefully cover all possible factors, constructing corresponding graph action modules for different granularity trajectory information for the first time, and unifying them in one framework.}

In addition, a few recent works are proposed to identify users across different mobility datasets~\cite{feng2019dplink,Feng2020dplink}. 
DPLink~\cite{feng2019dplink} is designed to model the correlations between trajectories to measure the trajectory similarity from different data source. 
% Although DPLink can be applied into TUL problem, xxx. 
% Recently, deep representation learning has been used for spatial-temporal data mining~\cite{}. 
Recently, attention mechanism~\cite{bahdanau2015neural} has also been studied in spatio-temporal data mining. Researchers combine attention with RNN for mobility modeling, such as mobility prediction~\cite{gao2019predicting} and mobility inference~\cite{zhou2019context}, and personalized route recommendation~\cite{wang2019empowering}. 
Thanks to the characteristics of attention mechanism, it can make up for the deficiency of RNN in capturing long-term dependencies to some extent. 
\textit{Different from the previous studies, to our best knowledge, we are the first to abandon RNN and adopt the fully attention neural network to solve TUL problem.}

\section{Preliminaries}

In this section, we first introduce some preliminary concepts and then formally define the problem of TUL. 

Let $\mathcal{U}=\{u_1,u_2,\dots,u_i\}$ denote a set of moving users.

\begin{Def}[Spatio-Temporal Point] A spatio-temporal point is a uniquely entity in the form of $\langle t, \ell \rangle$, where $t$ is the visited timestamp, and $\ell$ represents the geographical coordinates of the point (\ie, longitude and latitude). 
\end{Def}

% Notice that the location granularity of spatio-temporal points is different in different mobility data. For example, in GPS data, location $\ell$ is expressed in latitude and longitude, while location $\ell$ may be expressed as coarse grained check-in record (POI) in LBSN data. 

\begin{Def}[Trajectory] A trajectory is a sequence of spatio-temporal points $(\langle t_1, \ell_1 \rangle, \langle t_2, \ell_2 \rangle, \dots, \langle t_m, \ell_m \rangle)$ generated by user $u_i$ in chronological order during a certain time interval $\tau$, which is denoted by $Tr_{u_i}^{\tau}$.  
\end{Def}

The time interval can be one hour, one day, one week or even one month. 
A trajectory is called \textit{unlinked}, if we do not know the user who generated it. 
Let $T$ denote the whole time interval. 
The known trajectories generated in previous time intervals by all users, \ie, linked trajectories, are denoted by $Tr_{u}=\{Tr_{u_i}^{\tau}|u_i \in \mathcal{U} \land \tau \in T\}$.

Based on the above definitions, we now state our studied problem as below: 

\begin{Pro*}[Trajectory-User Linking]
Given a set of unlinked trajectories $\overline{Tr}$ generated by users $\mathcal{U}$ and corresponding linked trajectories $Tr_u$, our goal is to learn a mapping function $f: \overline{Tr} \rightarrow \mathcal{U}$ that links anonymous trajectories to users.  
\end{Pro*}

% The TUL problem is formalized as follows: Assume that we are given a number of unlinked trajectories $\overline{L}=\{l_1,l_2,..,l_m\}$ generated by some of the users in the set $U=\{u_1,u_2,..,u_n\}$. A trajectory $l$ for which it is not known who was the user generating it, is called unlinked ($l\in\overline{L}$). our goal is provide a mapping function $f:\overline{L} \mapsto U$  that links the unlinked trajectory to users. 

Key notations used in the paper are summarized in Table~\ref{tab:notation}. 

\begin{table}[h]
  \begin{center}
  \caption{Main notations and their definitions.}
  \label{tab:notation}
    % \fontsize{9}{12} \selectfont
    \begin{tabular}{c|c}
    \toprule
    Notation & Definition\\ \midrule
    $\mathcal{U}$ & the set of moving users\\
    $\langle t,\ell \rangle$ & a spatio-temporal point\\
    $Tr_i$ & a trajectory\\
    $m$ & the number of spatio-temporal points in $Tr_i$\\
    $\tau$ & the time interval\\
    $Tr_{u_i}^{\tau}$ & a trajectory generated by $u_i$\\
    $T$ & the whole time interval\\
    $Tr_{u}$ & the set of linked trajectories\\
    $\overline{Tr}$ & the set of unlinked trajectories\\
    $\mathcal{G}_l=(\mathcal{V}_l, \mathcal{E}_l)$ & the local spatial graph\\
    $\mathbf{A}_l$ & the adjacency matrix of local spatial graph\\
    $\mathbf{X}_{l}$ & the feature matrix of local spatial graph\\
    $\mathcal{G}_g=(\mathcal{V}_g, \mathcal{E}_g)$ & the global spatial graph\\
    $\mathbf{A}_g$ & the adjacency matrix of global spatial graph\\
    $\mathbf{X}_{g}$ & the feature matrix of global spatial graph\\
    $\mathbf{H}_l$ & the embeddings of all grids\\
    $\mathbf{H}_g$ & the global embeddings of all trajectories\\
    $z_i^l$ & the local representation for $Tr_i$\\
    $z_i^g$ & the global representation for $Tr_i$\\     
    \bottomrule
    \end{tabular}
  \end{center}
\end{table}

% \begin{table}[htbp]
% \centering
% \vspace{-2mm}
% \caption{Main notations and their definitions.}
% \label{tab:notation}
% \vspace{-2mm}
% \fontsize{9}{10} \selectfont
% \setlength{\tabcolsep}{2.8mm}{}	
% \begin{tabular}{c|c}
%      \toprule
%      Notation & Definition \\
%      \midrule
%      $Tr_{in}$			& the input trajectory\\
%      $Tr_{au}$			& the augmented long-term trajectory\\
%      $p_i, c_i, t_i$ &  POI, POI category and time of a check-in\\
%      $\mathbf{X}_{in}^{p}$ &  embeddings of POI sequence for $Tr_{in}$\\
%      $\mathbf{X}_{in}^{c}$ &  embeddings of category sequence for $Tr_{in}$\\
%      $\mathbf{X}_{au}^{p}$ &  embeddings of POI sequence for $Tr_{au}$\\
%      $\mathbf{X}_{au}^{c}$ &  embeddings of category sequence for $Tr_{au}$\\ 
%      $f_{\theta}$	&  RNN trajectory encoder\\
%      $f_{\phi}$		&  temporal-aware transformer trajectory encoder\\
%      $z_{in}^{\theta}$     &  representation of $Tr_{in}$ obtained by $f_{\theta}$\\
%      $z_{au}^{\phi}$     &  representation of $Tr_{au}$ obtained by $f_{\phi}$\\
%      $z_{in}^{\phi}$     &  representation of $Tr_{in}$ obtained by $f_{\phi}$\\
%      $z_{au}^{\theta}$     &  representation of $Tr_{au}$ obtained by $f_{\theta}$\\
%      \bottomrule
% \end{tabular}
% \vspace{-4mm}
% \end{table}

\section{Methodology}

\begin{figure*}[h]
    \begin{center}
    % \hspace{-8mm}
    % \includegraphics[width=0.999\textwidth]{figure/overview2.png}
    \includegraphics[width=1.02\textwidth]{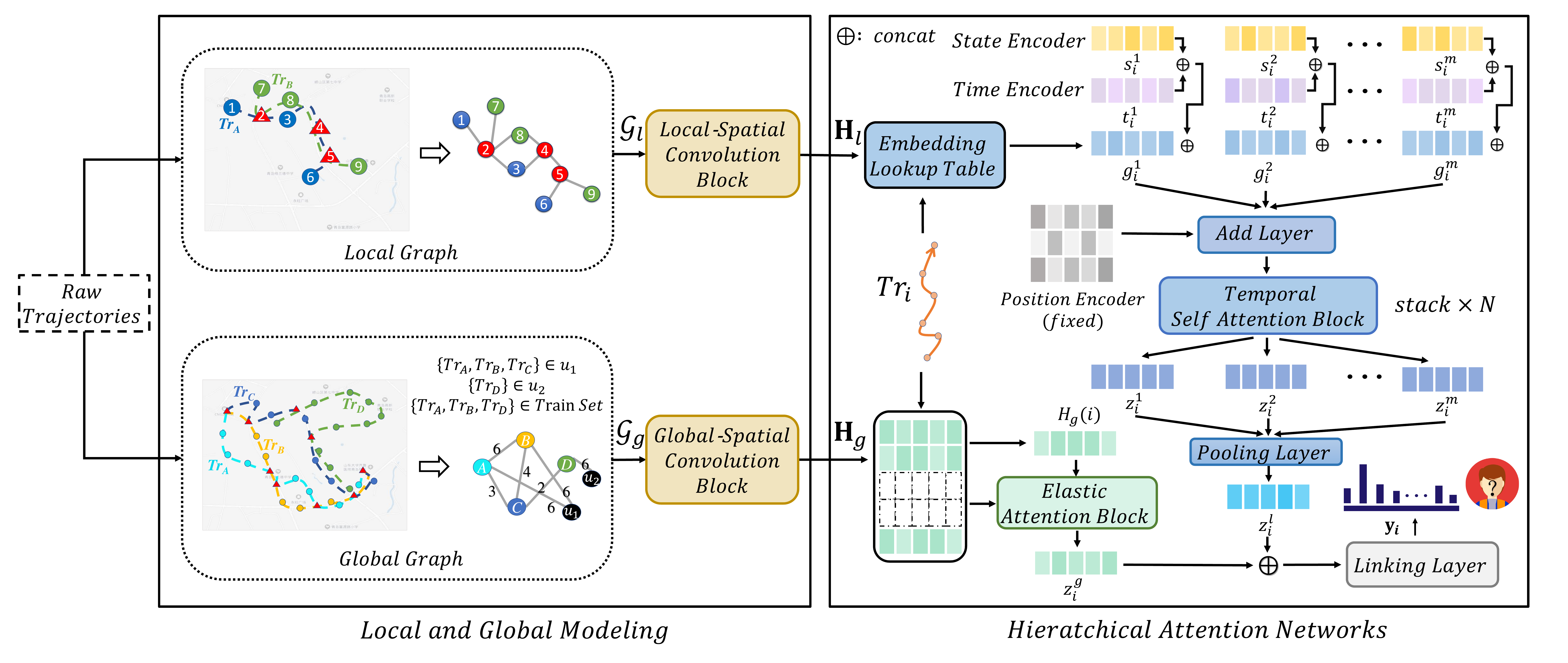}
    % \vspace{-5mm}
    \caption{The overview of the proposed framework}
    % \vspace{-6mm}
    \label{fig:overview}
    \end{center}
\end{figure*}

In this section, we present the details of our neural network model \model (as shown in Figure~\ref{fig:overview}), consisting of four key components: (1) \textit{local and global graph modeling}, (2) \textit{spatial convolutional networks}, (3) \textit{hierarchical spatio-temporal attention networks}, and (4) \textit{linking layer}. 
\textit{First}, we construct a local graph and a global graph to model micro and macro spatial relationships for all trajectories, respectively. 
\textit{Second}, we present the spatial convolutional networks on the two graphs to learn initial embedding for each divided grid and each trajectory. 
\textit{Third}, we employ a multi-head temporal self-attention network to capture the temporal dependency in local embeddings of each trajectory and fuse the local embeddings. We also design an global elastic attention encoder to obtain global representation. 
\textit{Finally}, we use a linking layer to classify trajectories by users. 

\subsection{Local and Global Graph Modeling}

In this section, we introduce how to construct local and global spatial graphs. 

\subsubsection{Preprocessing}

To facilitate the construction of local spatial graph, we employ a gridding technique to divide the geographic space of all trajectories into discrete grids. 
More specifically, given grid size $s$, the whole geographic space is divided into $n$ grids. Consequently, we constrict a grid mapping function $f_g: \ell_i \rightarrow g_i$, so that 
each trajectory $Tr = (\langle t_1, \ell_1 \rangle, \langle t_2, \ell_2 \rangle, \dots, \langle t_m, \ell_m \rangle)$ can be mapped to a grid sequence $Tr^{'} = (\langle t_1, g_1 \rangle, \langle t_2, g_2 \rangle, \dots, \langle t_m, g_m \rangle)$.

\subsubsection{Local Spatial Graph Construction}

To capture the correlations among spatial locations in all trajectories, we first construct a local spatial graph $\mathcal{G}_l=(\mathcal{V}_l, \mathcal{E}_l)$, where each grid is a node in $\mathcal{G}_l$, and edges indicate the connectivity between grids. That is, an edge is created between grid $g_i$ and grid $g_j$ if a trajectory contains a consecutive snippet from $g_i\backslash g_j$ to $g_j\backslash g_i$. 
In high-sampling trajectories, some consecutive spatio-temporal points may be mapped to same grid. For this case, we remove the self-loop edges in $\mathcal{G}_l$. 
The weight on edge $e_{i,j} \in \mathcal{E}_l$ is defined as the number of trajectories that contains the consecutive snippet $(g_i\backslash g_j \rightarrow g_j\backslash g_i)$. 

Furthermore, the semantic context of nodes can be expressed by a feature matrix. In this paper, we adopt a one-hot vector (the length is the number of grids) for each node $g_i$ to express the uniqueness of its location in the feature matrix. 
We use $\mathbf{A}_l$ and $\mathbf{X}_{l}$ to denote the adjacency matrix and feature matrix of local spatial graph, respectively.

\subsubsection{Global Spatial Graph Construction}

To model spatial correlations between trajectories and known linkages between users and trajectories, we next construct a global spatial graph $\mathcal{G}_g=(\mathcal{V}_g, \mathcal{E}_g)$, which is a heterogeneous graph that contains the above two relationships. 

Specifically, $\mathcal{G}_g$ includes two types of nodes -- trajectory nodes and user nodes, \ie, each trajectory is treated as a node and each user is also regarded as a node. 
If two trajectories share common grids, then an edge is created between these two trajectory nodes. The weight of the edge between two trajectory nodes is defined as the number of shared grids in two trajectories. 
In addition, the labeled trajectories are connected to their users, and the weights on the edges between user node and trajectory node are uniformly defined as the maximum weight between trajectory nodes, which indicates that the relationship strength between labeled trajectories and their users is maximized, and further the relationship between the trajectories of the same user is also strengthened.

For each trajectory node $Tr_i$ and user node $u_j$, we encode its spatial features (passed grids) into a multi-hot vector (the length is the number of grids) to represent their spatial characteristics. We use $\mathbf{A}_g$ and $\mathbf{X}_{g}$ to denote the adjacency matrix and feature matrix of global spatial graph, respectively.

Notably, when the number of trajectories is very large, if the traversal method is used to search for pairs of interactive trajectories (\ie, share common grids), the construction of the global spatial graph is actually very time-consuming. 
To improve the construction efficiency and reduce the time cost, we propose a fast implementation trick. We first use grid coding to represent each trajectory, that is, use one hot coding, for each grid, if the trajectory contains the grid, then the code is 1, otherwise 0.  Thus we can obtain the grid coding matrix of all trajectories, denoted by $\mathbf{C}_g$. 
Therefore, the adjacency matrix $\mathbf{A}_g$ can be calculated directly by: 
$\mathbf{A}_g=\mathbf{C}_g \cdot {\mathbf{C}_g}^{\mathsf{T}}.$

\subsection{Spatial Graph Convolutional Networks}

\subsubsection{Local Graph Convolution}
GCNs have been widely used in graph representation learning to capture graph topological structure by aggregating neighbor features and have achieved great success~\cite{wu2020comprehensive,hamilton2017inductive}.
% GCNs have been widely used in graph representation learning and have achieved great success. 
We employ GCN on the local spatial graph and global spatial graph constructed by different granularities to learn the hierarchical embeddings of trajectories.

To jointly capture the topological structures and spatial location information among grids, we first perform convolution operation on local graph $G_l$. 
Following~\cite{kipf2017semi}, multi-layer spatial convolution network performs following layer-wise propagation rule:
\begin{equation}
    \label{eq-globalgraph_w2}
    \mathbf{H}_{l}^{(i+1)}=\operatorname{ReLU}\left(\tilde{\mathbf{D}}_{l}^{-\frac{1}{2}} \tilde{\mathbf{A}}_{l} \tilde{\mathbf{D}}_{l}^{-\frac{1}{2}} \mathbf{H}_{l}^{(i)} \mathbf{W}_{l}^{(i)}\right),
\end{equation}
where $\mathbf{W}_{l}^{(i)}$ is a layer-specific trainable weight matrix,  $\tilde{\mathbf{A}}_{l}=\mathbf{A}_{l}+\mathbf{I}$, and $\tilde{\mathbf{D}}_{l_{ii}}=\sum_{j} \tilde{\mathbf{A}}_{l_{ij}}$. $\mathbf{H}_{l}^{(0)} = \mathbf{X}_l$, where $\mathbf{X}_l$ is the feature matrix of $\mathcal{G}_l$. $\mathbf{H}_{l}^{(i)} \in \mathbb{R}^{n \times d}$ is the output of $i$-th layer where $d$ is the embedding dimension, denoting the initial embeddings of all grids.

\subsubsection{Global Graph Convolution}

Although local graph convolution can learn the embeddings for all grids, it may not be able to capture the correlations between trajectories and between users and their produced trajectories from a global perspective. 
Therefore, to learn the embeddings of trajectories and users, we next perform convolution operation on global spatial graph: 
% However, local spatial graph convolution may not be able to capture some deep-level related information between trajectories and between trajectories and users. Specifically, the trajectory representation of spatiotemporal point granularity does not contain important information between trajectory levels. Intuitively, the more similar the attribute feature of trajectories are, the more similar the trajectory is. Therefore, we use $\mathbf{A}_g$ and $\mathbf{X}_g$ as input to perform the convolution operation:
\begin{equation}
    \label{eq-globalgraph_w2}
    \mathbf{H}_{g}^{(i+1)}=\operatorname{ReLU}\left(\tilde{\mathbf{D}}_{g}^{-\frac{1}{2}} \tilde{\mathbf{A}}_{g} \tilde{\mathbf{D}}_{g}^{-\frac{1}{2}} \mathbf{H}_{g}^{(i)} \mathbf{W}_{g}^{(i)}\right),
\end{equation}
where $\mathbf{W}_{g}^{(i)}$ is a layer-specific trainable weight matrix, and $\mathbf{H}_{g}^{(0)} = \mathbf{X}_g$.
We can learn the embedding for each trajectory and each user via global graph convolutional network, which captures the key spatial characteristics of trajectories from the whole.

\subsection{Hierarchical Spatio-Temporal Attention Networks}

\subsubsection{Semantic Location Encoder} 
\begin{figure}[h]
    \begin{center}
    \includegraphics[width=.85\columnwidth]{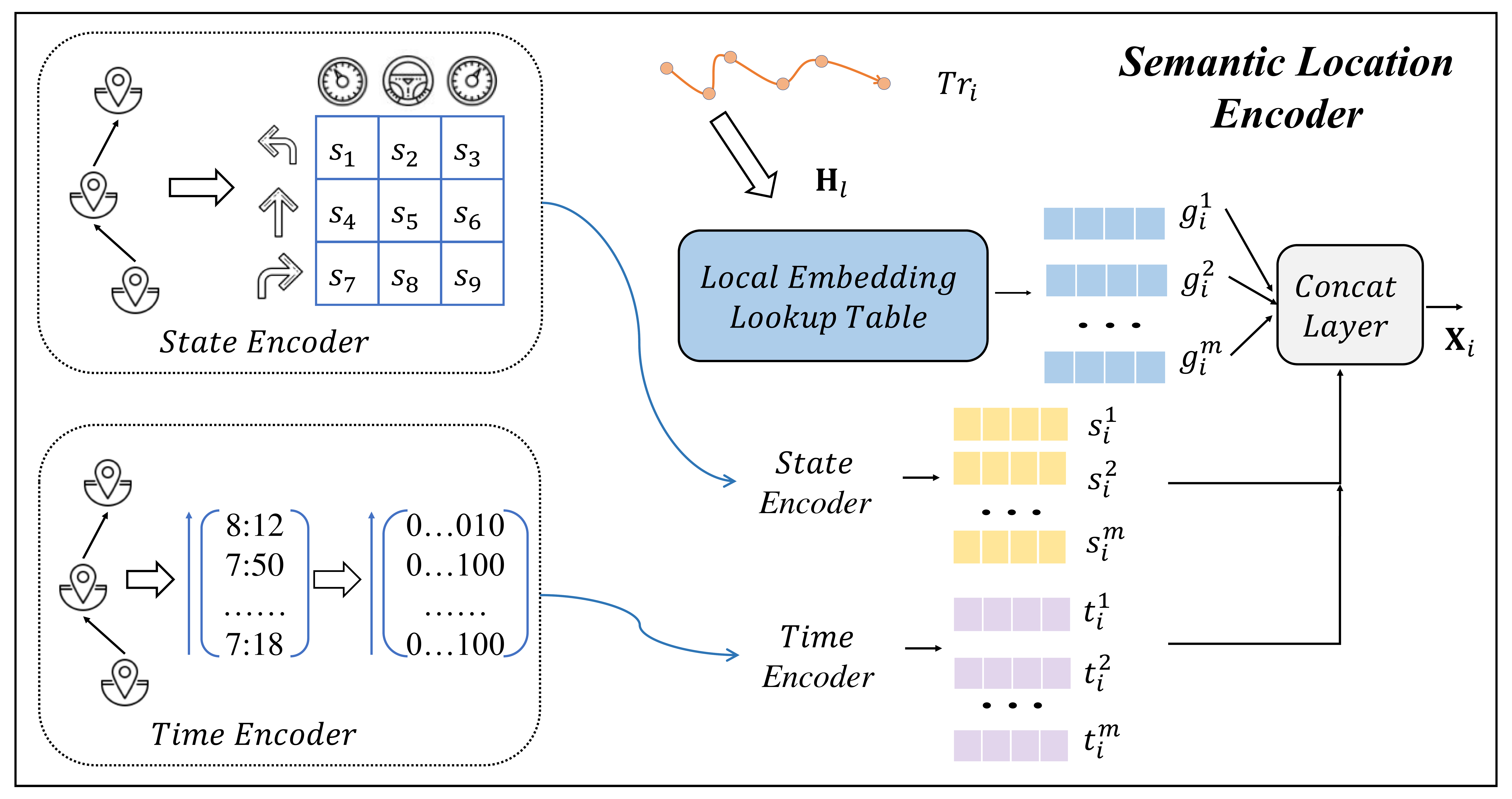}
    \caption{State/time encoder in semantic location encoder}
    \label{fig:state-time}
    \end{center}
\end{figure}

In addition to spatial information, contextual features in mobility data, \eg, motion state and time feature, can be considered into location embedding for trajectory sequence modeling. Therefore, location encoder is a multi-modal embedding module.  
Following~\cite{wang2019spatiotemporal}, we divide the motion state into $(\romannumeral1)$ speed-related operations (\ie, acceleration, deceleration, and constant speed) and $(\romannumeral2)$ direction-related operations (\ie, turning left, turning right, and moving straight), and combine them into nine motion states. 
For time feature, we divide the whole time into time windows according to a certain time granularity (\eg, 10 minutes).  
Figure~\ref{fig:state-time} shows an example of state encoder and time encoder in semantic location encoder. 

Specifically, we design two sparse linear embedding layers to encode motion state $s_i$ and time window $t_i$ (\eg, one-hot), and then concatenate the corresponding grid embedding to obtain an ensemble vector $x_i$. 
The formulation of location encoder is as follows: 
\begin{equation}
    \label{eq-locationemb}
    x_i=\operatorname{Tanh}\left(FC([\mathbf{W}_{t}t_i+b_t;\mathbf{W}_{s}s_i+b_s;\mathbf{H}_{l}(g_i)])\right),
\end{equation}
where $\mathbf{W}_t$, $\mathbf{W}_s$, $b_t$ and $b_s$ are learnable parameters of embedding layers,  $\mathbf{H}_{l}(g_i)$ denotes the local embedding of grid $g_i$, and $\mathbf{[; ;]}$ denotes the concatenate function. $\operatorname{Tanh}(\cdot)$ is the non-linear activation function, and $FC(\cdot)$ is a fully connected layer.

\subsubsection{Temporal Self-Attention Encoder}

To capture the long-term temporal dependencies in trajectory sequence which is difficult to model and learn by RNN-based models, we employ a multi-head temporal self-attention mechanism to learn the intra-trajectory correlations.

To supplement sequence information, the input first adds the position encoding: 
% The degradation problem of neural network is improved through a residual connection layer:
\begin{equation}
\begin{split}
% \label{eq-residual}
    \mathbf{M}_i=\mathbf{X}_i+\mathbf{P},\\
    % \mathbf{M}_{norm}=\mathbf{M}_{i}+\operatorname{norm}(\mathbf{M}_{i}),
\end{split}
\label{eq.4}
\end{equation}
where $\textbf{X}_i = \{x_1,x_2,\dots,x_m\}$ are location embeddings for all grids in trajectory $Tr_i$, and $\mathbf{P}$ denotes the position encoding, which aims to distinguish the sequence position. Notice that $\mathbf{P}$ is fixed and defined as in~\cite{vaswani2017attention}.

Then, $\mathbf{M}_i$ is then sent to the self-attention module to integrate the sequence information: 
\begin{equation}
\begin{split}
    % \label{eq-selfattn}
    \left(\mathbf{Q}_{i}, \mathbf{K}_{i}, \mathbf{V}_{i} \right)^\mathsf{T} =\mathbf{M}_{i} \left(\mathbf{W}^Q, \mathbf{W}^K, \mathbf{W}^{V}\right)^\mathsf{T},\\
    \mathbf{Z}_{i}=\operatorname{Softmax}\left(\frac{\mathbf{Q}_{i} \mathbf{K}_{i}^\mathsf{T}}{\sqrt{d}}\right) \mathbf{V}_{i},\\
\end{split}
\label{eq.5}
\end{equation}
where $\mathbf{W}^Q, \mathbf{W}^K, \mathbf{W}^V \in \mathbb{R}^{d \times d}$ are learnable projection matrices. 
The spatiotemporal correlation is calculated by dot product. 
$\operatorname{Softmax}(\cdot)$ is used to normalize the spatiotemporal dependencies, and the scale $\sqrt{d}$ prevents the saturation led by $\operatorname{Softmax}$ function. 
Besides, we also employ a residual connection to solve the problem of network degradation, followed by layer normalization.

In addition, multiple dependency patterns from location embeddings in a single trajectory can be learned through a multi-head attention mechanism: 
\begin{equation}
\begin{split}
    % \label{eq-multihead}
    \mathbf{Z}_i=FC(\operatorname{concat}(\mathbf{Z}_i^{(1)}, \mathbf{Z}_i^{(2)}, \dots, \mathbf{Z}_i^{(\#head)})),
\end{split}
\label{eq.6}
\end{equation}

By stacking multiple temporal self-attention layers, we can effectively capture the complex temporal dependencies in trajectory sequence.

Furthermore, the updated location embeddings $\mathbf{Z}_i=\{z_{i}^1,\\ z_{i}^2, \ldots, z_{i}^m\}$ of trajectory $Tr_i$ are sent to a pooling layer to extract important location information in the trajectory:
\begin{equation}
    \label{eq.pooling}
    z_i^{l}=\operatorname{Pooling}\left(z_{i}^1, z_{i}^2, \ldots, z_{i}^m\right), 
\end{equation}

For $\operatorname{Pooling}$, we select the Max-Pooling to detect the significant information. 
Consequently, the local representation $z_i^{l} \in \mathbb{R}^{d}$ for trajectory $Tr_i$ is obtained by fusing the local location information.

\subsubsection{Global Elastic Attention Encoder}

Inspired by~\cite{martins2016softmax,peters2019sparse}, we design an elastic attention module to select the most relevant global information. Specifically, we first find the global embedding $\mathbf{H}_g(i)$ for given trajectory $Tr_i$, and the attention score \wrt. trajectory $Tr_j$ is defined as: 
% \begin{equation}
% \begin{split}
%     \label{eq-elasticattn}
%     \mathbf{A}_{score}=\frac{q\mathbf{H}_{g}^\mathsf{T}}{\|q\|\left\|\mathbf{H}_{g}^\mathsf{T}\right\|},\\
%     \mathbf{W}_{g}=\operatorname{Sparsemax}\left(\mathbf{A}_{score}\right),
% \end{split}
% \end{equation}
\begin{equation}
\begin{split}
    \label{eq.elasticattn}
    a_{i,j}^s=\frac{\mathbf{H}_g(i)\mathbf{H}_{g}(j)^\mathsf{T}}{\|\mathbf{H}_g(i)\|\left\|\mathbf{H}_{g}(j)^\mathsf{T}\right\|},\\
    \mathbf{A}_i^s=\{a_{i,1}^s,a_{i,2}^s,\dots,a_{i,\|Tr\|}^s\}, \\
    \mathbf{W}_i^{g}=\operatorname{Sparsemax}\left(\mathbf{A}_i^s\right),
\end{split}
\end{equation}
where $\|Tr\|$ is the number of trajectories, and $\operatorname{Sparsemax}$ is an alternative to $\operatorname{Softmax}$ which tends to yield sparse probability distributions: 
\begin{equation}
    \label{eq.sparsemax}
    \operatorname{Sparsemax}(\boldsymbol{x}) = \underset{\boldsymbol{p} \in \Delta^{d}}{\operatorname{argmin}}\|\boldsymbol{p}-\boldsymbol{x}\|^{2},
\end{equation}
where $\Delta^{d} = \left\{\boldsymbol{p} \in \mathbb{R}^{d}: \boldsymbol{p} \geq 0,\|\boldsymbol{p}\|_{1}=1\right\}$. The predictive distribution $\boldsymbol{p}^{*} = \operatorname{Sparsemax}(\boldsymbol{x})$ is likely to assign exactly zero probability to low-scoring choices, which retains the most important factors. 

Finally, the output of global elastic attention encoder is the weighted sum of the global embedding vectors based on attention weights $\mathbf{W}_i^{g}$: 
\begin{equation}
    \label{eq.weight_sum}
    z_i^{g}=\mathbf{W}_i^{g} \mathbf{H}_{g},
\end{equation}
where $z_i^{g} \in \mathbb{R}^{d}$ is the global representation of trajectory ${Tr}_i$.

\subsection{Linking Layer}

The linking layer aims to obtain the probability distribution of users (labels) from the representations learned by hierarchical spatio-temporal attention networks. 
The linking layer first concatenates $z_i^l$ and $z_i^g$ for each trajectory to obtain a higher-level representation. 
Then, a fully connected layer is used to project the high-level representation into a vector with $|\mathcal{U}|$ dimension. The formulation is as follows:
\begin{equation}
    \label{linking}
    \mathbf{y}_i=\left(W_{c}\left[\begin{array}{c}
    z^{l}_{i} \\
    z^{g}_{i}
    \end{array}\right]+b_{c}\right)
\end{equation}
where $\mathbf{W}_c \in \mathbb{R}^{|\mathcal{U}| \times 2d}$ and $b_c \in \mathbb{R}^{|\mathcal{U}|}$ are learnable weight matrix and bias, and the vector $\mathbf{y}_i$ is the estimated probability of users for trajectory $Tr_i$.

To train our model, we apply cross-entropy as loss function and use back propagation algorithm to optimize our model. 
% The objective function is the negative log-likelihood of the true user label: 
We define our cross-entropy-based loss function as follows: 
\begin{equation}
\label{eq.obj}
    \mathcal{L}(\Theta)=-\frac{1}{\zeta} \sum_{i=1}^{\zeta} c_{i} \log \left(\sigma\left(\mathbf{y}_{i}\right)\right)+\frac{\lambda}{2}\|\Theta\|^{2}
\end{equation}
where $c_i$ is the one-hot ground truth label of trajectory ${Tr}_i$, $\sigma(\cdot)$ is the softmax function, $\zeta$ is the number of training trajectories,  
and $\Theta$ is the set of all trainable parameters. $\lambda$ is an L2 regularization hyperparameter to alleviate overfitting.

\subsection{Algorithm Pseudo-Code}

Algorithm~\ref{alg.FastLANE} shows the pseudo-code of our proposed \model framework guided by the above objective function (\ie, Eq.~\eqref{eq.obj}). 

\begin{algorithm}[h]
\renewcommand{\algorithmicrequire}{\textbf{Input:}}
\renewcommand{\algorithmicensure}{\textbf{Output:}}
\caption{The Learning Process of \model}
\label{alg.FastLANE}
\begin{algorithmic}[1]
\REQUIRE Input local spatial graph $\mathcal{G}_l$, global spatial graph $\mathcal{G}_g$, feature matrix $\mathbf{X}_l$ and $\mathbf{X}_g$, embedding dimension $d$, the number of convolution layers $l_n$
\ENSURE Model parameters $\mathbf{\Theta}$

\WHILE{! convergence}

\STATE Perform graph convolutional networks
\FOR{ $i=1$ to $l_n$}
\STATE $\mathbf{H}_{l}^{(i)} \gets \operatorname{ReLU}\left(\tilde{\mathbf{D}}_{l}^{-\frac{1}{2}} \tilde{\mathbf{A}}_{l} \tilde{\mathbf{D}}_{l}^{-\frac{1}{2}} \mathbf{H}_{l}^{(i-1)} \mathbf{W}_{l}^{(i-1)}\right)$;
\STATE $\mathbf{H}_{g}^{(i)} \gets \operatorname{ReLU}\left(\tilde{\mathbf{D}}_{g}^{-\frac{1}{2}} \tilde{\mathbf{A}}_{g} \tilde{\mathbf{D}}_{g}^{-\frac{1}{2}} \mathbf{H}_{g}^{(i-1)} \mathbf{W}_{g}^{(i-1)}\right)$;
\ENDFOR

\STATE $\mathbf{X} \gets \operatorname{Tanh}\left(FC([\mathbf{W}_{t}t+b_t;\mathbf{W}_{s}s+b_s;\mathbf{H}_{l}])\right)$; 

\STATE Get $z_i^l$ using Eqs.~\eqref{eq.4}-\eqref{eq.pooling};

% \STATE $\mathbf{M} \gets \mathbf{X} + \mathbf{P}$; 

% \FOR{ $i=1$ to $\#head$}
% \STATE $\left(\mathbf{Q}^{(i)}, \mathbf{K}^{(i)}, \mathbf{V}^{(i)} \right)^\mathsf{T} \gets  \mathbf{M} \left(\mathbf{W}^{Q^{(i)}}, \mathbf{W}^{K^{(i)}}, \mathbf{W}^{V^{(i)}}\right)^\mathsf{T}$;
% \STATE $\mathbf{Z}^{(i)} \gets \operatorname{Softmax}\left(\frac{\mathbf{Q}^{(i)} {\mathbf{K}^{(i)}}^\mathsf{T}}{\sqrt{d}}\right) \mathbf{V}^{(i)}$;
% \ENDFOR

% \COMMENT {Multi-head attention for lines 8-9: } 
% \STATE $\left(\mathbf{Q}, \mathbf{K}, \mathbf{V} \right)^\mathsf{T} \gets  \mathbf{M} \left(\mathbf{W}^{Q}, \mathbf{W}^{K}, \mathbf{W}^{V}\right)^\mathsf{T}$;
% \STATE $\mathbf{Z} \gets \operatorname{Softmax}\left(\frac{\mathbf{Q} \mathbf{K}^\mathsf{T}}{\sqrt{d}}\right) \mathbf{V}$;

% \STATE $\mathbf{Z} \gets  FC(\operatorname{concat}(\mathbf{Z}^{(1)}, \mathbf{Z}^{(2)}, \dots, \mathbf{Z}^{(\#head)}))$,

% \STATE $\mathbf{Z}^{l} \gets \operatorname{Pooling}(\mathbf{Z})$;

\STATE Get $z_i^g$ using Eqs.~\eqref{eq.sparsemax}-\eqref{eq.weight_sum};

% \STATE $\mathbf{Z}^{g} \gets \mathbf{W}^{g} \mathbf{H}_{g}$; 

% \STATE $\mathbf{Y} \gets \left(\mathbf{W}_{c}\left[\mathbf{Z}^{l}; \mathbf{Z}^{g}\right]+b_{c}\right)$;

\STATE $\mathbf{y}_i \gets \left(W_{c}\left[{z}_i^{l}; {z}_i^{g}\right]+b_{c}\right)$ 

\STATE Calculate $\mathcal{L}$ using Eq.~\eqref{eq.obj};
\STATE Back propagation and update parameters in \model;
\ENDWHILE
\STATE Return $\mathbf{\Theta}$;
\end{algorithmic}
\end{algorithm}

We now analyze the time complexity of our model. 
The time consumption of our model mainly lies in spatial GCNs, temporal self-attention encoder, and global elastic attention encoder.
First, the time complexity of spatial GCN block is $O(|\mathcal{E}|d)$ and $O(|\mathcal{V}|^2 d)$ for the GCN operation and the spatial attention operation, respectively, for every iteration.  
Second, for each trajectory, the time complexity of temporal self-attention encoder is $O(m^2 d)$ for the multi-head self-attention operation for each layer. 
It is clear that the time complexity of the global elastic attention encoder is much smaller than that of the temporal self-attention encoder. 
Therefore, the time complexity of our model is mainly composed of the spatial GCNs and the temporal self-attention encoder, and also depends on the number of layers of networks, the number of trajectories, and the average length of trajectories. 
\section{Experiments}

\begin{table}[b]
    \vspace{-6mm}
    \begin{center}
    % \caption{Statistics of the datasets. $|\mathcal{U}|$: number of users; $|\mathcal{P}|$: number of GPS points; $|\mathcal{C}|$: number of POIs; $|\mathcal{R}|$: average length of trajectories (before segmentation); $|\mathcal{T}_{r}|$: range of trajectory length}
     \caption{Statistics of the datasets. $\overline{len}$: average length of trajectories, and H: hour(s).}
    \label{tab:dataset}
    \fontsize{9}{12} \selectfont
    \setlength{\tabcolsep}{2mm}{}	
    \begin{tabular}{c|c|c|c|c|c|c}
        \toprule
        % \textbf{Dataset} & $|\mathcal{U}|$  & $|\mathcal{P}|$ & $|\mathcal{C}|$ & $|\mathcal{R}|$ & $|\mathcal{T}_{r}|$ \\
        \textbf{Datasets} & \#users  & \#trajectories & \#points & \#POIs & $\tau$ & $\overline{len}$  \\
        \midrule
        \multirow{2}{*}{Gowalla} & 547 & 38,567 & 15,967 & 15,892 & 6H & 2\\
        \cline{2-7}
        & 222 & 18,808 & 11,979 & 11,960 & 6H & 3\\
        \midrule
        \multirow{2}{*}{PrivateCar} & 71 & 4,178 & 135,085 & 10,493 & 1H & 32\\
        \cline{2-7}
        & 42 & 2,797 & 87,719 & 7,514 & 1H & 31\\
        \midrule
        \multirow{2}{*}{GeoLife} & 90 & 6,035 & 945,971 & 23,369 & 3H & 157\\
        \cline{2-7}
        & 56 & 4,064 & 768,533 & 19,384 & 3H & 189\\
        \bottomrule
    \end{tabular}
    \vspace{-6mm}
    \end{center}
\end{table}

In this section, we evaluate our proposed model on three
types of real-world mobility datasets. The following research
questions (RQs) are used to guide our experiments: 
\begin{itemize}
\item \textbf{RQ1.} How does our \model perform in TUL problem on real-world datasets compared to existing methods? 
\item \textbf{RQ2.} How does each component contribute to the performance of the proposed \model?\
\item \textbf{RQ3.} How does \model perform with different parameter settings (\eg, grid size $s$, dimension $d$, and time window size)?
\item \textbf{RQ4.} Whether the trajectory representation learned by our \model is better than DNN-based baselines? 
\item \textbf{RQ5.} How does our \model perform efficiently compared to existing methods?
\end{itemize}

% \textcolor{red}{RQX. needs to be supplemented here}

\subsection{Datasets} 

We conduct extensive experiments on three publicly available real-world datasets: Gowalla\footnote{\url{http://snap.stanford.edu/data/loc-gowalla.html}}, PrivateCar\footnote{\url{https://github.com/HunanUniversityZhuXiao/PrivateCarTrajectoryData}}, and GeoLife\footnote{\url{https://www.microsoft.com/en-us/research/project/geolife-building-social-networks-using-human-location-history/}}. 

\begin{itemize}
    \item \textbf{Gowalla}~\cite{gao2017identifying} is a user check-in dataset from a location based social network service. 
    For each user, we concatenate all check-in locations to form a trajectory which will be further divided into sub-trajectories based on given time interval.  
    
    \item %PrivateCar~\cite{huang2019road} is a GPS trajectory dataset of private cars collected in Shenzhen, China. It records GPS traces of 977 private cars from January 1 to January 15 in 2016. 
    \textbf{PrivateCar}~\cite{huang2019road} is a GPS trajectory dataset of private cars collected in Shenzhen, China from January 1 to January 15 in 2016. The sampling rate of the private car trajectory dataset is relatively high, 1$\sim$60 seconds (average: 15.3 seconds and standard deviation: 7.8 seconds). 
    
    \item \textbf{GeoLife}~\cite{zheng2010geolife} is a GPS trajectory dataset collected in Beijing from April 2007 to August 2012. This dataset includes a broad range of users’ outdoor movements, including sightseeing, hiking, cycling and so on. 91\% of the trajectories are logged densely, \eg, every 1$\sim$5 seconds or every 5$\sim$10 meters per point.

\end{itemize}

Notice that our model does not requires POI data. Some baselines (\eg, TULER and TULVAE) are designed to work on sparse check-in data. To ensure fair comparison, we map GPS trajectories to POI check-in data to make these baselines can handle GPS trajectory data. 
To fairly reproduce the results of baselines, we crawl POIs in Shenzhen and Beijing from BaiduMap\footnote{\url{https://map.baidu.com/}} as the additional geographical context for PrivateCar and GeoLife datasets. 

In order to check the robustness of our model, we select two different number of users from each dataset. 
The statistics of these datasets are summarized in Table~\ref{tab:dataset}. Following~\cite{gao2017identifying}, $\tau$ is set as a reasonable time interval on different datasets (\ie, 6 hours, 1 hour and 3 hours for Gowalla, PrivateCar and GeoLife datasets, respectively).

\subsection{Baselines}

In our evaluation, we compare our \model against the following two categories of baselines.

\subsubsection{Classic models}
\begin{itemize}
    \item \textbf{LCSS}~\cite{ying2010mining} -- It adopts LCSS to compute trajectory similarity, and searches the most similar trajectory in training set to find the corresponding user. 
    % \item \textbf{LCSS}~\cite{ying2010mining}: The Longest Common Sub Sequence is a common trajectory similarity query algorithm. We search the most similar trajectory in the train set to find the user corresponding to the trajectory of the test set, and apply LCSS to solve the TUL problem. 
    
    \item \textbf{LDA}~\cite{shahdoosti2017spectral} -- We apply Linear Discriminant Analysis (LDA) to TUL task by embedding trajectories into one-hot vectors and using SVD to decompose the within-class scatter matrix. 
    % The Linear Discriminant Analysis is a popular spatial data classification method. We apply LDA to TUL task by embedding trajectory into one-hot vectors and using SVD to decompose the within-class scatter matrix.
    
    \item \textbf{DT}~\cite{jiang2018survey} -- Decision Tree (DT) is a classic classification method for trajectory data. We use entropy as the criterion in TUL problem, which shows better performance than Gini index.
    
    \item \textbf{SR}~\cite{jin2019moving,jin2020trajectory} -- Signature Representation (SR) is a state-of-the-art trajectory similarity measure for moving object linking. 
    
\end{itemize}

\subsubsection{Deep neural network models}
\begin{itemize}
    \item \textbf{TULER}~\cite{gao2017identifying} -- This is the original RNN model for solving TUL task. There are three variants: RNN with Gated Recurrent Unit (\textbf{TULER-G}), Long Short-Term Memory (\textbf{TULER-L}) and bidirectional LSTM (\textbf{Bi-TULER}).
    
    \item \textbf{TULVAE}~\cite{zhou2018trajectory} -- It utilizes VAE to learn the hierarchical semantics of trajectory with stochastic latent variables that span hidden states in RNN. 
    
    \item \textbf{DeepTUL}~\cite{miao2020trajectory} -- This is a recurrent network with attention mechanism to solve TUL problem, which is a state-of-the-art method. It learns from labeled historical trajectory to capture multi-periodic nature of user mobility and alleviate the data sparsity problem. 
    
    \item \textbf{DPLink}~\cite{feng2019dplink,Feng2020dplink} -- DPLink is a state-of-the-art method to link user accounts from heterogeneous mobility data. %It cannot be directly applied to TUL problem, thus we extend it to perform user identification in a single mobility dataset in our experiments. 
    
    \item \textbf{T3S}~\cite{yang2021t3s} -- T3S is a state-of-the-art trajectory representation learning method for trajectory similarity computation. 

    \item \textbf{GNNTUL}~\cite{zhou2021trajectory} -- GNNTUL is the first GNN-based human mobility learning model exploiting implicit transition patterns behind sparse user traces in online social networks while extracting users’ unique movement features and discriminating the different trajectories.
\end{itemize}

Notice that we extend T3S with multi-class classification supervision to support TUL prediction. 
DPLink also cannot be directly applied to TUL problem, thus we extend it to perform user identification in a single mobility dataset in our experiments. The source code of our model is available at \url{https://github.com/Onedean/AttnTUL}.

\subsection{Evaluation Metrics}

We use the widely used $ACC@k$, Macro-P, Macro-R and Macro-F1 to evaluate the performance, which are common metrics in multi-classification task. 
Specifically, $ACC@K$ is used to evaluate the accuracy of TUL prediction as: 
\begin{equation}
    \label{acc@k}
    ACC@K=\frac{|\{Tr_{i} \in \overline{Tr} : u^{*}(Tr_{i}) \in \mathcal{U}_{K}(Tr_{i})\}|}{|\overline{Tr}|},
\end{equation}
where $u^{*}(Tr_{i})$ is the ground truth user, and $\mathcal{U}_{K}(Tr_{i})$ is the predicted top-$K$ user set. 
It is considered correct if the ground truth user $u^{*}(Tr_{i})$ lies within the predicted top-k user set $\mathcal{U}_{K}(Tr_{i})$. 

Macro-F1 is regarded as an overall performance indicator, taking into account the precision and recall across all classes in multi-classification task, which is defined as: 
\begin{equation}
\begin{split}
    \label{macrof1}
  Macro\text{--}P  =  \frac{1}{|c|}\sum^{|c|}_{i=1}{P_i}, \\
  Macro\text{--}R  =  \frac{1}{|c|}\sum^{|c|}_{i=1}{R_i}, \\
  Macro\text{--}F1  = \frac{1}{|c|}\sum^{|c|}_{i=1}{\frac{2 \times P_{i} \times R_{i}}{P_{i} + R_{i}}}
\end{split}
\end{equation}
where $|c|$ is number of classes, and $P_{i}$ and $R_{i}$ are \textit{precision} and \textit{recall} of each class (user in TUL) respectively.

\subsection{Experimental Settings}

In our experiments, we use the first 60\% of each user's trajectories as the training set on all datasets, the following 20\% as the validation set, and the remaining 20\% as test set.

We use the source code released by authors for baselines, and use the parameter setting recommended in the original paper and fine-tune them on each dataset to be optimal. 
% In the experiment, we set $d$ to 128, grid size $s$ to 120 meters, the number of GCN layers to 2, and the number of self-attention layers to 3 for our \model. For all learning methods, we set epoch to 80, batch size to 16 and tune learning rate in \{0.05, 0.001, 0.005, 0.0001, 0.0005\}. The batch size is set to 16, and drop out rate is set to 0.5.For fair comparison, for all learning methods, we set epoch to 80, batch size to 16 and tune learning rate in \{0.05, 0.001, 0.005, 0.0001, 0.0005\}. use early stopping mechanism, and set patience to 10 to avoid over fitting. 
In the experiment, we set embedding dimension $d$ to 128, grid size $s$ to 40 meters on Gowalla and 120 meters on the other two datasets, the length of time window to 2 hours, 10 minutes and 20 minutes on Gowalla, PrivateCar and GeoLife, the number of GCN layers to 2, the number of self-attention layers to 3, $\#head$ to 4, and $\lambda$ to $5e$-4 for our \model. 
For fair comparison, for all learning methods, we set epoch to 80, batch size to 16, dropout to 0.5, tune learning rate from 0.0001 to 0.01, use early stopping mechanism, and set patience to 10 to avoid over fitting. 
In each experiment, we repeat 10 runs and report the average.

All experiments for model efficiency evaluation are conducted on a machine with Intel Xeon Gold 6126@2.60GHz 12 cores CPU, 192GB memory, and NVIDIA Tesla V100-SXM2 (16GB) GPU.

% \textcolor{red}{Privacy issues description.} 
Notice that all the user personal information in our datasets are anonymized to protect user’s privacy. 
% Meanwhile, we store the data in a secure local server and only core researchers can access the data with strict non-disclosure agreements

\begin{table*}[t]
% \vspace{-6mm}
\centering
\caption{Performance comparison with deep neural network models on three real-world datasets.}
% \vspace{-3mm}
\label{tab:perform_compared}
% \fontsize{9}{12} \selectfont

\setlength{\tabcolsep}{0.8mm}{}	
\begin{tabular}{c|c|c|c|c|c|c|c|c|c|c|c}

\toprule

\multirow{2}{*}{Data} & \multirow{2}{*}{Methods} & ACC@1 & ACC@5 & Macro-P & Macro-R & Macro-F1 & ACC@1 & ACC@5 & Macro-P & Macro-R & Macro-F1\\

\cline{3-12}

& & \multicolumn{5}{c|}{$|\mathcal{U}|=222$} & \multicolumn{5}{c}{$|\mathcal{U}|=547$}\\

\midrule

\multirow{8}{*}{\rotatebox{90}{Gowalla}} &TULER-L &41.67\%  &51.23\%  &43.45\%  &34.11\%  &36.03\%  &37.61\%  &46.97\%  &40.96\%  &32.47\%  &34.24\%\\

\cline{2-12}

&TULER-G &41.56\%  &50.96\%  &41.06\%  &33.08\%  &34.64\%  &38.88\%  &48.61\%  &42.20\%  &33.69\%  &35.15\%\\

\cline{2-12}

&Bi-TULER &41.19\%  &50.36\%  &41.88\%  &33.92\%  &35.83\%  &36.88\%  &46.85\%  &41.09\%  &32.20\%  &33.80\%\\

\cline{2-12}

&TULVAE &40.13\%  &49.72\%  &39.71\%  &32.01\%  &33.68\%  &37.18\%  &46.44\%  &40.39\%  &32.30\%  &33.67\%\\

\cline{2-12}

&DeepTUL &42.36\%  &51.87\%  &43.81\%  &35.32\%  &\underline{37.22}\%  &37.99\%  &48.16\%  &41.15\%  &32.96\%  &34.48\%\\

\cline{2-12}

&DPLink &41.51\%  &\underline{52.94}\%  &41.32\%  &35.36\%  &36.15\%  &37.19\%  &49.72\%  &36.11\%  &33.18\%  &32.65\%\\

\cline{2-12}

&T3S &40.80\%  &50.04\%  &\underline{44.52}\%  &33.68\%  &35.77\%  &34.53\%  &45.97\%  &35.58\%  &30.76\%  &30.50\%\\

\cline{2-12}

&GNNTUL &\underline{42.61}\%  &52.45\%  &41.41\%  &\underline{36.27}\%  &35.82\%  &\underline{41.28}\%  &\underline{50.74}\%  &\underline{42.63}\%  &\underline{34.90}\%  &\underline{36.72}\%\\

\cline{2-12}

& \textbf{\model} &\textbf{46.17}\%  &\textbf{55.44}\%  &\textbf{45.92}\%  &\textbf{39.54}\%  &\textbf{40.03}\%  &\textbf{44.90}\%  &\textbf{54.72}\%  &\textbf{45.40}\%  &\textbf{38.44}\%  &\textbf{39.63}\%\\

\midrule

\multirow{8}{*}{\rotatebox{90}{PrivateCar}} & &\multicolumn{5}{c|}{$|\mathcal{U}|=42$} &\multicolumn{5}{c}{$|\mathcal{U}|=71$}\\

\cline{2-12}

&TULER-L &21.92\%  &44.90\%  &19.41\%  &21.05\%  &18.42\%  &15.43\%  &32.12\%  &15.20\%  &12.47\%  &12.16\%\\

\cline{2-12}

&TULER-G &21.92\%  &45.94\%  &19.35\%  &19.61\%  &18.01\%  &16.13\%  &32.77\%  &15.96\%  &14.37\%  &13.29\%\\

\cline{2-12}

&Bi-TULER &22.46\%  &47.79\%  &21.52\%  &21.54\%  &19.76\%  &16.57\%  &36.62\%  &15.76\%  &15.14\%  &13.80\%\\

\cline{2-12}

&TULVAE &23.98\%  &50.44\%  &25.34\%  &20.34\%  &20.45\%  &16.89\%  &31.81\%  &16.92\%  &15.38\%  &14.59\%\\

\cline{2-12}

&DeepTUL &22.99\%  &49.71\%  &22.95\%  &22.53\%  &20.16\%  &17.65\%  &33.09\%  &15.68\%  &16.97\%  &14.24\%\\

\cline{2-12}

&DPLink &23.47\%  &47.12\%  &25.36\%  &23.32\%  &22.23\%  &20.71\%  &41.79\%  &19.14\%  &16.83\%  &15.27\%\\

\cline{2-12}

&T3S &26.53\%  &47.90\%  &\underline{31.30}\%  &25.07\%  &25.59\%  &21.28\%  &39.45\%  &24.13\%  &20.05\%  &20.01\%\\

\cline{2-12}

&GNNTUL &\underline{32.84}\%  &\underline{56.89}\%  &30.94\%  &\underline{28.58}\%  &\underline{27.71}\%  &\underline{29.47}  &\underline{47.80}\%  &\underline{28.78}\%  &\underline{30.89}\%  &\underline{28.35}\%\\

\cline{2-12}

&\textbf{\model} &\textbf{35.11}\%  &\textbf{60.40}\%  &\textbf{33.24}\%  &\textbf{32.80}\%  &\textbf{31.49}\%  &\textbf{31.25}\%  &\textbf{54.74}\%  &\textbf{32.25}\%  &\textbf{32.24}\%  &\textbf{30.21}\%\\

\midrule

\multirow{8}{*}{\rotatebox{90}{GeoLife}} & &\multicolumn{5}{c|}{$|\mathcal{U}|=56$} &\multicolumn{5}{c}{$|\mathcal{U}|=90$}\\

\cline{2-12}

&TULER-L &41.79\%  &71.81\%  &33.78\%  &34.94\%  &31.70\%  &36.84\%  &60.28\%  &32.91\%  &30.41\%  &29.32\%\\

\cline{2-12}

&TULER-G &43.93\%  &70.08\%  &37.09\%  &36.50\%  &33.33\%  &35.51\%  &61.24\%  &33.88\%  &31.73\%  &30.25\%\\

\cline{2-12}

&Bi-TULER &44.50\%  &74.09\%  &38.14\%  &35.82\%  &33.76\%  &37.99\%  &61.85\%  &35.82\%  &33.16\%  &32.12\%\\

\cline{2-12}

&TULVAE &46.04\%  &70.99\%  &42.32\%  &39.73\%  &36.87\%  &39.27\%  &64.30\%  &36.23\%  &33.83\%  &32.31\%\\

\cline{2-12}

&DeepTUL &51.23\%  &79.19\%  &45.84\%  &41.82\%  &39.36\%  &44.92\%  &69.93\%  &38.19\%  &38.35\%  &35.99\%\\

\cline{2-12}

&DPLink &\underline{53.80}\%  &\underline{80.03}\%  &\underline{48.23}\%  &\underline{46.63}\%  &\underline{45.03}\%  &47.83\%  &73.95\%  &43.93\%  &40.80\%  &39.63\%\\

\cline{2-12}

&T3S &51.10\%  &77.40\%  &45.47\%  &43.84\%  &43.13\%  &44.33\%  &70.65\%  &40.18\%  &39.52\%  &38.18\%\\

\cline{2-12}

&GNNTUL &52.78\%  &76.13\%  &46.08\%  &45.14\%  &44.23\%  &\underline{48.56}\%  &\underline{75.86}\%  &\underline{46.74}\%  &\underline{43.89}\%  &\underline{43.40}\%\\

\cline{2-12}

&\textbf{\model} &\textbf{61.37}\%  &\textbf{85.87}\%  &\textbf{59.20}\%  &\textbf{59.59}\%  &\textbf{58.53}\%  &\textbf{53.92}\%  &\textbf{79.47}\%  &\textbf{49.94}\%  &\textbf{50.29}\%  &\textbf{48.24}\%\\

\bottomrule

\end{tabular}
% \vspace{-4mm}
\end{table*}

\begin{figure}[h]
\centering
% \hspace{-3mm}
\includegraphics[width=0.92\textwidth]{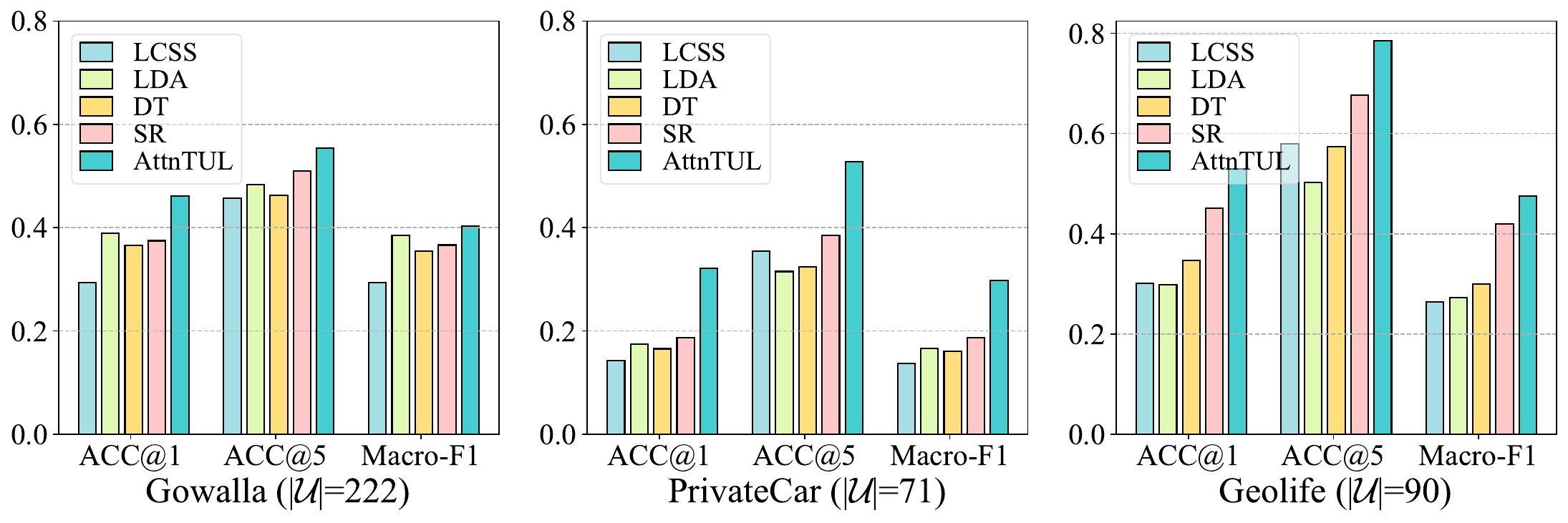}
\caption{Performance comparison with classic models.}
\label{fig:traditonal}
\end{figure}

\subsection{Experiment Results}

\subsubsection{Overall Performance (RQ1)}

We first evaluate the overall performance of the proposed model on three categories of trajectory datasets in comparison with state-of-the-art baselines. 
The overall performance is reported in Table~\ref{tab:perform_compared}, where the best is shown in \textbf{bold} and the second best is shown as \underline{underlined}. 
We also show the comparison with classic models in Figure~\ref{fig:traditonal}. 

Notice that SML-TUL~\cite{zhou2021self} and TGAN~\cite{zhou2021improving} are not compared in our experiments because there are no publicly available source codes for them. However, according to the results reported in~\cite{zhou2021self}, our \model significantly outperforms SML-TUL and TGAN in terms of Acc@$k$ and Macro-F1 on Gowalla, even though our \model links more users on the same Gowalla dataset. The specific results are as follows: 
\uwave{\model} vs. SML-TUL vs. TGAN: \uwave{46.17\%} vs. 45.71\% vs. 43.79\% in ACC@1, and \uwave{40.03\%} vs 36.15\% vs. 33.06\% in Macro-F1 on Gowalla. 

From the results in Figure~\ref{fig:traditonal}, our \model significantly outperforms all classic baseline methods in terms of all evaluation metrics, achieving 23.22\%, 67.02\%, 17.44\% improvements in terms of ACC@1 over SR method on Gowalla, PrivateCar, and Geolife datasets, respectively. 
SR performs better than other classic baselines in terms of ACC@1, ACC@5, and Macro-F1 on three datasets as it considers the similarity of two trajectories from four representation strategies (\ie, sequential, temporal ,spatial and spatiotemporal). However, SR is still based on the defined distance calculation, and cannot automatically learn the similarity or dependency between trajectories based on the relationship among trajectories on different mobility datasets. 

From the results in Table~\ref{tab:perform_compared}, we can see that \model achieves the best performance in terms of all metrics on three different categories of trajectory datasets. 
This is because our designed local and global graph model effectively captures both micro- and macro-spatial features for spatio-temporal trajectories. In addition, our hierarchical spatio-temporal attention network can learn the long-term dependencies in time dimension and adaptively fuse the local and global representations.  
This is main reason that our \model performs the best on both sparse check-in dataset (\ie, Gowalla) and dense GPS trajectory datasets (\ie, PrivateCar and GeoLife). 
Results in Table~\ref{tab:perform_compared} show that our model significantly outperforms state-of-the-art baselines by 6.04\%$\sim$14.07\% and 6.56\%$\sim$29.54\% improvements in terms of ACC@1 and Macro-F1. Specifically, \model achieves average gains of 9.20\% ACC@1 and 12.73\% Macro-F1 score in comparison to the best performed baseline across all datasets.
Considering that the performance gain in TUL prediction reported in recent works~\cite{miao2020trajectory,zhou2018trajectory} is usually around 2.74-9.26\% ACC@1 and 1.11-11.00\% Macro-F1, this performance improvement achieved by our \model is significant. 
Among various baselines, GNNTUL performs the second best on most of metrics, because they extract comprehensive trajectory features. However, our \model consistently outperforms GNNTUL by 6.04-16.28\% ACC@1 and 6.56-31.88\% Macro-F1 score across all datasets, which suggests the rationality of our designed hierarchical spatio-temporal attention network architecture.
Although GNNTUL uses graph-based representation models, it is still limited by the only consideration of the information within the trajectory, and do not consider the complex cross-trajectory correlations. 
% Another key difference is that our model is built over a fully attention neural model by addressing the limitation of non-parallel propagation in RNNs.

We also observe that model performance on data with fewer users is better than that on data with more users. This is intuitive as the more users the more difficult the classification becomes. Among all baselines, GNNTUL perform best in terms of ACC@1 and Macro-F1 on both Gowalla and PrivateCar datasets, and DPLink performs the best in terms of ACC@1 and Macro-F1 on GeoLife dataset. 
However, our \model achieves more improvement over the best performed baseline in terms of ACC@1 and Macro-F1 on data with more users, \eg, on Gowalla, 8.77\% improvement in ACC@1 for $|\mathcal{U}|=547$ vs. 8.35\% improvement for $|\mathcal{U}|=222$, and 7.92\% improvement in Macro-F1 for $|\mathcal{U}|=574$ vs. 7.55\% improvement for $|\mathcal{U}|=222$.  

\begin{table}[t]
% \vspace{-7mm}
\centering
\caption{Results of ablation study. Mac: Macro.}
\label{tab:ablation}
% \vspace{-2mm}
% \fontsize{9}{12} \selectfont
% \hspace{-6mm}
\setlength{\tabcolsep}{2mm}{}
\begin{tabular}{c|c|c|c|c|c|c}
\toprule

Datasets & Methods  & ACC@1 & ACC@5 & Macro-P & Macro-R & Macro-F1\\

\midrule

\multirow{6}{*}{\makecell{Gowalla\\ ($|\mathcal{U}|$=222)}} &\modell &43.66\%  &53.28\%  &45.14\%  &37.95\%  &38.56\%\\ 

\cline{2-7}

&\modelg &41.87\%  &52.61\%  &37.03\%  &33.31\%  &33.29\%\\

\cline{2-7}

& \modelsa &43.20\%  &52.70\%  &\underline{45.82}\%  &36.58\%  &37.99\%\\ 

\cline{2-7}

&\modelea &39.99\%  &51.58\%  &37.12\%  &31.73\%  &32.08\%\\

\cline{2-7}

&\modelts &\underline{44.01}\%  &\underline{54.13}\%  &45.47\%  &\underline{38.72}\%  &\underline{39.92}\%\\

\cline{2-7}

&\model &\textbf{46.17}\%  &\textbf{55.44}\%  &\textbf{45.92}\%  &\textbf{39.54}\%  &\textbf{40.03}\%\\

\midrule

\multirow{6}{*}{\makecell{PrivateCar\\ ($|\mathcal{U}|$=71)}} &\modell &19.83\%  &44.74\%  &18.26\%  &20.86\%  &16.01\%\\ 

\cline{2-7}

&\modelg &30.09\%  &53.14\%  &27.50\%  &31.05\%  &27.16\%\\

\cline{2-7}

&\modelsa &28.31\%  &\underline{53.92}\%  &27.40\%  &27.67\%  &24.97\%\\

\cline{2-7}

&\modelea &28.49\%  &52.12\%  &\underline{28.36}\%  &24.54\%  &23.64\%\\

\cline{2-7}

&\modelts &\underline{30.58}\%  &53.20\%  &26.96\%  &\underline{31.52}\%  &\underline{27.41}\%\\

\cline{2-7}

&\model &\textbf{31.25}\%  &\textbf{54.74}\%  &\textbf{32.25}\%  &\textbf{32.24}\%  &\textbf{30.21}\%\\

\midrule

\multirow{6}{*}{\makecell{GeoLife\\ ($|\mathcal{U}|$=90)}} &\modell &49.51\%  &78.43\%  &45.49\%  &45.48\%  &43.04\%\\

\cline{2-7}

&\modelg &48.97\%  &75.46\%  &43.82\%  &43.61\%  &42.03\%\\

\cline{2-7}

&\modelsa &51.06\%  &78.65\%  &47.01\%  &47.15\%  &44.01\%\\

\cline{2-7}

&\modelea &46.38\%  &74.35\%  &41.98\%  &40.56\%  &38.59\%\\

\cline{2-7}

&\modelts &\underline{52.07}\%  &\underline{79.25}\%  &\underline{47.58}\%  &\underline{47.41}\%  &\underline{45.02}\%\\

\cline{2-7}

&\model &\textbf{53.92}\%  &\textbf{79.47}\%  &\textbf{49.94}\%  &\textbf{50.29}\%  &\textbf{48.24}\%\\

\bottomrule

\end{tabular}
% \vspace{-1mm}
\end{table}

\subsubsection{Ablation Study (RQ2)}

To validate the effectiveness of each component in \model, we further conduct the ablation study. We compare our \model with five carefully designed variants. Despite the changed part(s), all variants have the same framework structure and parameter settings.
% (1) \textbf{\modell} -- removes the whole local spatial module; (2) \textbf{\modelg} -- removes the whole global spatial module; (3) \textbf{\modelsa} -- removes the self attention module; (4) \textbf{\modelea} -- uses $\operatorname{Softmax}$ to replace the elastic attention in global module; (5) \textbf{\modelts} -- removes the time and state encoders. 
\begin{itemize}
    \item \textbf{\modell} -- It removes the whole local spatial module to demonstrate the importance of local modeling.
    
    \item \textbf{\modelg} -- It removes the whole global spatial module to demonstrate the importance of global modeling.
    
    \item \textbf{\modelsa} -- It removes the self attention module and directly sends the location embeddings to pooling layer to fuse local information.
    
    \item \textbf{\modelea} -- It uses $\operatorname{Softmax}$ to replace our proposed elastic attention in global module.
    
    \item \textbf{\modelts} -- In this variant, we remove the time and state encoders to verify the importance of them. 
\end{itemize}

The results of ablation study on three datasets are shown in Table~\ref{tab:ablation}. 
As we can see, \model outperforms all variants on three datasets, which demonstrates that our key components all contribute to performance improvement of \model. 
A noteworthy phenomenon is that \modelts performs the second best on three datasets. A potential reason is that time information is not as important as spatial information in TUL problem. Similar conclusion has also been verified in~\cite{jin2019moving,jin2020trajectory}. 

The comparison between \modell and \model highlights the effectiveness of the proposed local modeling for capturing intra-trajectory dependencies in TUL problem. 
The comparison between \modelg and \model reflects the effectiveness of the proposed global modeling for extracting inter-trajectory corrections in TUL problem. 
Compared to \model, \modell and \modelg perform much worse demonstrating the importance of both local and global modelling in our model. 

In addition, we can see that \modelsa performs much worse than \model in terms of ACC@1 on PrivateCar and GeoLife, indicating that our designed temporal self-attention network could better capture long-term dependencies for long sequence trajectories. Nevertheless, \modelsa also performs worse than \model on Gowalla datasets, which demonstrates that the self-attention mechanism can also capture temporal dependencies for sparse check-in trajectories. 
Furthermore, we also find an important observation that \modelea produces worse results than \modelg in terms of most metrics on three datasets. This adequately demonstrates the crucial role of our designed elastic attention of global module in extracting relevant global representation for TUL prediction.

\begin{figure*}
\centering
% \hspace{-3mm}
\includegraphics[width=1.0\textwidth]{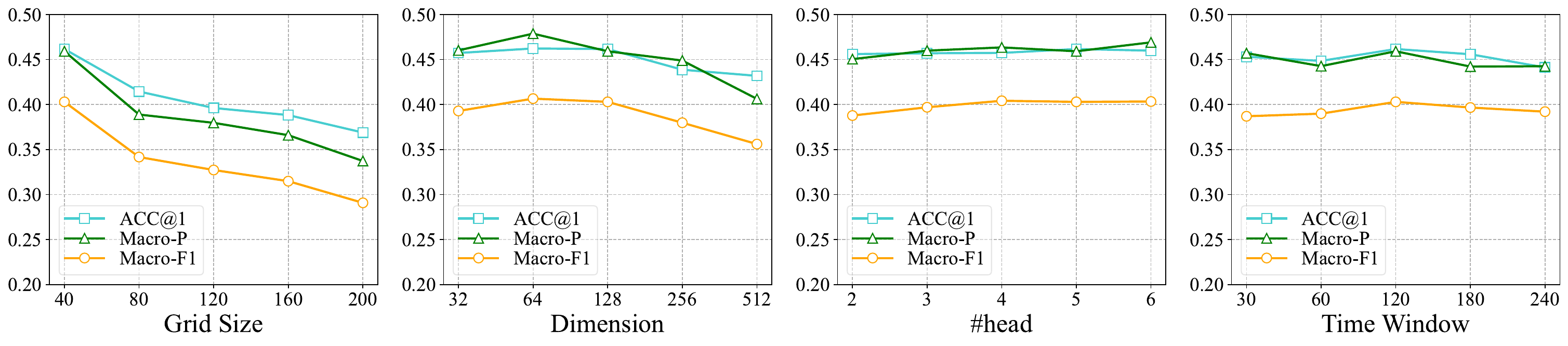}\\
(a) Gowalla\\
% \hspace{-3mm}
\includegraphics[width=1.0\textwidth]{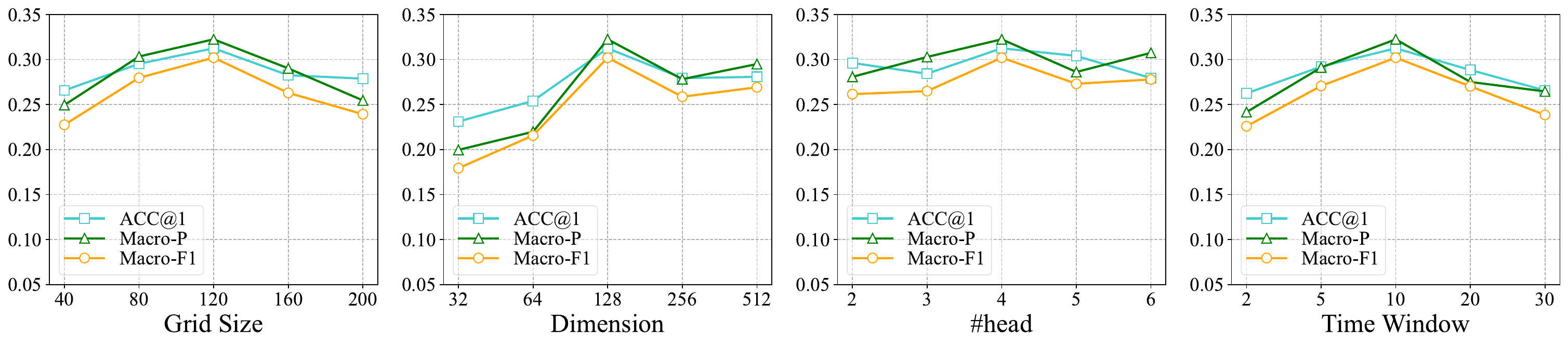}\\
(b) PrivateCar\\
\caption{Parameter sensitivity \wrt. grid size $s$, embedding dimensionality $d$, the number of attentive representation heads $\#head$ and time window length $T_W$ on Gowalla and PrivateCar data. Performances are measured by ACC@1, Macro-P, Macro-F1.}
\label{fig-parm}
\end{figure*}

\subsubsection{Robustness Analysis (RQ3)}

We also evaluate the robustness of \model \wrt. different settings of grid size $s$, embedding dimension $d$, the number of heads $\#head$ and the length of time window $T_W$. 
% We conduct the experiments on both sparse Gowalla and dense PrivateCar datasets. 
The results in terms of ACC@1, Macro-P and Macro-F1 on both sparse Gowalla ($|\mathcal{U}|=222$) and dense PrivateCar ($|\mathcal{U}|=71$) are depicted in Figure~\ref{fig-parm}.

\textit{The effect of the grid size.} 
From two left sub-figures in Figure~\ref{fig-parm}, we find that the performance of \model decreases as $s$ increases on Gowalla, while first increases and then decreases with the growth of $s$ from 40 to 200 meters. 
This phenomenon is easy to understand. Recall that Gowalla is a sparse check-in data, while PrivateCar is a dense GPS trajectory data. 
If grid is too large, the spatial information would be lost, which is reflected more obvious on sparse Gowalla than on dense PrivateCar. If grid is too small, some spatial interactions among trajectories would not be reflected, which affects the prediction performance. 

\textit{The effect of the embedding dimension.}
As shown in the two sub-figures of the second group, %the performance of \model has little effect \wrt. $d$ on Gowalla, and first increases and then decreases on PrivateCar as $d$ increases, reaching the best at $d=128$.  
the performance of \model first increases and then decreases on both Gowalla and PrivateCar as $d$ increases, reaching the best at $d=64$ and $d=128$ on Gowalla and PrivateCar, respectively. 
This is because Gowalla is very sparse, and thus few grids are involved. The lower embedding dimensionality has little effect on it, while PrivateCar is denser. When the dimensionality is very low, it has a greater impact on the prediction performance. 
However, larger embedding dimensionality may contain superfluous information which hurts the TUL prediction performance.

\begin{figure*}[b]
    \centering
    \hspace{-5mm}
    \subfigure[TULER-L]{
    \includegraphics[width=0.28\linewidth]{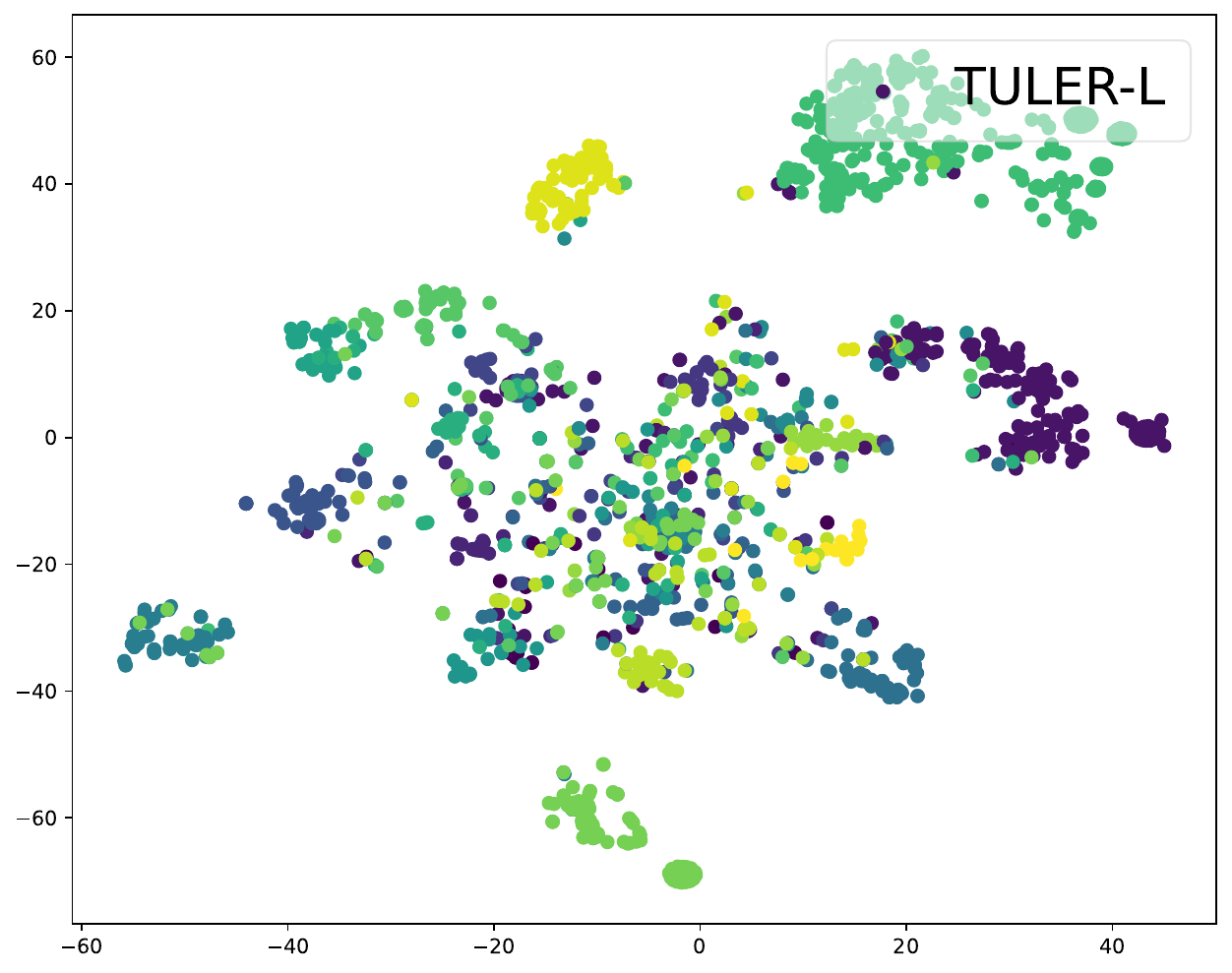}
    }
    \hspace{-3mm}
    \subfigure[TULER-G]{
    \includegraphics[width=0.28\linewidth]{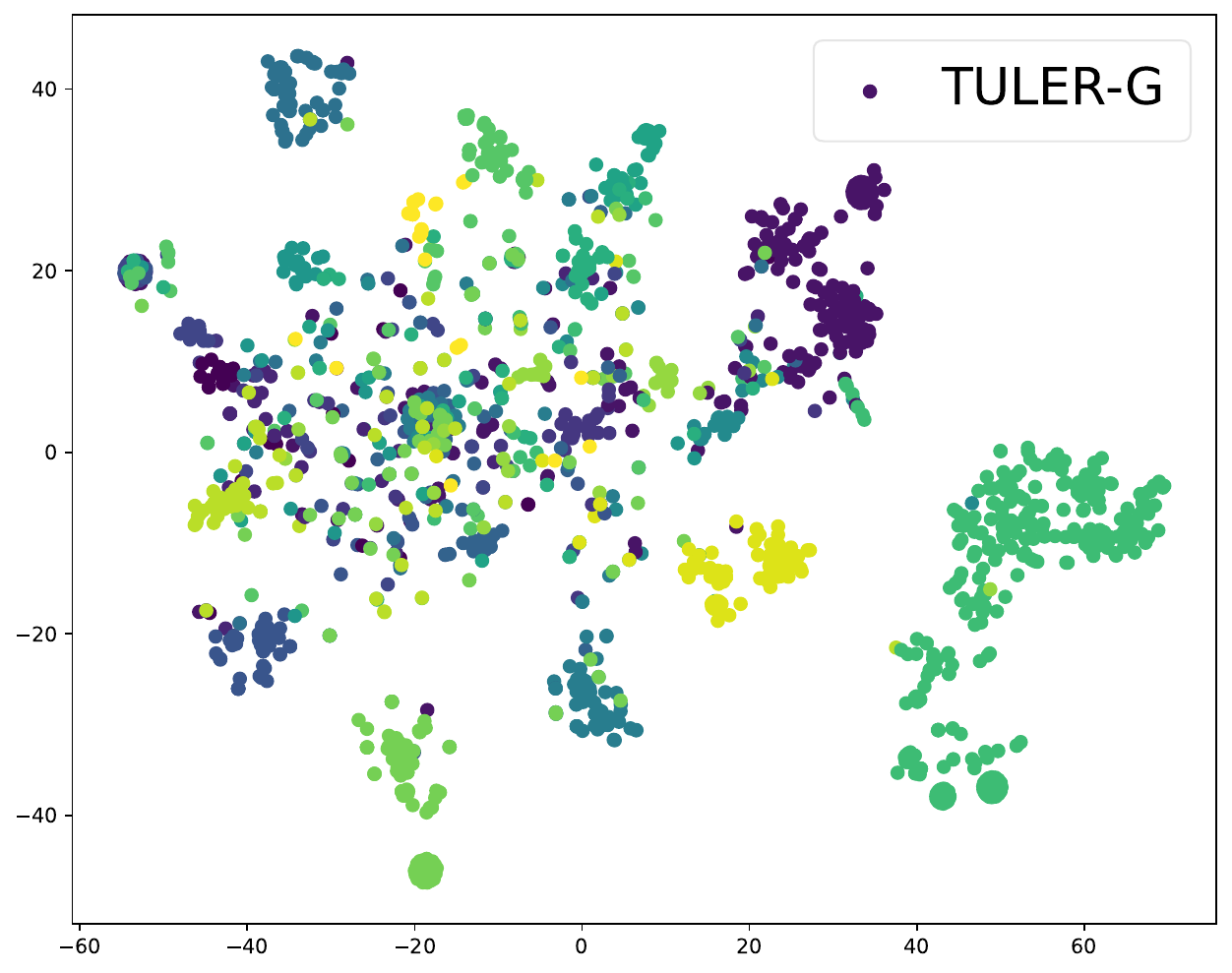}
    }
    \hspace{-3mm}
    \subfigure[Bi-TULER]{
    \includegraphics[width=0.28\linewidth]{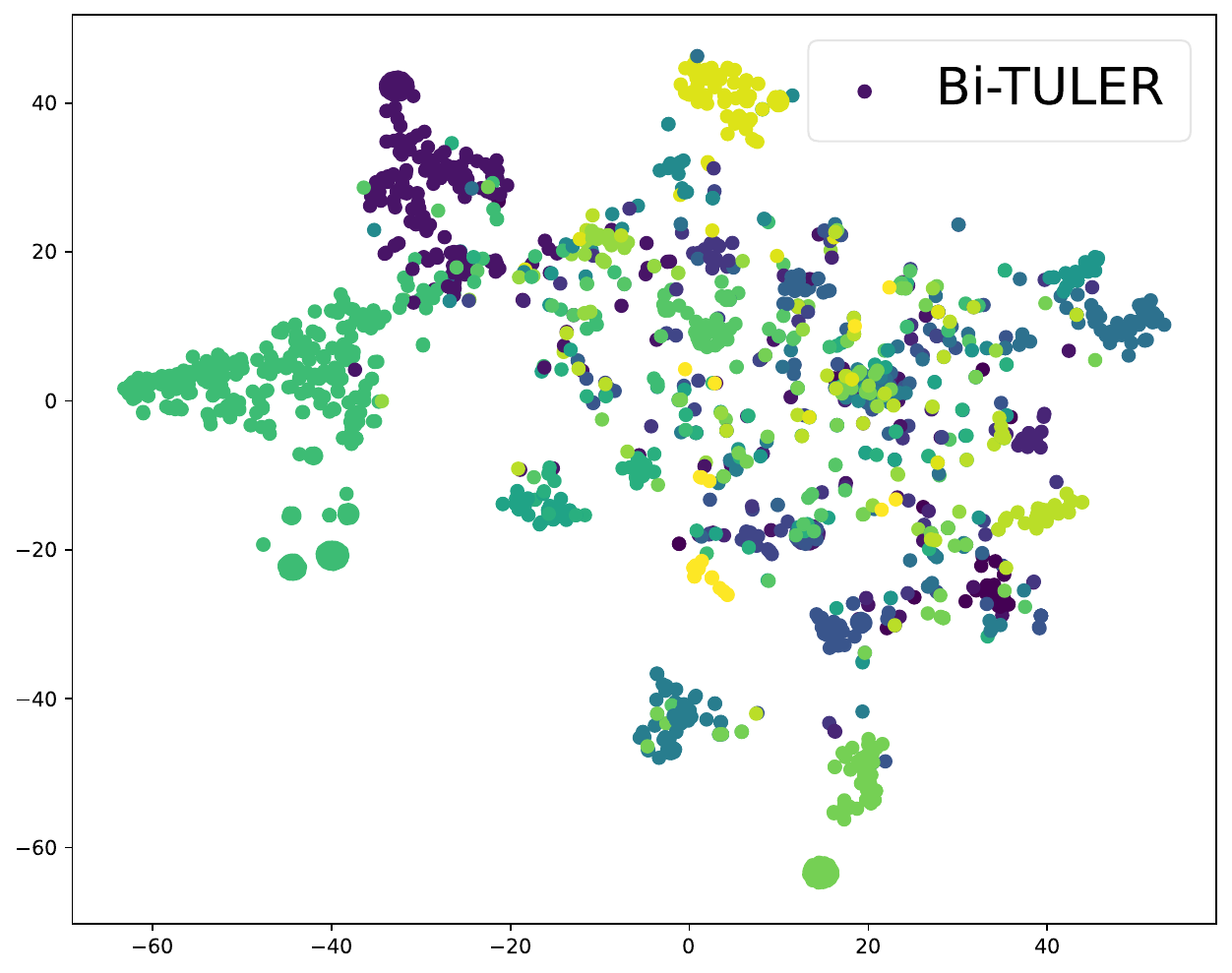}
    }\\
     \hspace{-5mm}
     \subfigure[TULVAE]{
    \includegraphics[width=0.28\linewidth]{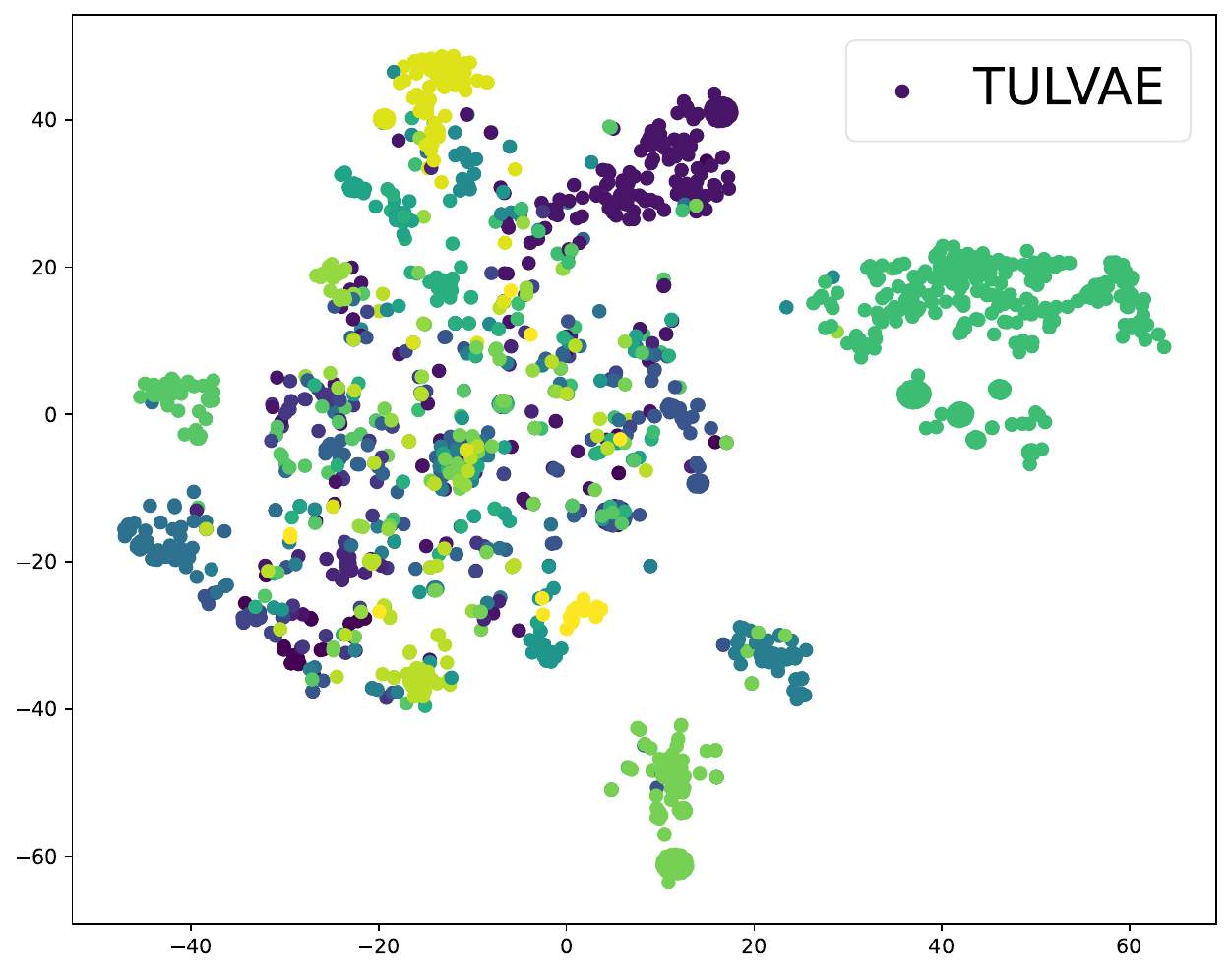}
    }
    \hspace{-3mm}
    \subfigure[DPLink]{
    \includegraphics[width=0.28\linewidth]{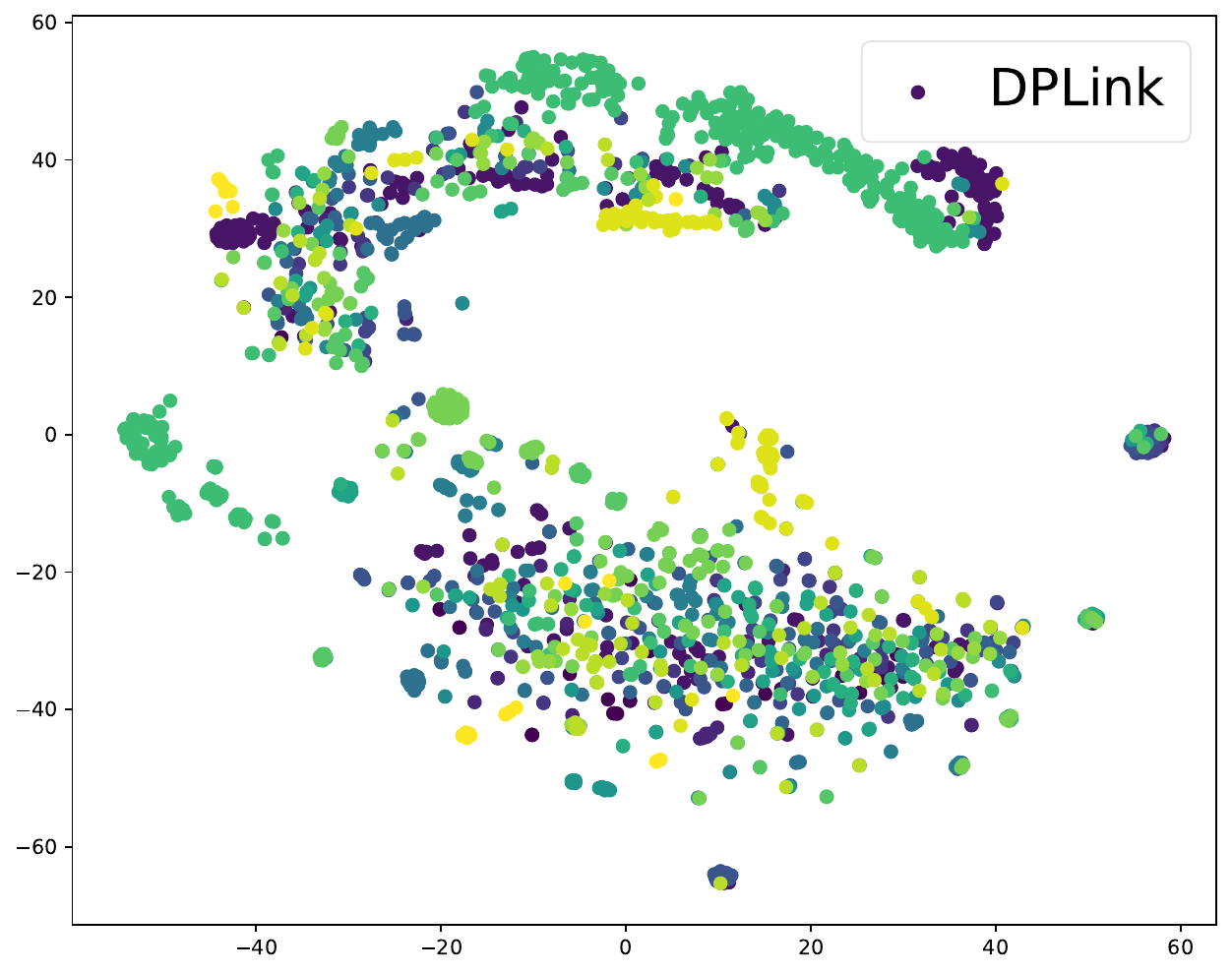}
    }
    \hspace{-3mm}
    \subfigure[T3S]{
    \includegraphics[width=0.28\linewidth]{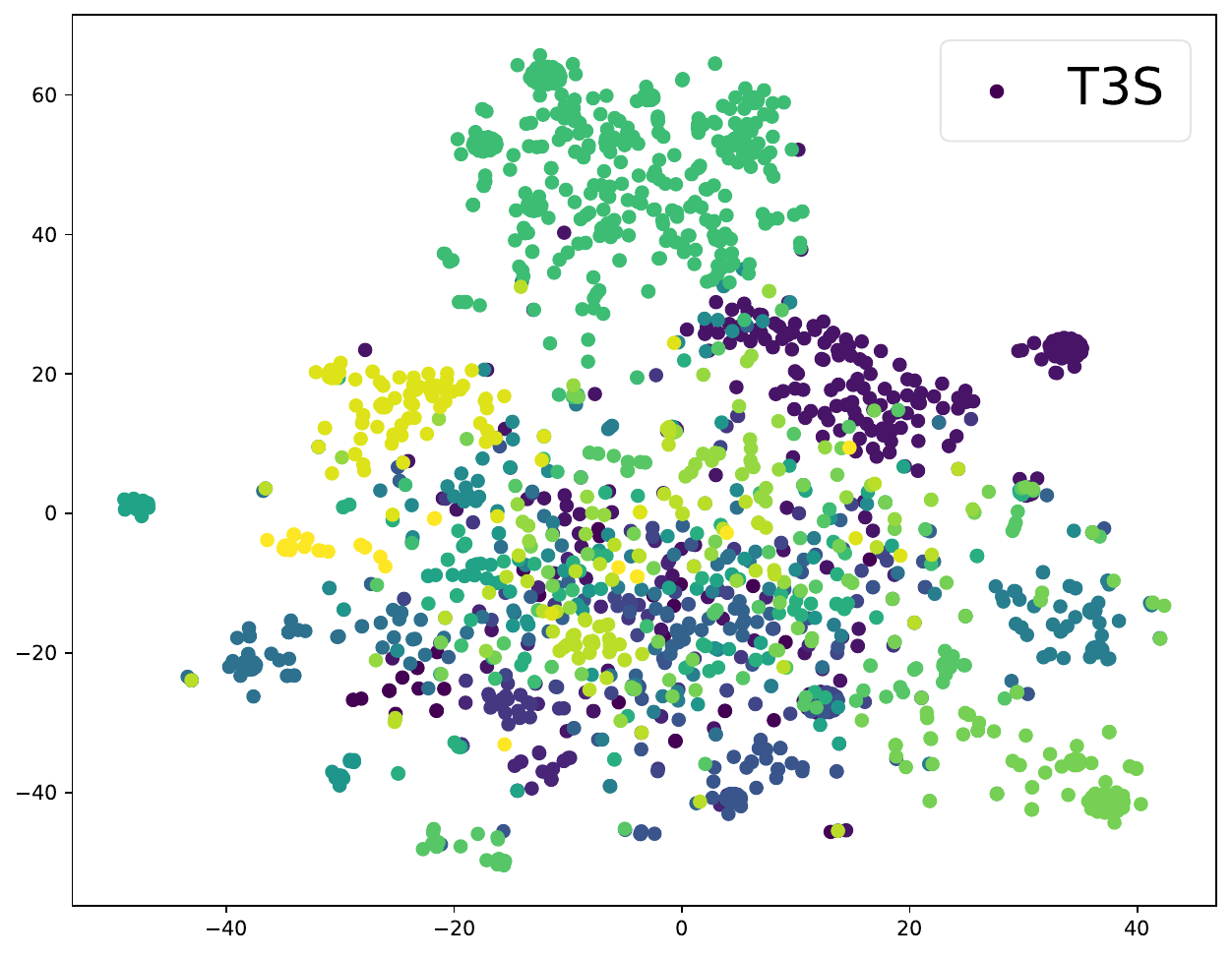}
    }\\
    \hspace{-5mm}
    \subfigure[DeepTUL]{
    \includegraphics[width=0.28\linewidth]{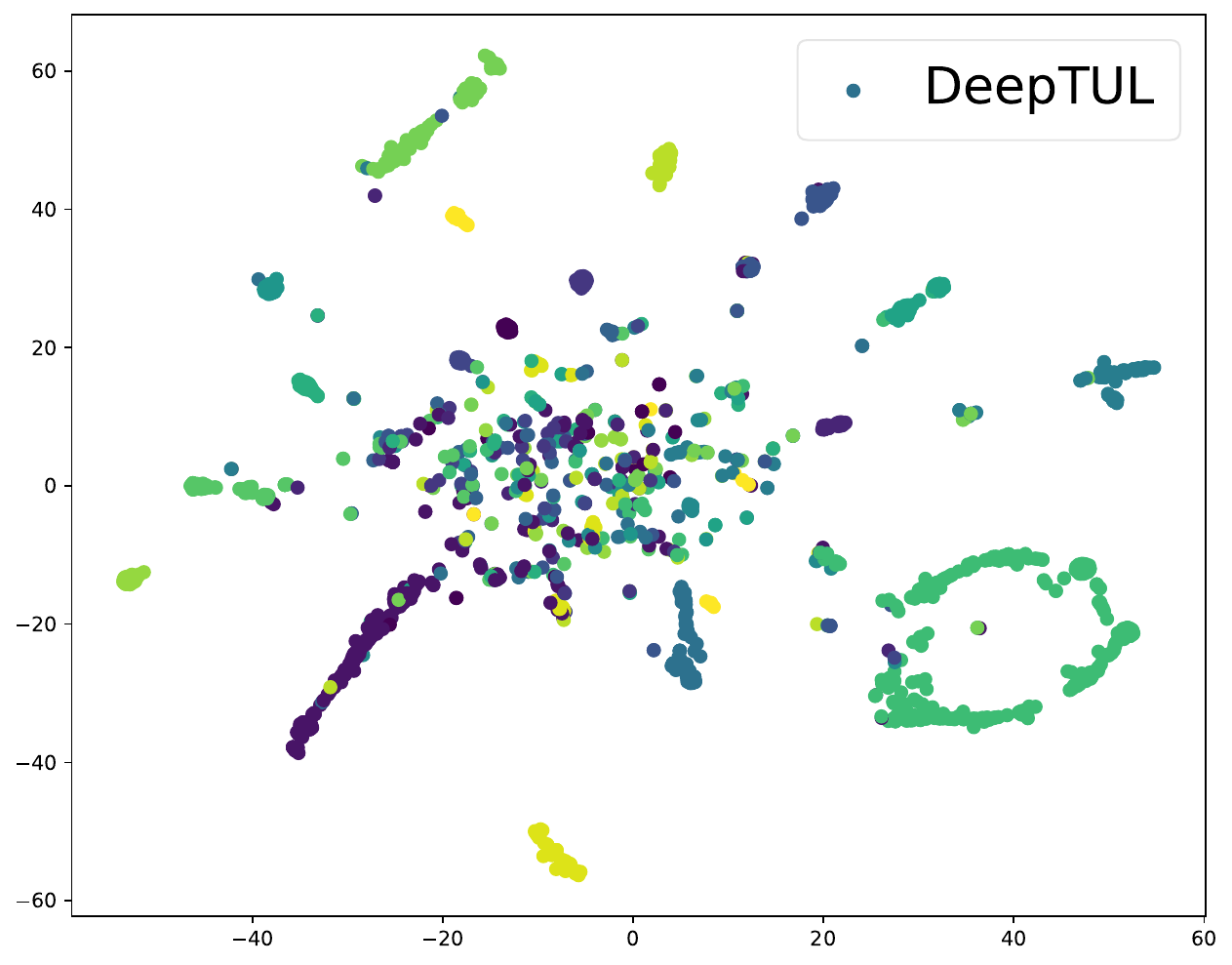}
    }
    \hspace{-3mm}
    \subfigure[GNNTUL]{
    \includegraphics[width=0.28\linewidth]{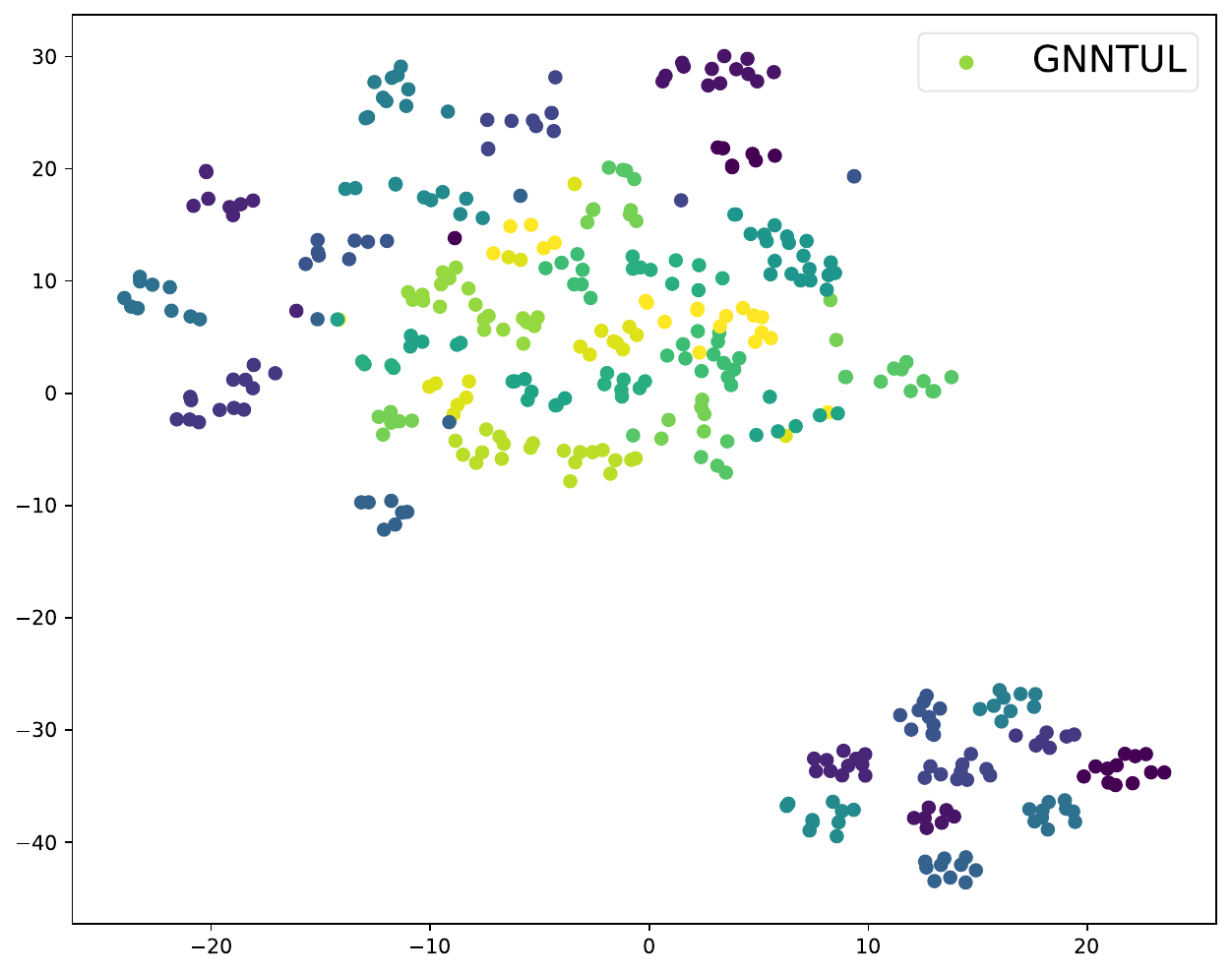}
    }
     \hspace{-3mm}
     \subfigure[\model]{
    \includegraphics[width=0.28\linewidth]{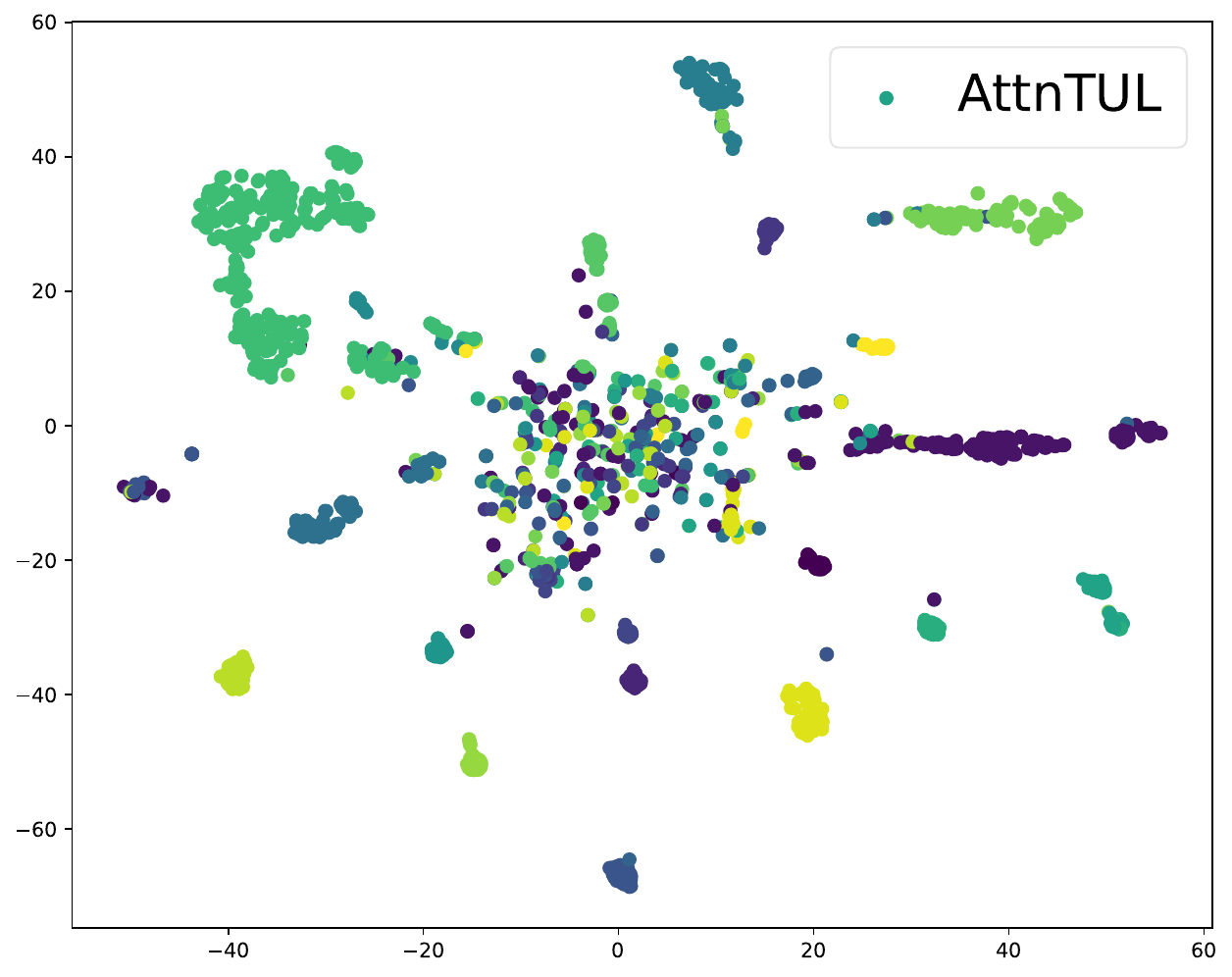}
    }
    \caption{Spatial-temporal representation visualization of trajectories learned by \model\ and other baselines on Gowalla.}
    \vspace{-0.1in}
    \label{fig:visual}
\end{figure*}

\begin{figure*}[t]
    \centering
    \hspace{-5mm}
    \subfigure[TULER-L]{
    \includegraphics[width=0.28\linewidth]{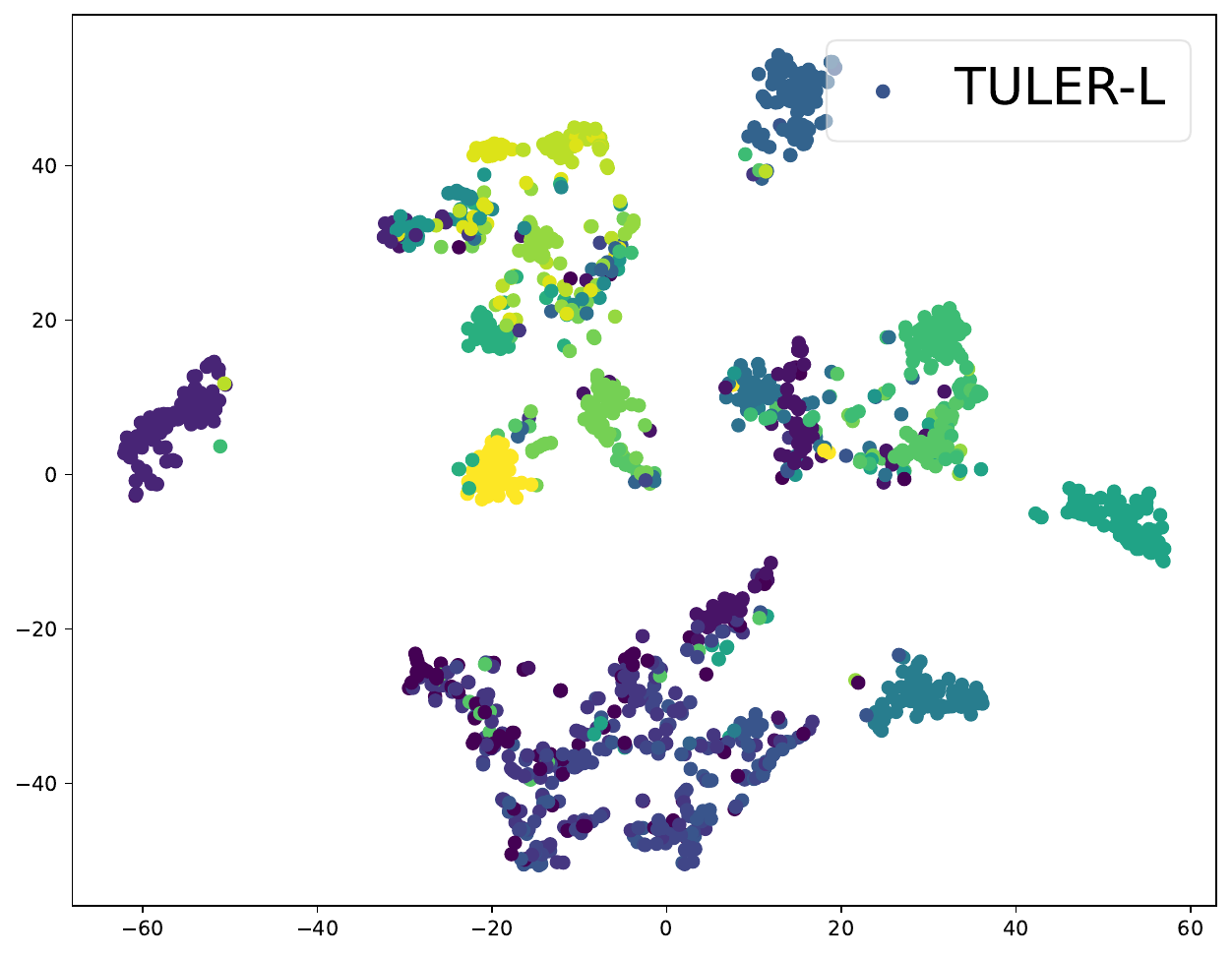}
    }
    \hspace{-3mm}
    \subfigure[TULER-G]{
    \includegraphics[width=0.28\linewidth]{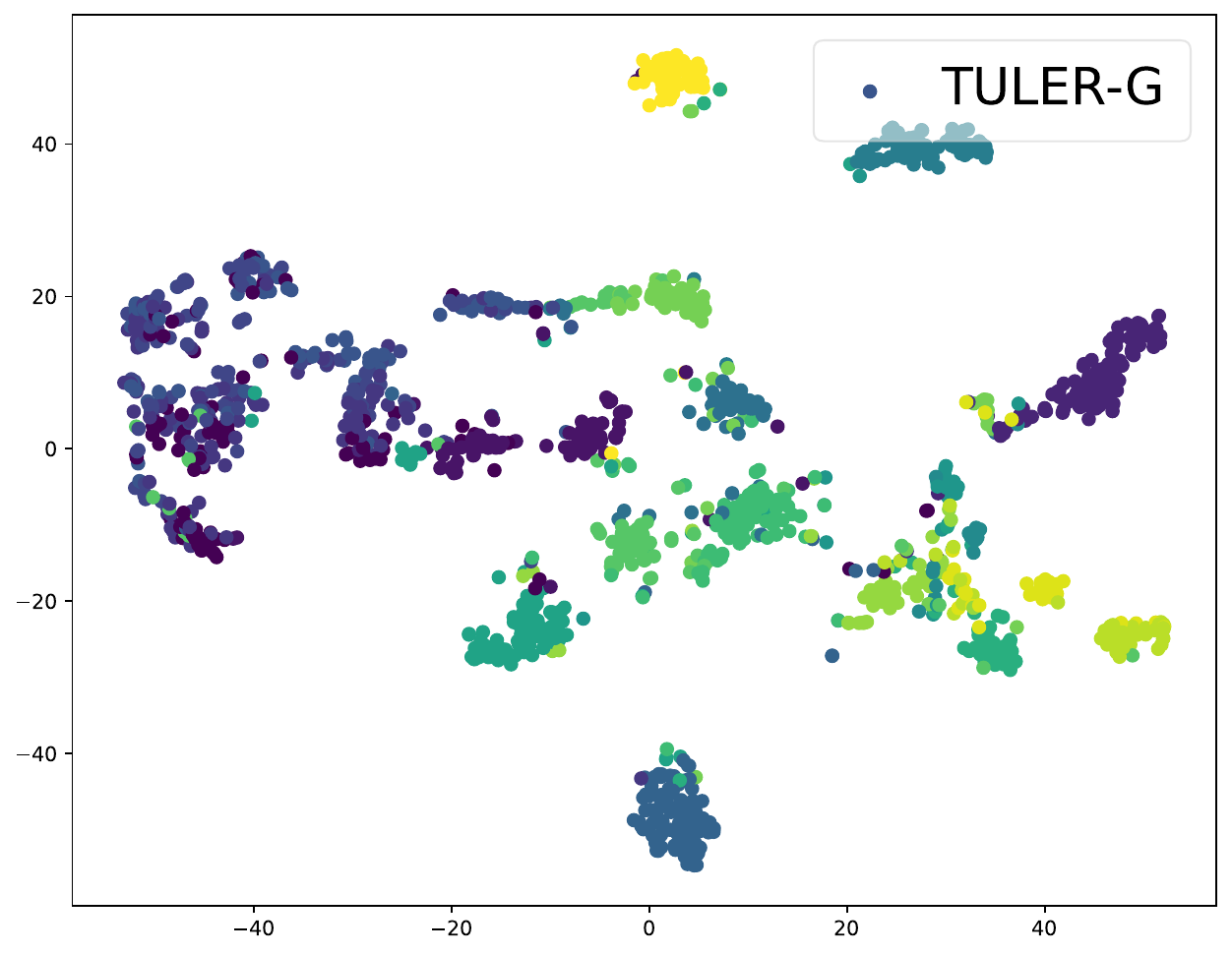}
    }
    \hspace{-3mm}
    \subfigure[Bi-TULER]{
    \includegraphics[width=0.28\linewidth]{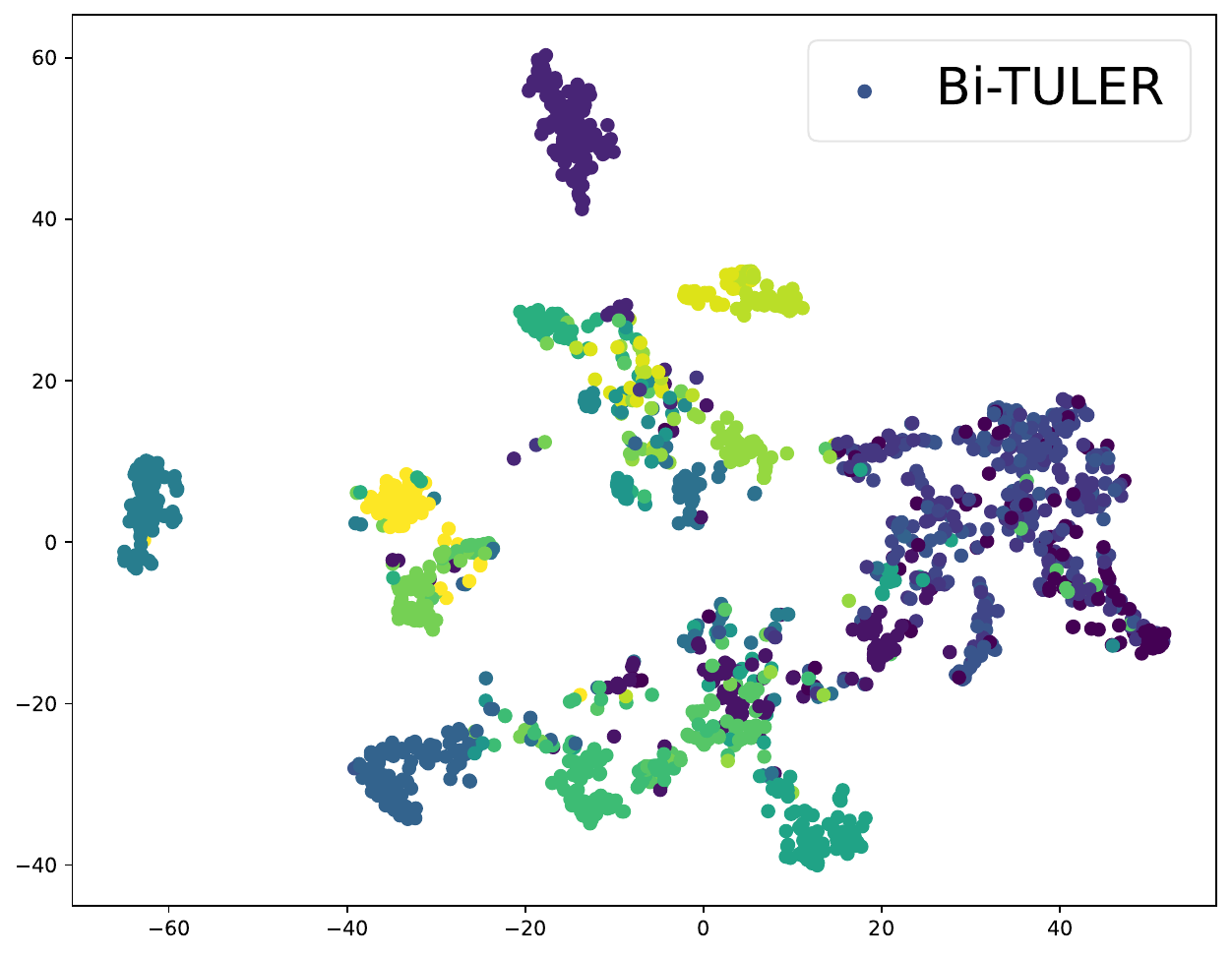}
    }\\
     \hspace{-5mm}
     \subfigure[TULVAE]{
    \includegraphics[width=0.28\linewidth]{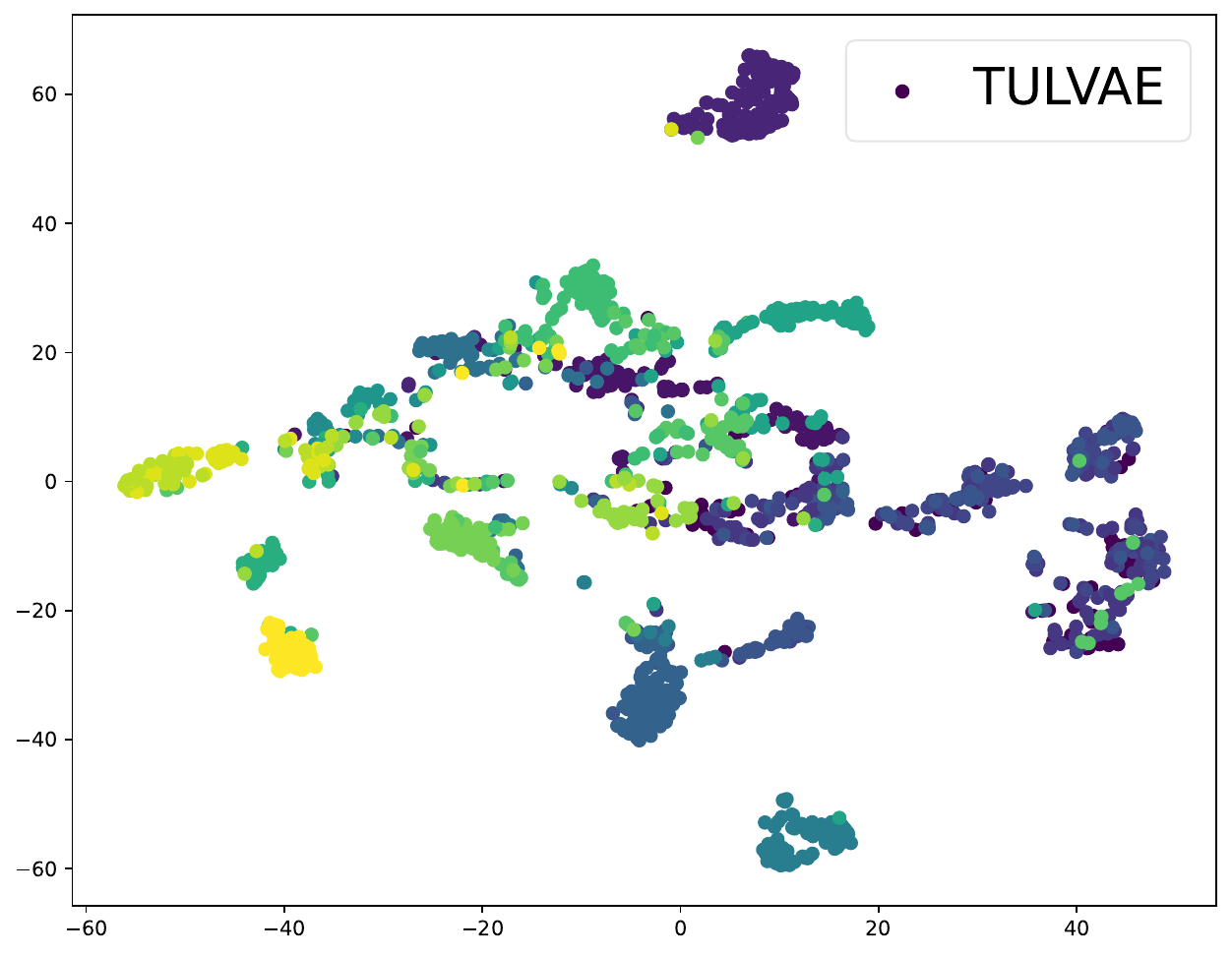}
    }
    \hspace{-3mm}
    \subfigure[DPLink]{
    \includegraphics[width=0.28\linewidth]{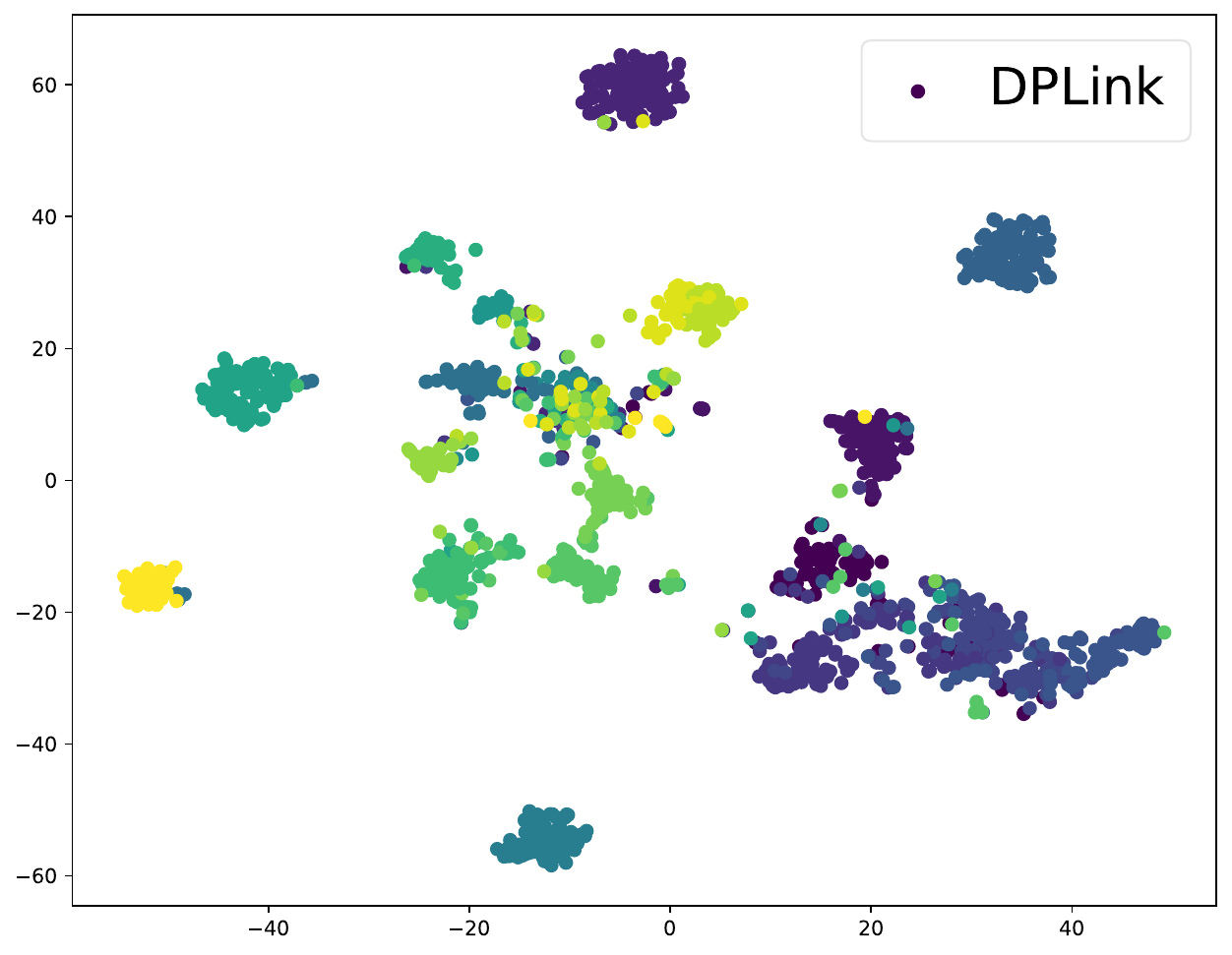}
    }
    \hspace{-3mm}
    \subfigure[T3S]{
    \includegraphics[width=0.28\linewidth]{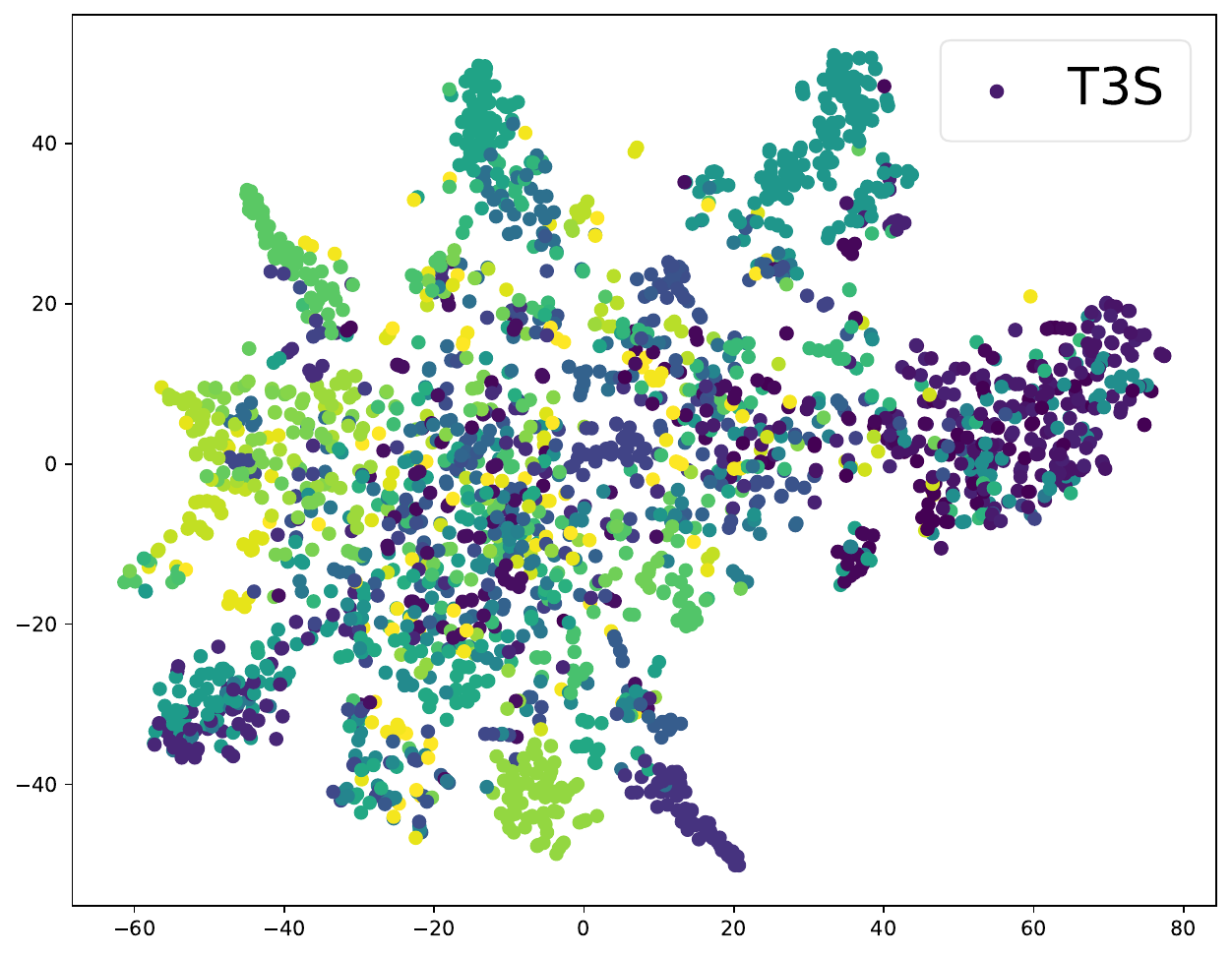}
    }\\
    \hspace{-5mm}
    \subfigure[DeepTUL]{
    \includegraphics[width=0.28\linewidth]{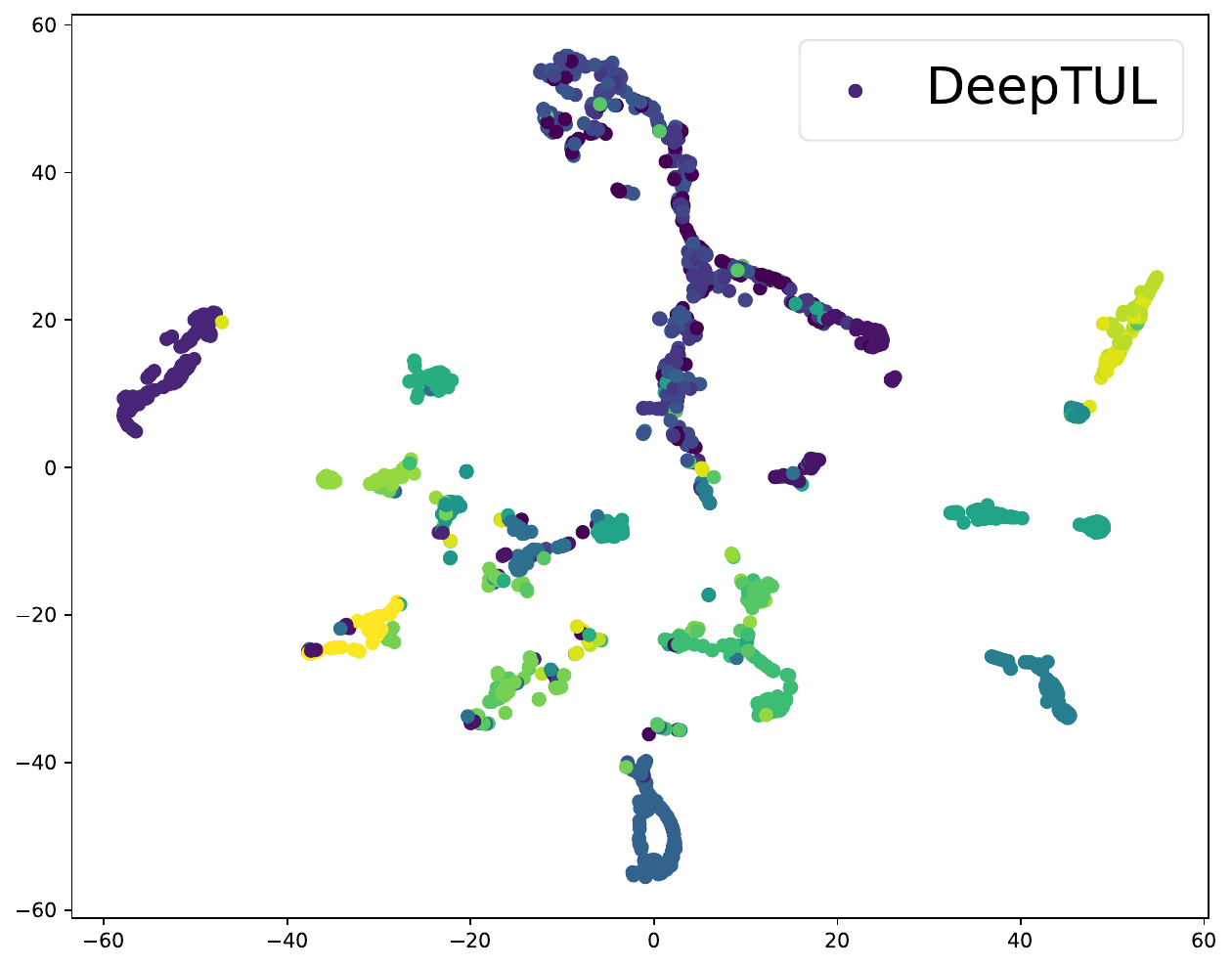}
    }
    \hspace{-3mm}
    \subfigure[GNNTUL]{
    \includegraphics[width=0.28\linewidth]{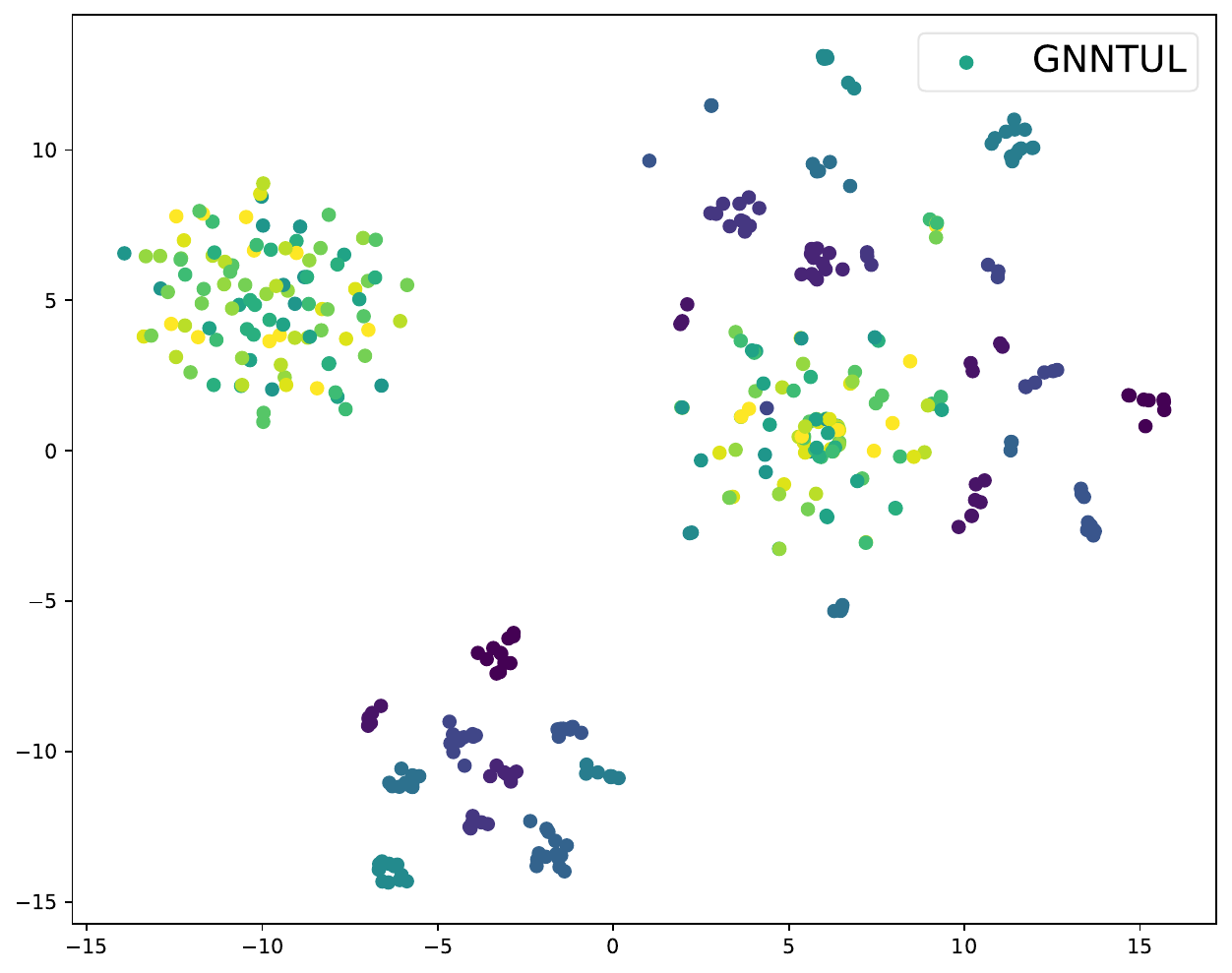}
    }
    \hspace{-3mm}
    \subfigure[\model]{
    \includegraphics[width=0.28\linewidth]{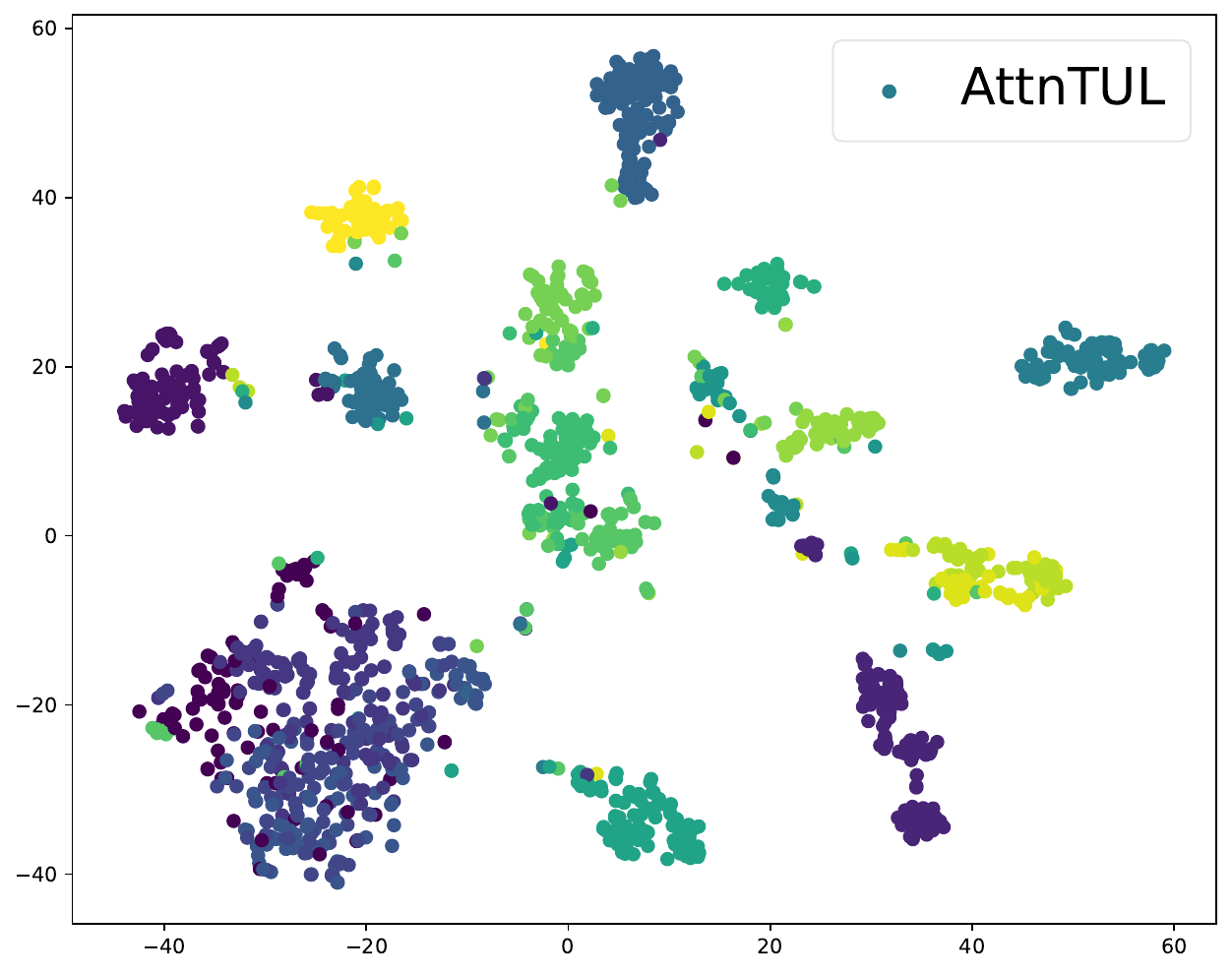}
    }
    \caption{Spatial-temporal representation visualization of trajectories learned by \model\ and other baselines on GeoLife.}
    \vspace{-0.1in}
    \label{fig:visual2}
\end{figure*}

\textit{The effect of the number of heads.} 
To investigate the effect of the number of heads in temporal self-attention encoder, we show the evaluation results with the settings of different head numbers in the third group of sub-figures in Figure~\ref{fig-parm}. 
As we can see, the performance of \model has little effect \wrt. the number of heads $\#head$ on Gowalla, first increases and then decreases on PrivateCar as $\#head$ increases, and the model performance achieves the best when $\#head=4$ on PrivateCar. 
The possible reason is that sub-trajectories in Gowalla are too sparse, so the intra-trajectory dependencies are not complicated, and thus the effect of multi-head attention mechanism is not obvious. While sub-trajectories in PrivateCar are relatively dense, and the use of multi-head attention mechanism effectively captures the complex temporal dependencies within the trajectory.

\textit{The effect of the length of time window.} 
From two right sub-figures in Figure~\ref{fig-parm}, we can see that the model performance also first increases and then decreases as the length of time window increases on both Gowalla and PrivateCar datasets. 
Considering the sparsity of Gowalla, denseness of PrivateCar, and the length of trajectories in these two datasets, we vary the length of time window from 30 to 240 minutes on Gowalla and from 2 to 30 minutes on PrivateCar. 
Specifically, the model performance achieves the best when $T_W$ is set to 2 hours on Gowalla and 10 minutes on PrivateCar. 
That is, on the sparse check-in mobility data Gowalla, when two-hour time window is set to the same encoding, the best results are obtained, indicating that check-ins within such a length of time can be considered to be indistinguishable in time.
Similarly, on the dense GPS trajectory data PrivateCar, the best performance is achieved with the same encoding for the spatial-temporal points within 10 minutes, which also shows that there is no temporal difference between the different sub-trajectories in 10-minute time window.

\subsubsection{Visualization (RQ4)}

To further verify the effectiveness of our model in learning spatiotemporal representation of trajectories, we adopt a visualization way to compare the representation vectors of trajectories learned by different models. 

To this end, we use t-SNE~\cite{van2008visualizing} to plot the latent space of trajectories learned by DNN-based models. Specifically, we randomly select 20 users and their corresponding trajectories from Gowalla and GeoLife. The learned representation of each trajectory is projected to the 2D space. 
The visualization results of learned trajectory reorientation on Gowalla ($|\mathcal{U}|=222$) and GeoLife ($|\mathcal{U}|=90$) are shown in Figure~\ref{fig:visual} and Figure~\ref{fig:visual2}, where points with the same color represent trajectories from the same user. 

From Figures~\ref{fig:visual} and~\ref{fig:visual2}, we can observe that the trajectory representations generated by our \model show an apparent clustering effect, and the clustering is tighter than other methods (\ie, Figure~\ref{fig:visual}(i) and Figure~\ref{fig:visual2}(i)). This indicates that \model is able to effectively distinguish trajectories generated by different users, which is crucial for TUL prediction task. Although GNNTUL and DeepTUL can also distinguish some clusters (users), there are obviously plenty of trajectories from different users intertwined together, and thus it is difficult to identify the correct users for these trajectories. 

Furthermore, we also observe that all baselines perform poorly on sparse check-in dataset (\ie, Gowalla) compared to the easily distinguishable dense GeoLife dataset, while our model effectively distinguishes the trajectories produced by most users (see Figure~\ref{fig:visual}(i)).

\subsubsection{Model Efficiency Analysis (RQ5)}

\begin{figure}[b]
    \begin{center}
    % \hspace{-5mm}
    \subfigure[Classic models]{
    \includegraphics[width=0.4\columnwidth]{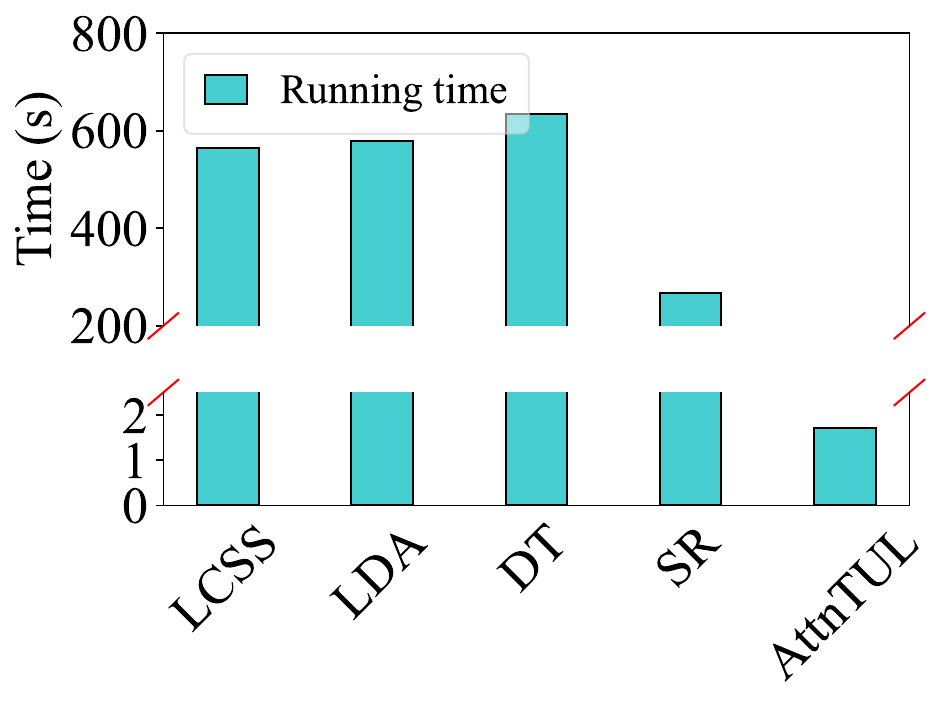}
    }
    \hspace{5mm}
    \subfigure[DNN models]{
    \includegraphics[width=0.4\columnwidth]{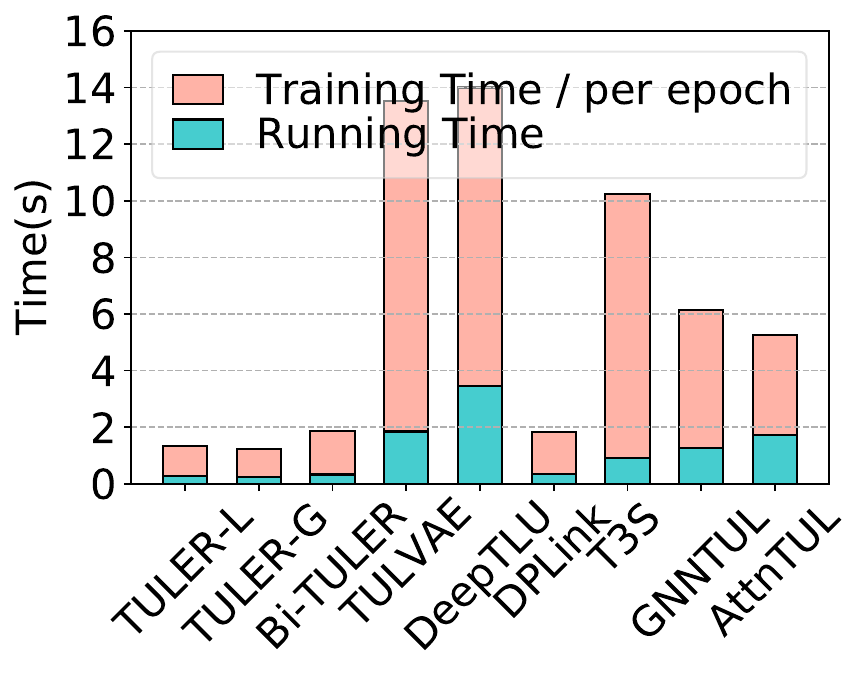}
    }
    \caption{Model Efficiency Evaluation}
    % \vspace{-4mm}
    \label{fig:time}
    \end{center}
\end{figure}

We also compare the efficiency of our \model with all baselines on Gowalla ($|\mathcal{U}|=222$) data. 
We report the experiment results \wrt. classic methods and DNN-based methods in Figure~\ref{fig:time}. 

Since classic methods have no training time, we only compare the running time. As can be seen from Figure~\ref{fig:time}(a), the prediction time of our \model is much less than the running time of classic methods. This is because the classic methods need to calculate a large number of distances between trajectories, so the time cost is relatively high. For example, LCSS, LDA, and DT all consume more than 300 times the prediction time of our model.

From the results in Figure~\ref{fig:time}(b), we can see that the running time (\ie, the testing time) of DNN-based models is much less than that of classic methods. Although the testing time of our model is not the best, it is faster than state-of-the-art GNNTUL, and the testing time is also less than 1 seconds.
Since we adopt the early stopping mechanism for all DNN-based models, the number of rounds of early stopping is different for different models. For a fair comparison, we report the average training time per each epoch. 
In terms of model training time, the efficiency of our \model is in the middle level, better than state-of-the-art models (\ie, TULVAE, T3S, DeepTUL, and GNNTUL), and worse than the simple TUL models (\ie, TULER and its variants) and DPLink. It takes around 5.6 seconds for each epoch on Gowalla using a Tesla V100 GPU card. 
Specifically, our \model is 2.4 times and 7.3 times faster than TULVAE and DeepTUL per epoch, respectively. 
The potential reason is that DeepTUL needs to use historical trajectory data, so a large number of trajectory distances need to be calculated, while TULVAE needs to use variational inference, which requires expensive time cost.

% \vspace{-2mm}
\section{Conclusion}
In this paper, we present a novel hierarchical spatio-temporal attention network, called \model, for TUL problem. 
\model effectively learns the local and global representations for each trajectory by the designed hierarchical spatio-temporal attention network to classify trajectories by users. 
\model first employs GCN on the constructed local and global graphs to learn embeddings for grids and trajectories. 
Then, it uses a hierarchical spatio-temporal attention network to obtain the local and global representations for each trajectory, respectively. 
Eventually, a linking layer is designed to fuse the two representation to classify trajectories by users. 
% We first construct a local spatial graph and a global spatial graph to model micro- and macro-spatial relationships among users' trajectories. 
% Then we employ GCN on both local and global graphs to learn embeddings for grids and trajectories, respectively. 
% Afterwords, a hierarchical spatio-temporal attention network is designed to obtain the local and global representations for each trajectory, respectively. 
% Eventually, a linking layer is designed to fuse the two representation to classify trajectories by users. 
Experiments on three real-world mobility datasets demonstrate that our model significantly outperforms state-of-the-art baselines in terms of all metrics for TUL problem. For future work, we plan to enhance our proposed \model\ model with the ability of handling streaming mobility trace data for real-time trajectory-user linking scenario.
% For future work, we are interested in inducing a more complex attention model to integrate rich semantic contexts for TUL prediction. 

\section*{Acknowledgment}
This work is partially supported by the National Natural Science Foundation of China under grant Nos. 62176243, 61773331 and 41927805. %and the National Key Research and Development Program of China under grant Nos. 2018AAA0100602 and 2019YFC1509100. Corresponding author: Yanwei Yu. 

%
% \clearpage
% \balance
\bibliographystyle{ACM-Reference-Format}
\bibliography{cite}


\begin{thebibliography}{50}

%%% the Web address.

\ifx \showCODEN    \undefined \def \showCODEN     #1{\unskip}     \fi
\ifx \showDOI      \undefined \def \showDOI       #1{#1}\fi
\ifx \showISBNx    \undefined \def \showISBNx     #1{\unskip}     \fi
\ifx \showISBNxiii \undefined \def \showISBNxiii  #1{\unskip}     \fi
\ifx \showISSN     \undefined \def \showISSN      #1{\unskip}     \fi
\ifx \showLCCN     \undefined \def \showLCCN      #1{\unskip}     \fi
\ifx \shownote     \undefined \def \shownote      #1{#1}          \fi
\ifx \showarticletitle \undefined \def \showarticletitle #1{#1}   \fi
\ifx \showURL      \undefined \def \showURL       {\relax}        \fi
% invisible to TeX
\providecommand\bibfield[2]{#2}
\providecommand\bibinfo[2]{#2}
\providecommand\natexlab[1]{#1}
\providecommand\showeprint[2][]{arXiv:#2}

\bibitem[Atev et~al\mbox{.}(2010)]%
        {Atev2010Hausdorff}
\bibfield{author}{\bibinfo{person}{Stefan Atev}, \bibinfo{person}{Grant
  Miller}, {and} \bibinfo{person}{Nikolaos~P. Papanikolopoulos}.}
  \bibinfo{year}{2010}\natexlab{}.
\newblock \showarticletitle{Clustering of Vehicle Trajectories}.
\newblock \bibinfo{journal}{\emph{IEEE Transactions on Intelligent
  Transportation Systems}} \bibinfo{volume}{11}, \bibinfo{number}{3}
  (\bibinfo{year}{2010}), \bibinfo{pages}{647--657}.
\newblock
\urldef\tempurl%
\url{https://doi.org/10.1109/TITS.2010.2048101}
\showDOI{\tempurl}


\bibitem[Bahdanau et~al\mbox{.}(2015)]%
        {bahdanau2015neural}
\bibfield{author}{\bibinfo{person}{Dzmitry Bahdanau},
  \bibinfo{person}{Kyung~Hyun Cho}, {and} \bibinfo{person}{Yoshua Bengio}.}
  \bibinfo{year}{2015}\natexlab{}.
\newblock \showarticletitle{Neural machine translation by jointly learning to
  align and translate}. In \bibinfo{booktitle}{\emph{ICLR}}.
\newblock


\bibitem[Byun et~al\mbox{.}(2023)]%
        {byun2023aspect}
\bibfield{author}{\bibinfo{person}{Hyungho Byun}, \bibinfo{person}{Younhyuk
  Choi}, {and} \bibinfo{person}{Chong-Kwon Kim}.}
  \bibinfo{year}{2023}\natexlab{}.
\newblock \showarticletitle{Aspect-oriented unsupervised social link inference
  on user trajectory data}.
\newblock \bibinfo{journal}{\emph{Information Sciences}}  \bibinfo{volume}{626}
  (\bibinfo{year}{2023}), \bibinfo{pages}{249--261}.
\newblock


\bibitem[Chen et~al\mbox{.}(2022)]%
        {chen2022MainTUL}
\bibfield{author}{\bibinfo{person}{Wei Chen}, \bibinfo{person}{Shuzhe Li},
  \bibinfo{person}{Chao Huang}, \bibinfo{person}{Yanwei Yu},
  \bibinfo{person}{Yongguo Jiang}, {and} \bibinfo{person}{Junyu Dong}.}
  \bibinfo{year}{2022}\natexlab{}.
\newblock \showarticletitle{Mutual Distillation Learning Network for
  Trajectory-User Linking}. In \bibinfo{booktitle}{\emph{IJCAI}}.
\newblock


\bibitem[Chen et~al\mbox{.}(2021)]%
        {chen2021curriculum}
\bibfield{author}{\bibinfo{person}{Yudong Chen}, \bibinfo{person}{Xin Wang},
  \bibinfo{person}{Miao Fan}, \bibinfo{person}{Jizhou Huang},
  \bibinfo{person}{Shengwen Yang}, {and} \bibinfo{person}{Wenwu Zhu}.}
  \bibinfo{year}{2021}\natexlab{}.
\newblock \showarticletitle{Curriculum Meta-Learning for Next POI
  Recommendation}. In \bibinfo{booktitle}{\emph{KDD}}.
  \bibinfo{pages}{2692--2702}.
\newblock


\bibitem[Deng et~al\mbox{.}(2023)]%
        {deng2023s2tul}
\bibfield{author}{\bibinfo{person}{Liwei Deng}, \bibinfo{person}{Hao Sun},
  \bibinfo{person}{Yan Zhao}, \bibinfo{person}{Shuncheng Liu}, {and}
  \bibinfo{person}{Kai Zheng}.} \bibinfo{year}{2023}\natexlab{}.
\newblock \showarticletitle{S2TUL: A Semi-Supervised Framework for
  Trajectory-User Linking}. In \bibinfo{booktitle}{\emph{Proceedings of the
  Sixteenth ACM International Conference on Web Search and Data Mining}}.
  \bibinfo{pages}{375--383}.
\newblock


\bibitem[Feng et~al\mbox{.}(2020)]%
        {Feng2020dplink}
\bibfield{author}{\bibinfo{person}{Jie Feng}, \bibinfo{person}{Yong Li},
  \bibinfo{person}{Mingyang Zhang}, \bibinfo{person}{Zeyu Yang},
  \bibinfo{person}{Huandong Wang}, \bibinfo{person}{Han Cao}, {and}
  \bibinfo{person}{Depeng Jin}.} \bibinfo{year}{2020}\natexlab{}.
\newblock \showarticletitle{User Identity Linkage via Co-Attentional Neural
  Network From Heterogeneous Mobility Data}.
\newblock \bibinfo{journal}{\emph{TKDE}} (\bibinfo{year}{2020}),
  \bibinfo{pages}{1--1}.
\newblock
\urldef\tempurl%
\url{https://doi.org/10.1109/TKDE.2020.2989732}
\showDOI{\tempurl}


\bibitem[Feng et~al\mbox{.}(2019)]%
        {feng2019dplink}
\bibfield{author}{\bibinfo{person}{Jie Feng}, \bibinfo{person}{Mingyang Zhang},
  \bibinfo{person}{Huandong Wang}, \bibinfo{person}{Zeyu Yang},
  \bibinfo{person}{Chao Zhang}, \bibinfo{person}{Yong Li}, {and}
  \bibinfo{person}{Depeng Jin}.} \bibinfo{year}{2019}\natexlab{}.
\newblock \showarticletitle{Dplink: User identity linkage via deep neural
  network from heterogeneous mobility data}. In
  \bibinfo{booktitle}{\emph{WWW}}. \bibinfo{pages}{459--469}.
\newblock


\bibitem[Gao et~al\mbox{.}(2020)]%
        {gao2020adversarial}
\bibfield{author}{\bibinfo{person}{Qiang Gao}, \bibinfo{person}{Fengli Zhang},
  \bibinfo{person}{Fuming Yao}, \bibinfo{person}{Ailing Li},
  \bibinfo{person}{Lin Mei}, {and} \bibinfo{person}{Fan Zhou}.}
  \bibinfo{year}{2020}\natexlab{}.
\newblock \showarticletitle{Adversarial mobility learning for human trajectory
  classification}.
\newblock \bibinfo{journal}{\emph{IEEE Access}}  \bibinfo{volume}{8}
  (\bibinfo{year}{2020}), \bibinfo{pages}{20563--20576}.
\newblock


\bibitem[Gao et~al\mbox{.}(2019)]%
        {gao2019predicting}
\bibfield{author}{\bibinfo{person}{Qiang Gao}, \bibinfo{person}{Fan Zhou},
  \bibinfo{person}{Goce Trajcevski}, \bibinfo{person}{Kunpeng Zhang},
  \bibinfo{person}{Ting Zhong}, {and} \bibinfo{person}{Fengli Zhang}.}
  \bibinfo{year}{2019}\natexlab{}.
\newblock \showarticletitle{Predicting human mobility via variational
  attention}. In \bibinfo{booktitle}{\emph{WWW}}. \bibinfo{pages}{2750--2756}.
\newblock


\bibitem[Gao et~al\mbox{.}(2017)]%
        {gao2017identifying}
\bibfield{author}{\bibinfo{person}{Qiang Gao}, \bibinfo{person}{Fan Zhou},
  \bibinfo{person}{Kunpeng Zhang}, \bibinfo{person}{Goce Trajcevski},
  \bibinfo{person}{Xucheng Luo}, {and} \bibinfo{person}{Fengli Zhang}.}
  \bibinfo{year}{2017}\natexlab{}.
\newblock \showarticletitle{Identifying Human Mobility via Trajectory
  Embeddings}. In \bibinfo{booktitle}{\emph{IJCAI}}, Vol.~\bibinfo{volume}{17}.
  \bibinfo{pages}{1689--1695}.
\newblock


\bibitem[Hamilton et~al\mbox{.}(2017)]%
        {hamilton2017inductive}
\bibfield{author}{\bibinfo{person}{William~L Hamilton}, \bibinfo{person}{Rex
  Ying}, {and} \bibinfo{person}{Jure Leskovec}.}
  \bibinfo{year}{2017}\natexlab{}.
\newblock \showarticletitle{Inductive representation learning on large graphs}.
  In \bibinfo{booktitle}{\emph{NeurIPS}}. \bibinfo{pages}{1025--1035}.
\newblock


\bibitem[Huang et~al\mbox{.}(2018)]%
        {2018deepcrime}
\bibfield{author}{\bibinfo{person}{Chao Huang}, \bibinfo{person}{Junbo Zhang},
  \bibinfo{person}{Yu Zheng}, {and} \bibinfo{person}{Nitesh~V Chawla}.}
  \bibinfo{year}{2018}\natexlab{}.
\newblock \showarticletitle{DeepCrime: Attentive hierarchical recurrent
  networks for crime prediction}. In \bibinfo{booktitle}{\emph{CIKM}}.
  \bibinfo{pages}{1423--1432}.
\newblock


\bibitem[Huang et~al\mbox{.}(2019)]%
        {huang2019road}
\bibfield{author}{\bibinfo{person}{Yourong Huang}, \bibinfo{person}{Zhu Xiao},
  \bibinfo{person}{Xiaoyou Yu}, \bibinfo{person}{Dong Wang},
  \bibinfo{person}{Vincent Havyarimana}, {and} \bibinfo{person}{Jing Bai}.}
  \bibinfo{year}{2019}\natexlab{}.
\newblock \showarticletitle{Road network construction with complex
  intersections based on sparsely sampled private car trajectory data}.
\newblock \bibinfo{journal}{\emph{TKDD}} \bibinfo{volume}{13},
  \bibinfo{number}{3} (\bibinfo{year}{2019}), \bibinfo{pages}{1--28}.
\newblock


\bibitem[Jiang(2018)]%
        {jiang2018survey}
\bibfield{author}{\bibinfo{person}{Zhe Jiang}.}
  \bibinfo{year}{2018}\natexlab{}.
\newblock \showarticletitle{A survey on spatial prediction methods}.
\newblock \bibinfo{journal}{\emph{TKDE}} \bibinfo{volume}{31},
  \bibinfo{number}{9} (\bibinfo{year}{2018}), \bibinfo{pages}{1645--1664}.
\newblock


\bibitem[Jin et~al\mbox{.}(2019)]%
        {jin2019moving}
\bibfield{author}{\bibinfo{person}{Fengmei Jin}, \bibinfo{person}{Wen Hua},
  \bibinfo{person}{Jiajie Xu}, {and} \bibinfo{person}{Xiaofang Zhou}.}
  \bibinfo{year}{2019}\natexlab{}.
\newblock \showarticletitle{Moving object linking based on historical trace}.
  In \bibinfo{booktitle}{\emph{ICDE}}. IEEE, \bibinfo{pages}{1058--1069}.
\newblock


\bibitem[Jin et~al\mbox{.}(2020)]%
        {jin2020trajectory}
\bibfield{author}{\bibinfo{person}{Fengmei Jin}, \bibinfo{person}{Wen Hua},
  \bibinfo{person}{Thomas Zhou}, \bibinfo{person}{Jiajie Xu},
  \bibinfo{person}{Matteo Francia}, \bibinfo{person}{Maria Orowska}, {and}
  \bibinfo{person}{Xiaofang Zhou}.} \bibinfo{year}{2020}\natexlab{}.
\newblock \showarticletitle{Trajectory-Based Spatiotemporal Entity Linking}.
\newblock \bibinfo{journal}{\emph{TKDE}} (\bibinfo{year}{2020}).
\newblock


\bibitem[Keogh and Pazzani(2000)]%
        {keogh2000scaling}
\bibfield{author}{\bibinfo{person}{Eamonn~J Keogh} {and}
  \bibinfo{person}{Michael~J Pazzani}.} \bibinfo{year}{2000}\natexlab{}.
\newblock \showarticletitle{Scaling up dynamic time warping for datamining
  applications}. In \bibinfo{booktitle}{\emph{KDD}}. \bibinfo{pages}{285--289}.
\newblock


\bibitem[Khandelwal et~al\mbox{.}(2018)]%
        {khandelwal2018sharp}
\bibfield{author}{\bibinfo{person}{Urvashi Khandelwal}, \bibinfo{person}{He
  He}, \bibinfo{person}{Peng Qi}, {and} \bibinfo{person}{Dan Jurafsky}.}
  \bibinfo{year}{2018}\natexlab{}.
\newblock \showarticletitle{Sharp Nearby, Fuzzy Far Away: How Neural Language
  Models Use Context}. In \bibinfo{booktitle}{\emph{ACL}}.
  \bibinfo{pages}{284--294}.
\newblock


\bibitem[Kipf and Welling(2017)]%
        {kipf2017semi}
\bibfield{author}{\bibinfo{person}{Thomas~N. Kipf} {and} \bibinfo{person}{Max
  Welling}.} \bibinfo{year}{2017}\natexlab{}.
\newblock \showarticletitle{Semi-Supervised Classification with Graph
  Convolutional Networks}. In \bibinfo{booktitle}{\emph{ICLR}}.
\newblock


\bibitem[Li et~al\mbox{.}(2018)]%
        {li2018deep}
\bibfield{author}{\bibinfo{person}{Xiucheng Li}, \bibinfo{person}{Kaiqi Zhao},
  \bibinfo{person}{Gao Cong}, \bibinfo{person}{Christian~S Jensen}, {and}
  \bibinfo{person}{Wei Wei}.} \bibinfo{year}{2018}\natexlab{}.
\newblock \showarticletitle{Deep representation learning for trajectory
  similarity computation}. In \bibinfo{booktitle}{\emph{ICDE}}. IEEE,
  \bibinfo{pages}{617--628}.
\newblock


\bibitem[Liu et~al\mbox{.}(2019)]%
        {liu2019geo}
\bibfield{author}{\bibinfo{person}{Wei Liu}, \bibinfo{person}{Zhi-Jie Wang},
  \bibinfo{person}{Bin Yao}, {and} \bibinfo{person}{Jian Yin}.}
  \bibinfo{year}{2019}\natexlab{}.
\newblock \showarticletitle{Geo-ALM: POI Recommendation by Fusing Geographical
  Information and Adversarial Learning Mechanism.}. In
  \bibinfo{booktitle}{\emph{IJCAI}}, Vol.~\bibinfo{volume}{7}.
  \bibinfo{pages}{1807--1813}.
\newblock


\bibitem[Martins and Astudillo(2016)]%
        {martins2016softmax}
\bibfield{author}{\bibinfo{person}{Andre Martins} {and} \bibinfo{person}{Ramon
  Astudillo}.} \bibinfo{year}{2016}\natexlab{}.
\newblock \showarticletitle{From softmax to sparsemax: A sparse model of
  attention and multi-label classification}. In
  \bibinfo{booktitle}{\emph{ICML}}. PMLR, \bibinfo{pages}{1614--1623}.
\newblock


\bibitem[Miao et~al\mbox{.}(2020)]%
        {miao2020trajectory}
\bibfield{author}{\bibinfo{person}{Congcong Miao}, \bibinfo{person}{Jilong
  Wang}, \bibinfo{person}{Heng Yu}, \bibinfo{person}{Weichen Zhang}, {and}
  \bibinfo{person}{Yinyao Qi}.} \bibinfo{year}{2020}\natexlab{}.
\newblock \showarticletitle{Trajectory-user linking with attentive recurrent
  network}. In \bibinfo{booktitle}{\emph{AAMAS}}. \bibinfo{pages}{878--886}.
\newblock


\bibitem[Peters et~al\mbox{.}(2019)]%
        {peters2019sparse}
\bibfield{author}{\bibinfo{person}{Ben Peters}, \bibinfo{person}{Vlad Niculae},
  {and} \bibinfo{person}{Andr{\'e}~FT Martins}.}
  \bibinfo{year}{2019}\natexlab{}.
\newblock \showarticletitle{Sparse Sequence-to-Sequence Models}. In
  \bibinfo{booktitle}{\emph{ACL}}. \bibinfo{pages}{1504--1519}.
\newblock


\bibitem[Shahdoosti and Mirzapour(2017)]%
        {shahdoosti2017spectral}
\bibfield{author}{\bibinfo{person}{Hamid~Reza Shahdoosti} {and}
  \bibinfo{person}{Fardin Mirzapour}.} \bibinfo{year}{2017}\natexlab{}.
\newblock \showarticletitle{Spectral--spatial feature extraction using
  orthogonal linear discriminant analysis for classification of hyperspectral
  data}.
\newblock \bibinfo{journal}{\emph{European Journal of Remote Sensing}}
  \bibinfo{volume}{50}, \bibinfo{number}{1} (\bibinfo{year}{2017}),
  \bibinfo{pages}{111--124}.
\newblock


\bibitem[Shang et~al\mbox{.}(2017)]%
        {shang2017trajectory}
\bibfield{author}{\bibinfo{person}{Shuo Shang}, \bibinfo{person}{Lisi Chen},
  \bibinfo{person}{Zhewei Wei}, \bibinfo{person}{Christian~S{\o}ndergaard
  Jensen}, \bibinfo{person}{Kai Zheng}, {and} \bibinfo{person}{Panos Kalnis}.}
  \bibinfo{year}{2017}\natexlab{}.
\newblock \showarticletitle{Trajectory similarity join in spatial networks}.
\newblock \bibinfo{journal}{\emph{VLDB}} \bibinfo{volume}{10},
  \bibinfo{number}{11} (\bibinfo{year}{2017}).
\newblock


\bibitem[Sun et~al\mbox{.}(2021)]%
        {sun2021trajectory}
\bibfield{author}{\bibinfo{person}{Tao Sun}, \bibinfo{person}{Yongjun Xu},
  \bibinfo{person}{Fei Wang}, \bibinfo{person}{Lin Wu},
  \bibinfo{person}{Tangwen Qian}, {and} \bibinfo{person}{Zezhi Shao}.}
  \bibinfo{year}{2021}\natexlab{}.
\newblock \showarticletitle{Trajectory-User Link with Attention Recurrent
  Networks}. In \bibinfo{booktitle}{\emph{ICPR}}. IEEE,
  \bibinfo{pages}{4589--4596}.
\newblock


\bibitem[Van~der Maaten and Hinton(2008)]%
        {van2008visualizing}
\bibfield{author}{\bibinfo{person}{Laurens Van~der Maaten} {and}
  \bibinfo{person}{Geoffrey Hinton}.} \bibinfo{year}{2008}\natexlab{}.
\newblock \showarticletitle{Visualizing data using t-SNE.}
\newblock \bibinfo{journal}{\emph{Journal of machine learning research}}
  \bibinfo{volume}{9}, \bibinfo{number}{11} (\bibinfo{year}{2008}).
\newblock


\bibitem[Vaswani et~al\mbox{.}(2017)]%
        {vaswani2017attention}
\bibfield{author}{\bibinfo{person}{Ashish Vaswani}, \bibinfo{person}{Noam
  Shazeer}, \bibinfo{person}{Niki Parmar}, \bibinfo{person}{Jakob Uszkoreit},
  \bibinfo{person}{Llion Jones}, \bibinfo{person}{Aidan~N Gomez},
  \bibinfo{person}{{\L}ukasz Kaiser}, {and} \bibinfo{person}{Illia
  Polosukhin}.} \bibinfo{year}{2017}\natexlab{}.
\newblock \showarticletitle{Attention is all you need}. In
  \bibinfo{booktitle}{\emph{NeurIPS}}. \bibinfo{pages}{5998--6008}.
\newblock


\bibitem[Wang et~al\mbox{.}(2019b)]%
        {wang2019empowering}
\bibfield{author}{\bibinfo{person}{Jingyuan Wang}, \bibinfo{person}{Ning Wu},
  \bibinfo{person}{Wayne~Xin Zhao}, \bibinfo{person}{Fanzhang Peng}, {and}
  \bibinfo{person}{Xin Lin}.} \bibinfo{year}{2019}\natexlab{b}.
\newblock \showarticletitle{Empowering A* search algorithms with neural
  networks for personalized route recommendation}. In
  \bibinfo{booktitle}{\emph{KDD}}. \bibinfo{pages}{539--547}.
\newblock


\bibitem[Wang et~al\mbox{.}(2019a)]%
        {wang2019spatiotemporal}
\bibfield{author}{\bibinfo{person}{Pengyang Wang}, \bibinfo{person}{Xiaolin
  Li}, \bibinfo{person}{Yu Zheng}, \bibinfo{person}{Charu Aggarwal}, {and}
  \bibinfo{person}{Yanjie Fu}.} \bibinfo{year}{2019}\natexlab{a}.
\newblock \showarticletitle{Spatiotemporal representation learning for driving
  behavior analysis: A joint perspective of peer and temporal dependencies}.
\newblock \bibinfo{journal}{\emph{TKDE}} (\bibinfo{year}{2019}).
\newblock


\bibitem[Wang et~al\mbox{.}(2020)]%
        {wang2020deep}
\bibfield{author}{\bibinfo{person}{Senzhang Wang}, \bibinfo{person}{Jiannong
  Cao}, {and} \bibinfo{person}{Philip Yu}.} \bibinfo{year}{2020}\natexlab{}.
\newblock \showarticletitle{Deep learning for spatio-temporal data mining: A
  survey}.
\newblock \bibinfo{journal}{\emph{TKDE}} (\bibinfo{year}{2020}).
\newblock


\bibitem[Wu et~al\mbox{.}(2020)]%
        {wu2020comprehensive}
\bibfield{author}{\bibinfo{person}{Zonghan Wu}, \bibinfo{person}{Shirui Pan},
  \bibinfo{person}{Fengwen Chen}, \bibinfo{person}{Guodong Long},
  \bibinfo{person}{Chengqi Zhang}, {and} \bibinfo{person}{S~Yu Philip}.}
  \bibinfo{year}{2020}\natexlab{}.
\newblock \showarticletitle{A comprehensive survey on graph neural networks}.
\newblock \bibinfo{journal}{\emph{TNNLS}} \bibinfo{volume}{32},
  \bibinfo{number}{1} (\bibinfo{year}{2020}), \bibinfo{pages}{4--24}.
\newblock


\bibitem[Yang et~al\mbox{.}(2021)]%
        {yang2021t3s}
\bibfield{author}{\bibinfo{person}{Peilun Yang}, \bibinfo{person}{Hanchen
  Wang}, \bibinfo{person}{Ying Zhang}, \bibinfo{person}{Lu Qin},
  \bibinfo{person}{Wenjie Zhang}, {and} \bibinfo{person}{Xuemin Lin}.}
  \bibinfo{year}{2021}\natexlab{}.
\newblock \showarticletitle{T3S: Effective Representation Learning for
  Trajectory Similarity Computation}. In \bibinfo{booktitle}{\emph{ICDE}}.
  IEEE, \bibinfo{pages}{2183--2188}.
\newblock


\bibitem[Yao et~al\mbox{.}(2019)]%
        {yao2019computing}
\bibfield{author}{\bibinfo{person}{Di Yao}, \bibinfo{person}{Gao Cong},
  \bibinfo{person}{Chao Zhang}, {and} \bibinfo{person}{Jingping Bi}.}
  \bibinfo{year}{2019}\natexlab{}.
\newblock \showarticletitle{Computing trajectory similarity in linear time: A
  generic seed-guided neural metric learning approach}. In
  \bibinfo{booktitle}{\emph{ICDE}}. IEEE, \bibinfo{pages}{1358--1369}.
\newblock


\bibitem[Yao et~al\mbox{.}(2020)]%
        {yao2020linear}
\bibfield{author}{\bibinfo{person}{Di Yao}, \bibinfo{person}{Gao Cong},
  \bibinfo{person}{Chao Zhang}, \bibinfo{person}{Xuying Meng},
  \bibinfo{person}{Rongchang Duan}, {and} \bibinfo{person}{Jingping Bi}.}
  \bibinfo{year}{2020}\natexlab{}.
\newblock \showarticletitle{A Linear Time Approach to Computing Time Series
  Similarity based on Deep Metric Learning}.
\newblock \bibinfo{journal}{\emph{TKDE}} (\bibinfo{year}{2020}).
\newblock


\bibitem[Ying et~al\mbox{.}(2010)]%
        {ying2010mining}
\bibfield{author}{\bibinfo{person}{Josh Jia-Ching Ying}, \bibinfo{person}{Eric
  Hsueh-Chan Lu}, \bibinfo{person}{Wang-Chien Lee}, \bibinfo{person}{Tz-Chiao
  Weng}, {and} \bibinfo{person}{Vincent~S Tseng}.}
  \bibinfo{year}{2010}\natexlab{}.
\newblock \showarticletitle{Mining user similarity from semantic trajectories}.
  In \bibinfo{booktitle}{\emph{SIGSPATIAL Workshop on LBSNs}}.
  \bibinfo{pages}{19--26}.
\newblock


\bibitem[Yu et~al\mbox{.}(2020)]%
        {yu2020tulsn}
\bibfield{author}{\bibinfo{person}{Yong Yu}, \bibinfo{person}{Haina Tang},
  \bibinfo{person}{Fei Wang}, \bibinfo{person}{Lin Wu},
  \bibinfo{person}{Tangwen Qian}, \bibinfo{person}{Tao Sun}, {and}
  \bibinfo{person}{Yongjun Xu}.} \bibinfo{year}{2020}\natexlab{}.
\newblock \showarticletitle{TULSN: Siamese Network for Trajectory-user
  Linking}. In \bibinfo{booktitle}{\emph{IJCNN}}. IEEE, \bibinfo{pages}{1--8}.
\newblock


\bibitem[Zhang et~al\mbox{.}(2020)]%
        {zhang2020trajectory}
\bibfield{author}{\bibinfo{person}{Hanyuan Zhang}, \bibinfo{person}{Xinyu
  Zhang}, \bibinfo{person}{Qize Jiang}, \bibinfo{person}{Baihua Zheng},
  \bibinfo{person}{Zhenbang Sun}, \bibinfo{person}{Weiwei Sun}, {and}
  \bibinfo{person}{Changhu Wang}.} \bibinfo{year}{2020}\natexlab{}.
\newblock \showarticletitle{Trajectory Similarity Learning with Auxiliary
  Supervision and Optimal Matching}. In \bibinfo{booktitle}{\emph{IJCAI}}.
  AAAI, \bibinfo{pages}{3209--3215}.
\newblock


\bibitem[Zhang et~al\mbox{.}(2019)]%
        {zhang2019decomposition}
\bibfield{author}{\bibinfo{person}{Mingyang Zhang}, \bibinfo{person}{Tong Li},
  \bibinfo{person}{Hongzhi Shi}, \bibinfo{person}{Yong Li},
  \bibinfo{person}{Pan Hui}, {et~al\mbox{.}}} \bibinfo{year}{2019}\natexlab{}.
\newblock \showarticletitle{A decomposition approach for urban anomaly
  detection across spatiotemporal data}. In \bibinfo{booktitle}{\emph{IJCAI}}.
\newblock


\bibitem[Zheng(2015)]%
        {zheng2015trajectory}
\bibfield{author}{\bibinfo{person}{Yu Zheng}.} \bibinfo{year}{2015}\natexlab{}.
\newblock \showarticletitle{Trajectory data mining: an overview}.
\newblock \bibinfo{journal}{\emph{TIST}} \bibinfo{volume}{6},
  \bibinfo{number}{3} (\bibinfo{year}{2015}), \bibinfo{pages}{1--41}.
\newblock


\bibitem[Zheng et~al\mbox{.}(2008)]%
        {zheng2008understanding}
\bibfield{author}{\bibinfo{person}{Yu Zheng}, \bibinfo{person}{Quannan Li},
  \bibinfo{person}{Yukun Chen}, \bibinfo{person}{Xing Xie}, {and}
  \bibinfo{person}{Wei-Ying Ma}.} \bibinfo{year}{2008}\natexlab{}.
\newblock \showarticletitle{Understanding mobility based on GPS data}. In
  \bibinfo{booktitle}{\emph{UbiComp}}. \bibinfo{pages}{312--321}.
\newblock


\bibitem[Zheng et~al\mbox{.}(2010)]%
        {zheng2010geolife}
\bibfield{author}{\bibinfo{person}{Yu Zheng}, \bibinfo{person}{Xing Xie},
  \bibinfo{person}{Wei-Ying Ma}, {et~al\mbox{.}}}
  \bibinfo{year}{2010}\natexlab{}.
\newblock \showarticletitle{Geolife: A collaborative social networking service
  among user, location and trajectory.}
\newblock \bibinfo{journal}{\emph{IEEE Data Eng. Bull.}} \bibinfo{volume}{33},
  \bibinfo{number}{2} (\bibinfo{year}{2010}), \bibinfo{pages}{32--39}.
\newblock


\bibitem[Zhou et~al\mbox{.}(2021a)]%
        {zhou2021trajectory}
\bibfield{author}{\bibinfo{person}{Fan Zhou}, \bibinfo{person}{Shupei Chen},
  \bibinfo{person}{Jin Wu}, \bibinfo{person}{Chengtai Cao}, {and}
  \bibinfo{person}{Shengming Zhang}.} \bibinfo{year}{2021}\natexlab{a}.
\newblock \showarticletitle{Trajectory-user linking via graph neural network}.
  In \bibinfo{booktitle}{\emph{ICC 2021-IEEE International Conference on
  Communications}}. IEEE, \bibinfo{pages}{1--6}.
\newblock


\bibitem[Zhou et~al\mbox{.}(2021b)]%
        {zhou2021self}
\bibfield{author}{\bibinfo{person}{Fan Zhou}, \bibinfo{person}{Yurou Dai},
  \bibinfo{person}{Qiang Gao}, \bibinfo{person}{Pengyu Wang}, {and}
  \bibinfo{person}{Ting Zhong}.} \bibinfo{year}{2021}\natexlab{b}.
\newblock \showarticletitle{Self-supervised human mobility learning for next
  location prediction and trajectory classification}.
\newblock \bibinfo{journal}{\emph{Knowledge-Based Systems}}
  (\bibinfo{year}{2021}), \bibinfo{pages}{107214}.
\newblock


\bibitem[Zhou et~al\mbox{.}(2018)]%
        {zhou2018trajectory}
\bibfield{author}{\bibinfo{person}{Fan Zhou}, \bibinfo{person}{Qiang Gao},
  \bibinfo{person}{Goce Trajcevski}, \bibinfo{person}{Kunpeng Zhang},
  \bibinfo{person}{Ting Zhong}, {and} \bibinfo{person}{Fengli Zhang}.}
  \bibinfo{year}{2018}\natexlab{}.
\newblock \showarticletitle{Trajectory-User Linking via Variational
  AutoEncoder}. In \bibinfo{booktitle}{\emph{IJCAI}}.
  \bibinfo{pages}{3212--3218}.
\newblock


\bibitem[Zhou et~al\mbox{.}(2021c)]%
        {zhou2021improving}
\bibfield{author}{\bibinfo{person}{Fan Zhou}, \bibinfo{person}{Ruiyang Yin},
  \bibinfo{person}{Goce Trajcevski}, \bibinfo{person}{Kunpeng Zhang},
  \bibinfo{person}{Jin Wu}, {and} \bibinfo{person}{Ashfaq Khokhar}.}
  \bibinfo{year}{2021}\natexlab{c}.
\newblock \showarticletitle{Improving human mobility identification with
  trajectory augmentation}.
\newblock \bibinfo{journal}{\emph{GeoInformatica}} \bibinfo{volume}{25},
  \bibinfo{number}{3} (\bibinfo{year}{2021}), \bibinfo{pages}{453--483}.
\newblock


\bibitem[Zhou et~al\mbox{.}(2019)]%
        {zhou2019context}
\bibfield{author}{\bibinfo{person}{Fan Zhou}, \bibinfo{person}{Xiaoli Yue},
  \bibinfo{person}{Goce Trajcevski}, \bibinfo{person}{Ting Zhong}, {and}
  \bibinfo{person}{Kunpeng Zhang}.} \bibinfo{year}{2019}\natexlab{}.
\newblock \showarticletitle{Context-aware variational trajectory encoding and
  human mobility inference}. In \bibinfo{booktitle}{\emph{WWW}}.
  \bibinfo{pages}{3469--3475}.
\newblock


\bibitem[Zhu et~al\mbox{.}(2012)]%
        {zhu2012inferring}
\bibfield{author}{\bibinfo{person}{Yin Zhu}, \bibinfo{person}{Yu Zheng},
  \bibinfo{person}{Liuhang Zhang}, \bibinfo{person}{Darshan Santani},
  \bibinfo{person}{Xing Xie}, {and} \bibinfo{person}{Qiang Yang}.}
  \bibinfo{year}{2012}\natexlab{}.
\newblock \showarticletitle{Inferring taxi status using gps trajectories}.
\newblock \bibinfo{journal}{\emph{arXiv preprint arXiv:1205.4378}}
  (\bibinfo{year}{2012}).
\newblock


\end{thebibliography}

\end{document}